\documentclass[mnsc,nonblindrev]{informs3}

\OneAndAHalfSpacedXI

\usepackage{booktabs}
\usepackage{algorithm,algorithmic}
\usepackage{tikz}
\usetikzlibrary{automata,positioning}
\usetikzlibrary{shadows}
\usetikzlibrary{shapes,shapes.geometric}
\usepackage{subcaption}
\usepackage{soul}

\usepackage{thmtools}
\usepackage{thm-restate}
\usepackage{multirow}
\usepackage[shortlabels]{enumitem}

\def\bold#1{{\sffamily \bfseries#1}}
\def\leaveline{\vspace{0.3cm}}

\def\begi{\begin{itemize}}
\def\endi{\end{itemize}}

\def\fx{\pmb{x}}

\def\bS{\mathbb{S}}
\def\bA{\mathbb{A}}
\def\bE{\mathbb{E}}
\def\cP{\mathcal{P}}
\def\blambda{\pmb{\lambda}}
\def\cM{\mathcal{M}}
\def\cH{\mathcal{H}}
\def\ALG{\texttt{ALG}}
\def\PR{\texttt{PR}}
\def\vmin{v_{\text{min}}}

\graphicspath{ {images/} }

\usepackage{natbib}
\bibpunct[, ]{(}{)}{,}{a}{}{,}%
\def\bibfont{\small}%
\TheoremsNumberedThrough     
\ECRepeatTheorems

\EquationsNumberedThrough    

\MANUSCRIPTNO{}

\begin{document}

\RUNAUTHOR{\bold{Iyengar and Singal}}
\RUNTITLE{\sffamily Conversion Funnel Optimization}
\TITLE{\Large \bold{Model-Free Approximate Bayesian Learning \\ for Large-Scale Conversion Funnel Optimization}} 

\ARTICLEAUTHORS{%
  \AUTHOR{\sffamily Garud Iyengar}
  \AFF{IEOR, Columbia University,  \EMAIL{\texttt{garud@ieor.columbia.edu}}}
  \AUTHOR{\sffamily Raghav Singal}
  \AFF{Tuck School of Business, Dartmouth College,  \EMAIL{\texttt{singal@dartmouth.edu}}}
} 

\ABSTRACT{%
The flexibility of choosing the ad action as a function of the consumer state is critical for modern-day marketing campaigns. We study the problem of identifying the optimal sequential personalized interventions that maximize the adoption probability for a new product. We model consumer behavior by a conversion funnel that captures the state of each consumer (e.g., interaction history with the firm) and allows the consumer behavior to vary as a function of both her state and firm's sequential interventions. We show our model captures consumer behavior with very high accuracy (out-of-sample AUC of over 0.95) in a real-world email marketing dataset. However, it results in a very large-scale learning problem, where the firm must learn the state-specific effects of various interventions from consumer interactions. We propose a novel attribution-based decision-making algorithm for this problem that we call \emph{model-free approximate Bayesian learning}. Our algorithm inherits the interpretability and scalability of Thompson sampling for bandits and maintains an approximate belief over the value of each state-specific intervention. The belief is updated as the algorithm interacts with the consumers. Despite being an approximation to the Bayes update, we prove the asymptotic optimality of our algorithm and analyze its convergence rate. We show that our algorithm significantly outperforms traditional approaches on extensive simulations calibrated to a real-world email marketing dataset. 
}%

\KEYWORDS{Sequential marketing, Markov decision process, scalability, interpretability, attribution}  

\HISTORY{This version: January 2024. }

\maketitle

\section{Introduction} \label{sec:intro}
Over the last two decades, digitization has been drastically shifting the way businesses operate. A significant portion of a modern-day organization's operations (ranging from marketing to sales) happen over the Internet, boosting the size of the digital economy to around 15\% of global GDP \citep{2022WB}. This shift from offline to online operations has provided businesses access to ``big data'', enabling them to understand customer behavior at a \textit{personalized} level. The challenge now is to map this ``big data'' into better decisions, in order to enhance customer experience and ultimately, boost revenue. 

\begin{figure}[ht]
\centering
\begin{subfigure}[b]{0.3\textwidth}
\centering
\includegraphics[scale=0.3]{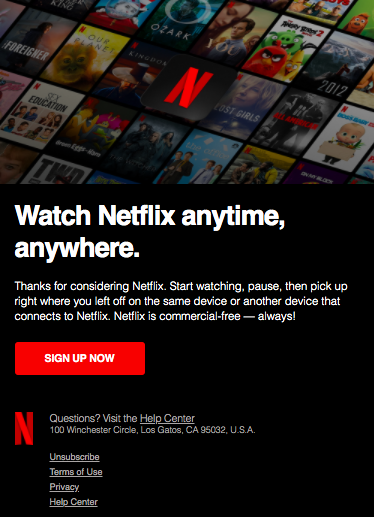}
\caption*{\small \#1: May 3, 2020}
\label{fig:CFONetflixOne}
\end{subfigure}
\begin{subfigure}[b]{0.3\textwidth}
\centering
\includegraphics[scale=0.3]{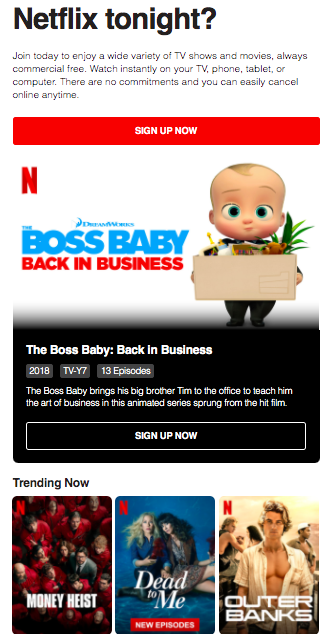}
\caption*{\small \#2: May 18, 2020}
\label{fig:CFONetflixTwo}
\end{subfigure}
\begin{subfigure}[b]{0.3\textwidth}
\centering
\includegraphics[scale=0.3]{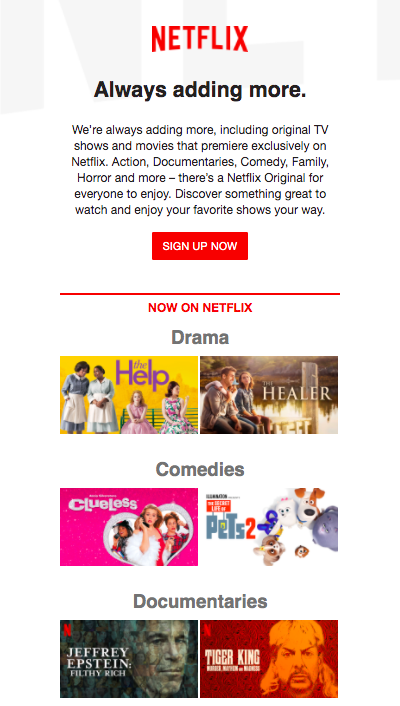}
\caption*{\small \#3: June 8, 2020}
\label{fig:CFONetflixThree}
\end{subfigure}
\caption{\normalfont Sequence of emails received by one of the authors after providing his email to Netflix (but not subscribing to the membership). The emails were sent with a 15-20 days gap in between (May 3, May 18, and June 8) and the contents of each of the email were unique. The subject line of the three emails were ``Movies \& TV shows your way'', ``Watch TV shows \& movies anytime, anywhere'', and ``Netflix - something for everyone'', respectively.} 
\label{fig:CFONetflix}
\end{figure}

We consider the decision problem of a firm promoting a product or
service online (e.g., Netflix selling its paid membership service), which we refer to as the \textit{conversion funnel optimization} problem (formally defined in \S\ref{sec:model}).  
Informally, the
goal is to maximize the \textit{conversion probability}, i.e.,  maximize the probability that a new
customer lead (\textit{consumer})
buys its product. To do so, the firm performs \textit{sequential
  interventions} (e.g., Netflix sends 
promotional emails as shown in Figure~\ref{fig:CFONetflix}) as a function 
of the information it collects on the consumer (\textit{state}). The
intervention space of the firm includes multiple \textit{actions} (e.g., type of
email) and the consumer's state evolves dynamically as a function of the firm's actions and her state (in an endogenous manner).  For instance, a
consumer who has interacted with a previous intervention (e.g., opening an
email) might exhibit a different behavior 
(and hence, might be in a different ``state'') as compared to a consumer
who has ``avoided'' previous interventions.  
Adding more to the complexity, the firm does not know apriori the effect
of its state-specific interventions on the consumer's behavior and hence,
needs to \textit{learn} or \textit{estimate} as it interacts with various
consumers.    
This maps to the  setting where the firm is promoting a new product, or an existing product to an unexplored consumer segment such as
a new geography (covariate shift), or perhaps there has been a change in the underlying consumer behavior possibly due to shifts in macroeconomic conditions, competitors’ policies, or other marketing activities of the
firm itself (concept shift). 

Note that the problem of conversion funnel optimization
is widespread in practice. For instance,
87\% of marketers develop interventions specific to the consumer state in
her journey to conversion and the effects of such interventions are
state-specific \citep{2020sej}. Generic interventions such as a
``how-to guide'' or a ``landing page'' perform better at the ``top of the
funnel'' whereas product-specific interventions such as ``product
overview'' or a ``customer review'' perform better towards the ``bottom of
the funnel'' \citep{2020sej}. 
There are multiple firms operating in the space of email campaign
management\footnote{See \url{https://www.privy.com/},
  \url{https://www.klaviyo.com/}, \url{https://mailchimp.com/},
  \url{https://www.constantcontact.com/}, and \url{https://sendgrid.com/} for a sample of firms operating in this
  space.} and the global market size for just email marketing is estimated
to be around USD 7.5 billion, with a projected growth to USD 17.9 billion
by 2027 \citep{2021ReportLinker}.  
To define the focus of our work, we highlight an important characteristic
of the email marketing industry. In particular, when a
consumer journey begins (for instance, a consumer providing her email to
Netflix), the firm has very limited data on her. All the firm has is the consumer's email and perhaps her 
name. Hence, we do not assume the firm has access
to the \textit{features} of the consumers (age, sex, location, for
example) but we let the firm learn about the consumers' preferences as it
interacts with them via sequential interventions.  
This is not only consistent with our personal experience (Figure
\ref{fig:CFONetflix}) but also with the discussions we had with our
industry partner\footnote{Our partner is one of the biggest service providers in the space of digital marketing.} that provided us the real-world data.
Further, a quick review of practice-driven blog articles suggests this is in fact consistent with practice.\footnote{See for example \url{https://www.markettailor.io/blog/role-of-analytics-in-email-marketing}.}

Optimizing the conversion funnel is challenging, primarily due to the complex consumer
behavior. The consumer behavior can be affected by the earlier
interventions (carryover and spillover effects) and the consumer's
interactions with such interventions. For example, a consumer who
interacts with an initial intervention (opening a promotional email) might
be more likely to convert in the future than a consumer who does
not. These effects can be non-linear \citep{chatterjee2003modeling}. There
can also be temporal effects \citep{sahni2015effect}, including marketing
fatigue \citep{sinha2007over}. For example, sending
promotional emails to a consumer every day might annoy her, leading her to
leave the system (by unsubscribing from the firm's email list).  
The key challenge here is that consumer behavior can be a complicated function of her state (past interactions, time since last intervention,
etc.), and apriori, 
the consumer response to interventions as a function of her state and the intervention is unknown and must be learned.
Doing so while optimizing the funnel at the same time is
challenging. 

\subsection{Our Approach and Contributions} \label{sec:approach}

We take a data-driven approach that maps
observable consumer-level data to
scalable and interpretable decisions.  We summarize our approach
in three modules: data, model, and decisions. 

\bold{Data.} We motivate our problem via a practical application
in \S\ref{sec:data}. We present a real-world large-scale dataset and
analyze it to explain consumer-level behavior.  Our dataset consists of 
sequential email interventions 
advertising a software product of a Fortune 500
firm to millions of consumers.  The firm uses four types of
emails. For each consumer, we observe her ``path'', i.e., her interactions
with the sequence of emails she receives (e.g.,  which emails she
opens/clicks) and whether she converts (purchases the product) or not.  
As discussed above, such consumer behavior is complicated by various
carryover and spillover effects.  To capture such effects, we consider
four dimensions of the data: (1) temporal (time of email), (2) consumer
awareness (number of type-specific emails received by a consumer), (3)
consumer engagement (number of type-specific emails opened/clicked by
a consumer), and (4) email type.  We use these dimensions to predict a
consumer's behavior, i.e.,  how will she behave to the next email she
receives as a function of these features? Since the interaction effects
between features can be non-linear, we give ourselves the flexibility to
capture them by fitting boosted trees. Our key finding is that though
intricate,  consumer behavior can be predicted by such micro-level data to
very high accuracy (out-of-sample AUC of over 0.95), which strongly
motivates the need to use such information when modeling consumer
behavior. 

\bold{Model.} In~\S\ref{sec:model}, we propose a Markov decision
process (MDP) for consumer behavior (\emph{conversion funnel MDP}). The states of this MDP are
constructed using the features identified in our data analysis.  As
these features predict consumer behavior very accurately, we expect the
MDP to correctly model the consumer journey.
In the MDP model,  at each point in time,
a consumer is in some observable state, which summarizes her interaction
history with the firm.
The firm decides on which intervention to perform (if any). 
The consumer transitions to a
new state that is a stochastic function of her
current state and the firm's intervention. The firm then performs
another intervention and the process repeats until the consumer converts
or quits.  This MDP only has a terminal reward, a feature we exploit in
designing our decision-making algorithm. 

\bold{Decisions.} 
Although our conversion funnel MDP is able to accurately predict
consumer-level behavior,  it is very high dimensional as 
we track quantities such as the number of type-specific emails
received/opened/clicked by a consumer in the MDP state. 
Thus, identifying the optimal policy is
challenging, and motivates the need for the
decision-making algorithm to be \emph{scalable} and \emph{interpretable}.
Further, the number of parameters needed to fully specify the 
conversion funnel MDP is of the order of hundreds of thousands. Thus, it is
impractical (especially for a new product or
under concept/covariate shift) to obtain a good estimate for the
parameters \emph{before} intervening, i.e., one is 
forced to \emph{learn} in an online manner.
Our key contribution is to exploit the terminal reward structure of the conversion funnel MDP and propose a novel attribution-based learning algorithm (\S\ref{sec:MFABL}), which guides the firm in terms of what intervention to perform given the state of a consumer.  Our algorithm, \emph{model-free\footnote{\label{ft:CFOModelFreeOne}We use the terminology
``model-free'' to be consistent with the existing literature 
\citep{sutton2018reinforcement} and note that ``model-free'' does not
mean that there does not exist a true underlying model. Instead, it
means that there exists a true underlying model but our algorithm does
not learn/estimate the model. Approaches that learn/estimate the model
are called ``model-based''. We revisit this discussion in
\S\ref{sec:MFABL} (Footnote~\ref{ft:CFOmodelfree}), where we
formally define our algorithm.} approximate Bayesian learning} (MFABL),
extends the simplicity 
of Thompson
sampling~(TS) algorithm~\citep{thompson1933likelihood}, 
proposed for multi-armed bandits,
to the conversion funnel MDP.  Similar to TS,  MFABL
maintains a Beta belief on the value of
each intervention, but in a state-specific manner.  However, given the sequential nature
of our model,  updating the Beta belief is challenging and 
hence, MFABL approximates the Bayes update by mimicking the Beta-Bernoulli style attribution of TS.  This lends MFABL scalability and interpretability, and despite being an approximate Bayes update,  we prove that MFABL is asymptotically optimal (Theorem \ref{theorem:CFOConvergence}).  We also analyze its convergence rate (Theorem \ref{theorem:ConRate}) and supplement our theoretical developments via extensive numerics
in \S\ref{sec:numerics} and \S\ref{sec:scalability}, where we benchmark
MFABL with multiple approaches and highlight the value of scalability.  

\subsection{Related Literature} \label{sec:literature}
Our work relates to various works in advertising (sequential marketing, conversion funnel modeling, and attribution) and the literature on reinforcement learning. We briefly review them next.

\bold{Sequential Marketing.} Maximizing a firm's expected reward by showing a sequence of ads has been studied for over two decades.
Works here can broadly be classified into two categories: (1) estimation-based and (2) learning-based.  Estimation-based approaches use historical data to first estimate the consumer behavior model and then use the estimated model to design an optimal marketing policy.  Some examples are \cite{bitran1996mailing} and \cite{simester2006dynamic}.
However, when data is not available apriori (e.g., promoting a new product) or when the underlying consumer behavior changes (concept shift) or when targeting a new consumer segment (covariate shift), such approaches might be inappropriate \citep{simester2020targeting}.
This motivates learning-based approaches, which continuously update the estimate/belief over the consumer behavior model as more data is collected. Examples include \citet{abe2002empirical},  \cite{rafieian2022optimizing}, and \cite{liu2022dynamic}.  None of these works focus on the conversion funnel MDP, which is our primary focus. In particular, we explicitly leverage the terminal reward structure of the conversion funnel and show how it can be used to design an interpretable and scalable learning algorithm.

\bold{Conversion Funnel Modeling.} There is a large literature on conversion funnel modeling in marketing, going back to \cite{strong1925psychology}.  Such models capture the
journey of the consumer via small-sized state-based models, where the state is typically hidden.
The states capture the consumer journey from 
being unaware of the product to
becoming interested, and finally purchasing.  
On the contrary, we leverage consumer-level granular data 
and work with a large state space with thousands of states.
Though such high-dimensional modeling  
allows us to explain consumer behavior with very high accuracy,  
it makes the conversion funnel optimization problem much more challenging,
for which we develop an interpretable and a scalable algorithm 
that exploits the terminal reward structure of the conversion funnel.

\bold{Attribution in Online Advertising.} With the growth of online advertising,  understanding the contribution of various ad actions to a product purchase (conversion) has become an important topic.  Works such as \cite{singal2022shapley} discuss various attribution strategies (e.g., last-touch, uniform,  removal effect, and Shapley value).  Their focus is primarily on \emph{measuring} the contributions but not on \emph{prescriptions}, which differentiates our work since we provide a systematic way to map attribution to marketing decisions. We explicitly decompose our decision-making algorithm into two modules: (1) optimization and (2) attribution. The optimization module picks a marketing action based on current beliefs and the attribution module updates the beliefs by crediting a conversion event to the actions that led to it, which then feeds into the optimization (and so on).
We propose two simple attribution rules.  Under the first rule, an action is credited positive/negative attribution if it transitions the consumer to a ``good''/``bad'' state.
The second rule is analogous to the widely used uniform attribution as it uniformly attributes a positive (or negative) credit to all the actions that led to a conversion (or lack thereof).
We note that works such as \cite{berman2018beyond} also map attribution to decisions, but their focus is on stylized models rather than data-driven optimization.

\bold{Reinforcement Learning (RL).} Our model of consumer behavior is an
MDP and we let the firm learn the parameters of the MDP by interacting
with consumers in an online manner.  
Therefore,  our problem is an instance
of RL. We refer the reader to \cite{sutton2018reinforcement} for an overview
of RL.  The key distinction in our work is that the conversion funnel MDP
exhibits a special structure such that there is no immediate reward but
only a terminal reward (consumer converting).  
Hence,  our MDP is in a middle ground between bandits
and general MDPs.  
Like bandits, there is only a terminal reward but
we allow for sequential interventions and the evolution
of consumer state.  We exploit this 
specific structure to
design a novel decision-making algorithm that extends the simplicity and
scalability of Thompson sampling for
bandits~\citep{thompson1933likelihood} to the conversion funnel MDP in a model-free manner.  
Though \citet{dearden1998bayesian} and
\citet{ryzhov2019bayesian} also maintain a \textit{model-free} belief, 
the proposed methods cannot
exploit the terminal reward structure inherent in the conversion funnel and
hence,  the update is not interpretable when compared to 
the simple interpretable ``Beta-Bernoulli'' update used in MFABL. Further,
their action selection can be quite different from that of MFABL. For
example, \citet{ryzhov2019bayesian} use a knowledge-gradient technique for
action selection, which in fact, requires one to estimate the transition
probabilities $\mathcal{P}$, making the overall algorithm
model-based. Recent work by \citet{jin2018q} attempt
at solving similar problems in a model-free manner, but by using upper
confidence bound (UCB) algorithms as opposed to TS.  
We also briefly mention an alternative line of work
\citep{strens2000bayesian, osband2013more,ouyang2017learning} that attempts at solving similar problems but by maintaining a
\textit{model-based} belief over the underlying MDP. However, such
model-based algorithms require one to
compute the optimal policy for an
MDP in each
``iteration'', which poses a significantly higher computational burden than the
model-free approaches, especially when the number of states $S$ or actions $A$ is
large. Furthermore, since model-based approaches store belief over the
transition probabilities $\mathcal{P}$, their storage requirement is
$\mathcal{O}(S^2 A)$, compared with $\mathcal{O}(SA)$ of MFABL. In our
numerical study (\S\ref{sec:numerics}), we benchmark MFABL with such
state-of-the-art algorithms.

Though our framework is general, our numerics focus on an email
marketing application,  an area explored by others as well. For example,
\cite{kadiyala2021data} propose a data-driven approach to targeting
promotional emails while capturing delayed / long-term
incentives. However,  in contrast to our sequential learning framework,
they focus on a one-step estimate-then-optimize approach. We refer the
reader to \cite{kadiyala2021data} for a more detailed review of related
work. 

\bold{Outline.}
We motivate our problem via a practical application in \S\ref{sec:data}, where we present a dataset and analyze it to explain consumer-level behavior.  Our data analysis motivates the conversion funnel MDP model of consumer behavior, which we use to formulate the conversion funnel optimization problem in \S\ref{sec:model}.  In \S\ref{sec:MFABL},  we propose the MFABL algorithm for solving the conversion funnel optimization, followed by numerical experiments in \S\ref{sec:numerics} and \S\ref{sec:scalability}.
We conclude in \S\ref{sec:conc}. 

\section{Motivating Application} \label{sec:data}
We now present a motivating application by discussing a real-world
dataset.  After describing the dataset in the paragraph below,  we 
perform an exploratory data analysis (\S\ref{sec:eda})
and then,  provide a model for
consumer-level behavior (\S\ref{sec:behavior}). We
hope to convince the reader that though consumer behavior is complicated,
it can be captured via micro-level data. 

Our data is for a software product promoted and sold on the
Internet by a Fortune 500 firm.  Microscopic data
statistics are not disclosed for the sake of anonymity. 
The dataset consists of a few million consumer paths, out of which a few
tens of thousands of consumers made a purchase. Each path corresponds to a
unique consumer and starts with the consumer creating an account on the
firm's website (\emph{sign-up}). Once a consumer signs up, the firm sends
her a sequence of emails on a daily basis, with possibly days in a path
with no email.  
The firm classifies these emails into four types: (1)
``awareness'', (2) ``promotional'', (3) ``return-on-investment (ROI)'', and (4) ``type 4''\footnote{The actual name of ``type 4'' email is not disclosed for anonymity
  reasons since it reveals the name of the product being promoted and
  hence, risks revealing the identity of the firm.}.  The ``awareness'' and
``promotional'' emails are designed with the goal of promoting the firm brand
and product awareness (``top of the funnel'') whereas the ``ROI'' emails
are designed for conversions (``bottom of the funnel''). The ``type 4'' email
highlights the features of the product being promoted. 
A consumer can receive the same type of email multiple times.
For each email, the firm knows the time it was sent and the time when the
consumer opened it (if at all). In addition, for ``awareness'' and
``ROI'' emails,  the
firm can track the time at 
which the consumer clicks on an embedded link in the email (if at
all). From the date of the sign-up, the data tracks a consumer for several
months to give her enough time to make a decision. 
Consumers sign up on different dates and the path lengths can vary. Some
convert within a few days and some never do. We focus on a horizon of
$T=14$ days since a 
large fraction of conversions (approximately 80\%) happen within
two weeks. This can be thought of as the ``lifetime'' of a user (time taken by
a user to make a decision). We note that the product being sold is
relatively inexpensive (as compared to a product such as a car). 

\subsection{Exploratory Data Analysis} \label{sec:eda}
\begin{figure}
\centering
\begin{subfigure}[b]{0.3\textwidth}
\centering
\includegraphics[width=1\linewidth]{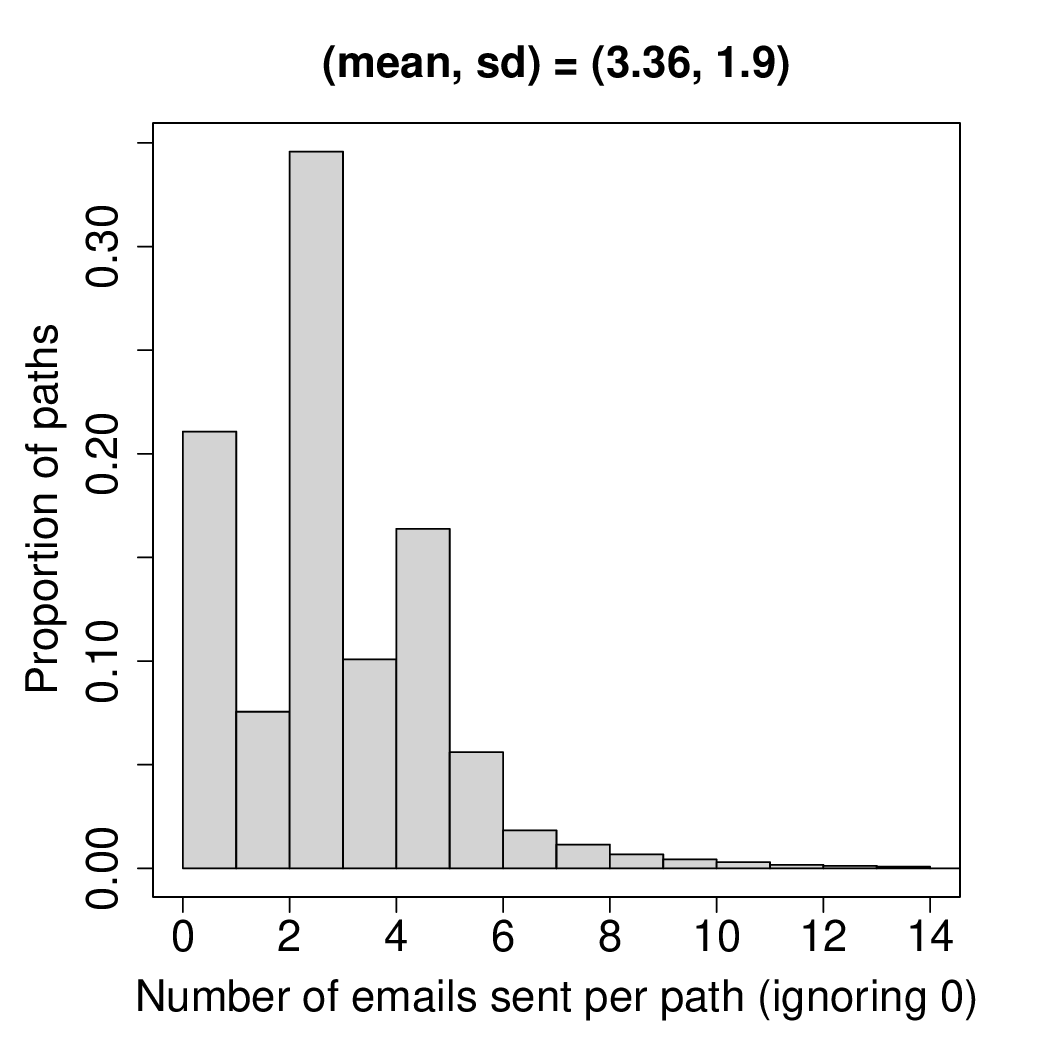}
\caption{\small Received}
\label{fig:nEmailsSent}
\end{subfigure}
\begin{subfigure}[b]{0.3\textwidth}
\centering
\includegraphics[width=1\linewidth]{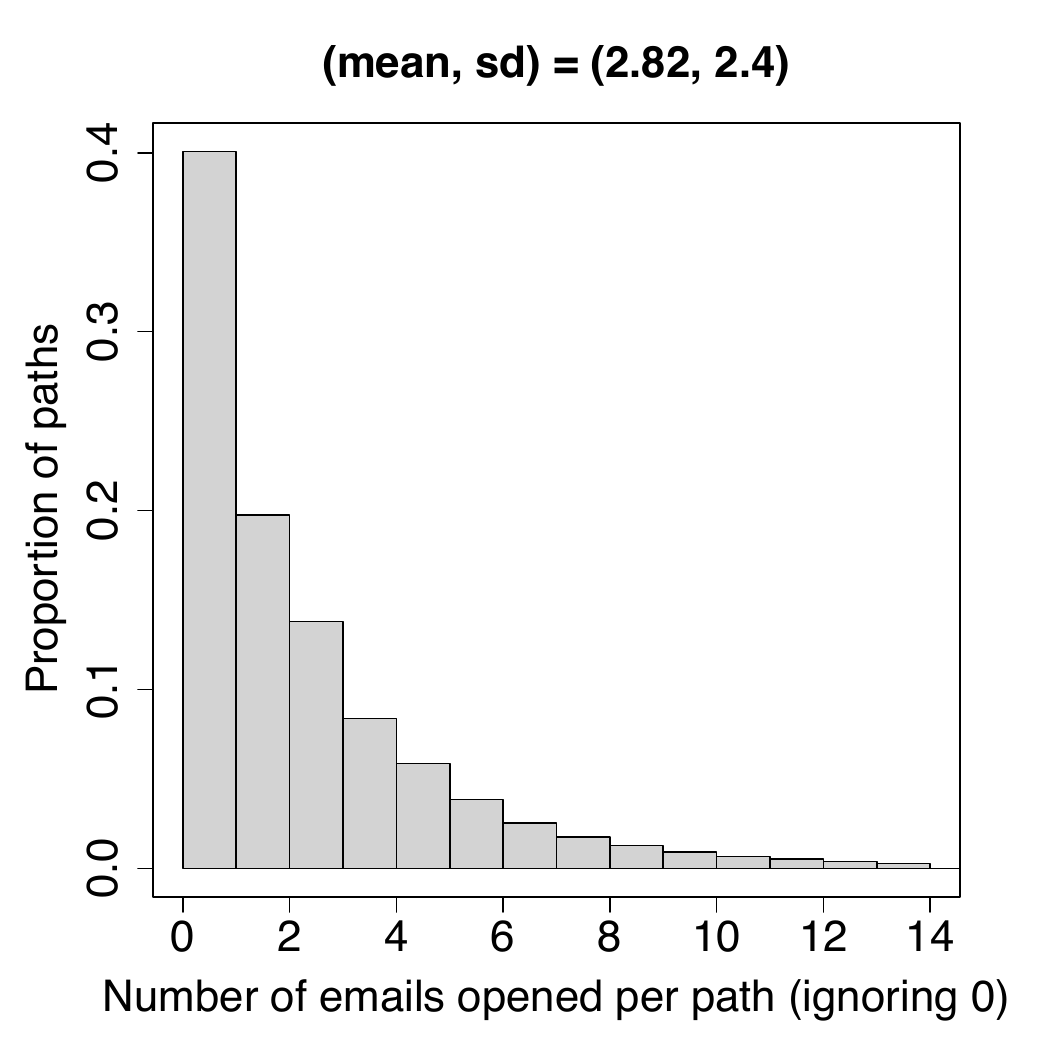}
\caption{\small Opened}
\label{fig:nEmailsOpened}
\end{subfigure}
\begin{subfigure}[b]{0.3\textwidth}
\centering
\includegraphics[width=1\linewidth]{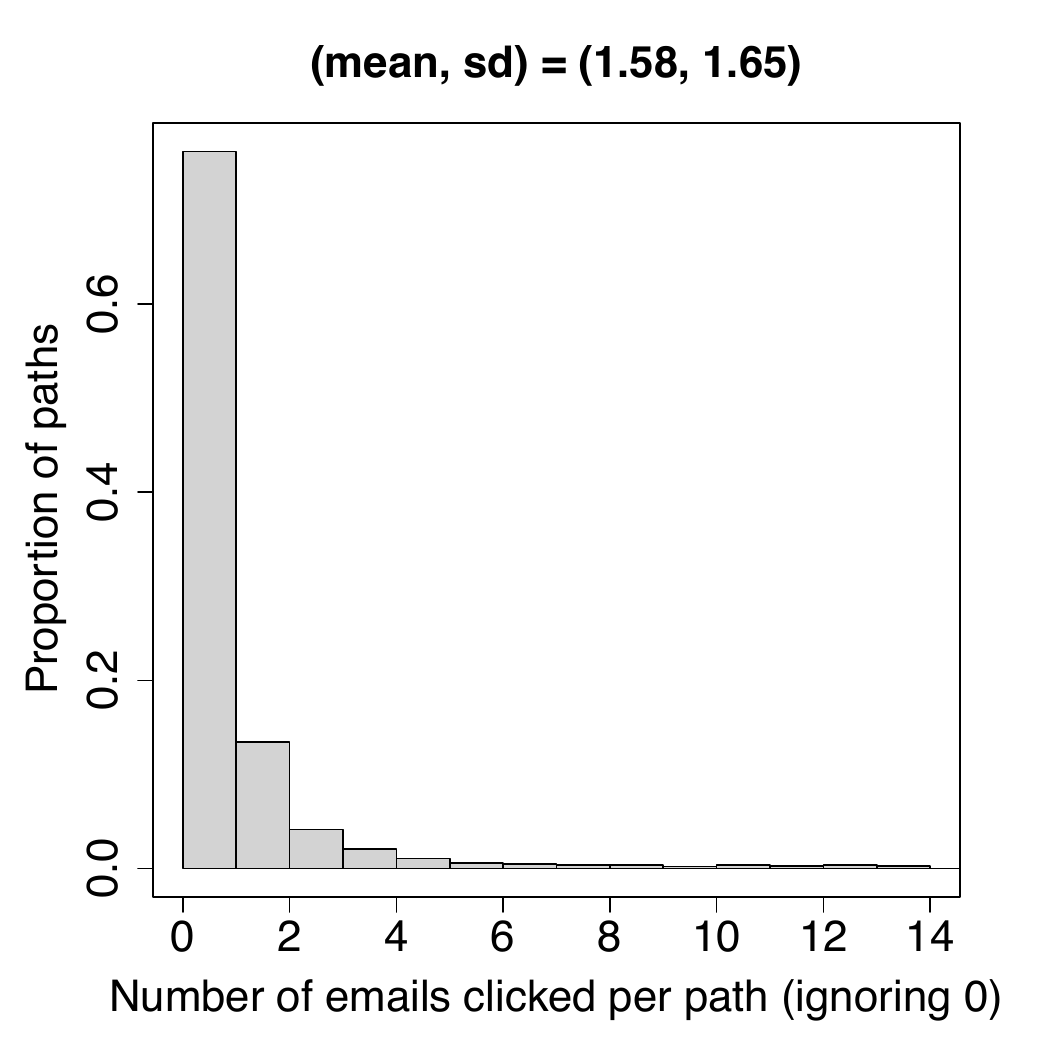}
\caption{\small Clicked}
\label{fig:nEmailsClicked}
\end{subfigure}
\caption{\normalfont Distribution of number of emails received, opened, and clicked. For each distribution, we ignore the value of 0 to understand paths with some activity.  Hence, the first bucket (x-axis) corresponds to a value of 1. Given our time horizon of $T=14$ days and a daily frequency of emails, the maximum value on the x-axis is 14.} 
\label{fig:nEmails}
\end{figure}

\begin{figure}
\centering
\begin{subfigure}[b]{0.3\textwidth}
\centering
\includegraphics[width=1\linewidth]{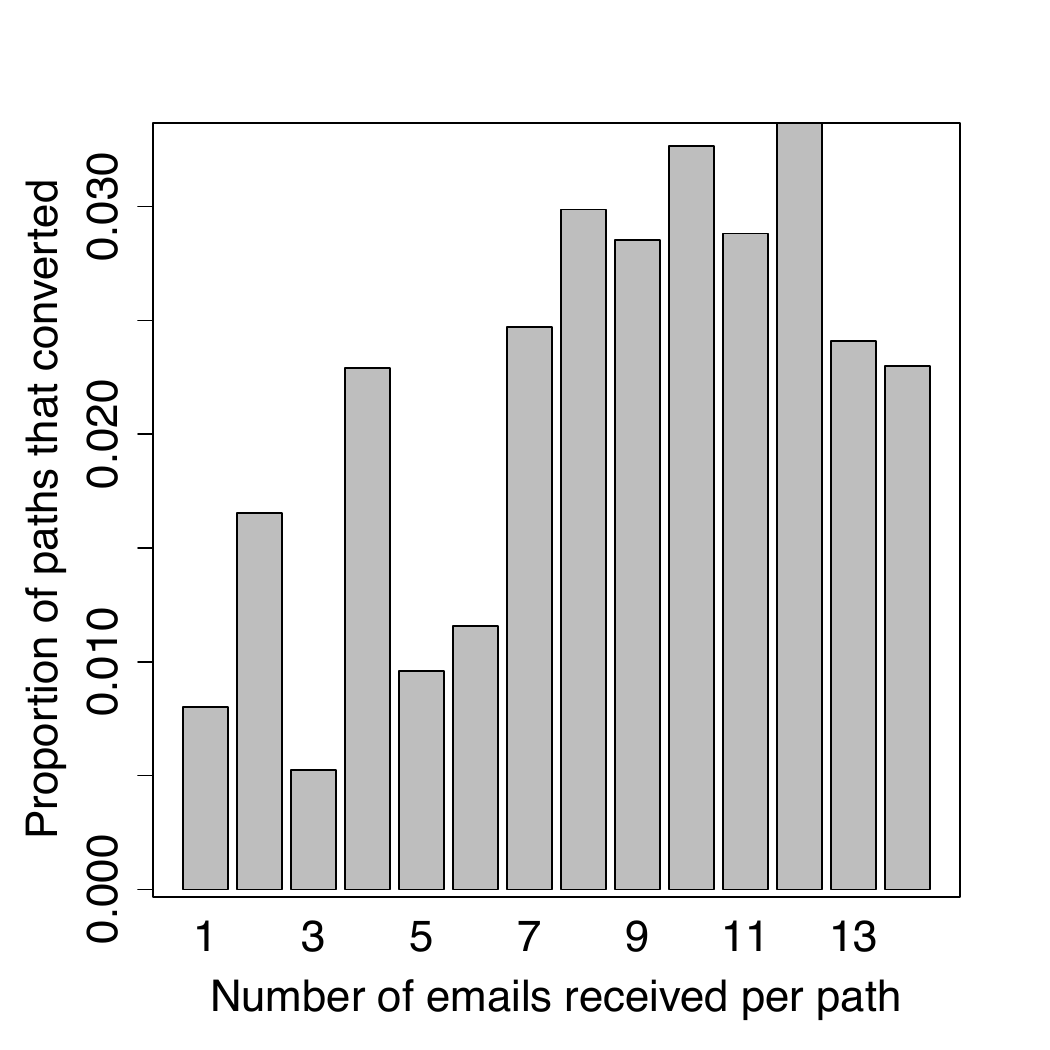}
\caption{\small Received}
\label{fig:ConversionVsSent}
\end{subfigure}
\begin{subfigure}[b]{0.3\textwidth}
\centering
\includegraphics[width=1\linewidth]{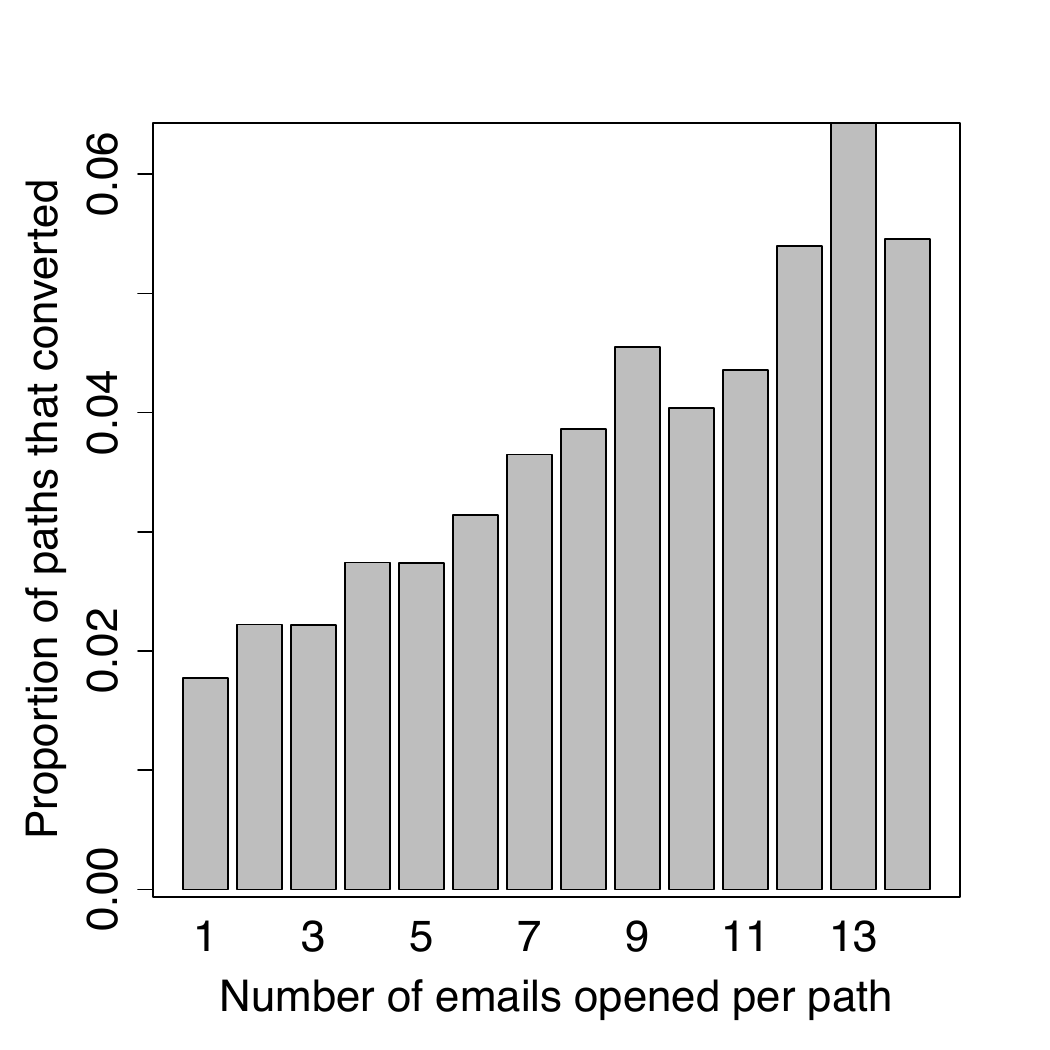}
\caption{\small Opened}
\label{fig:ConversionVsOpen}
\end{subfigure}
\begin{subfigure}[b]{0.3\textwidth}
\centering
\includegraphics[width=1\linewidth]{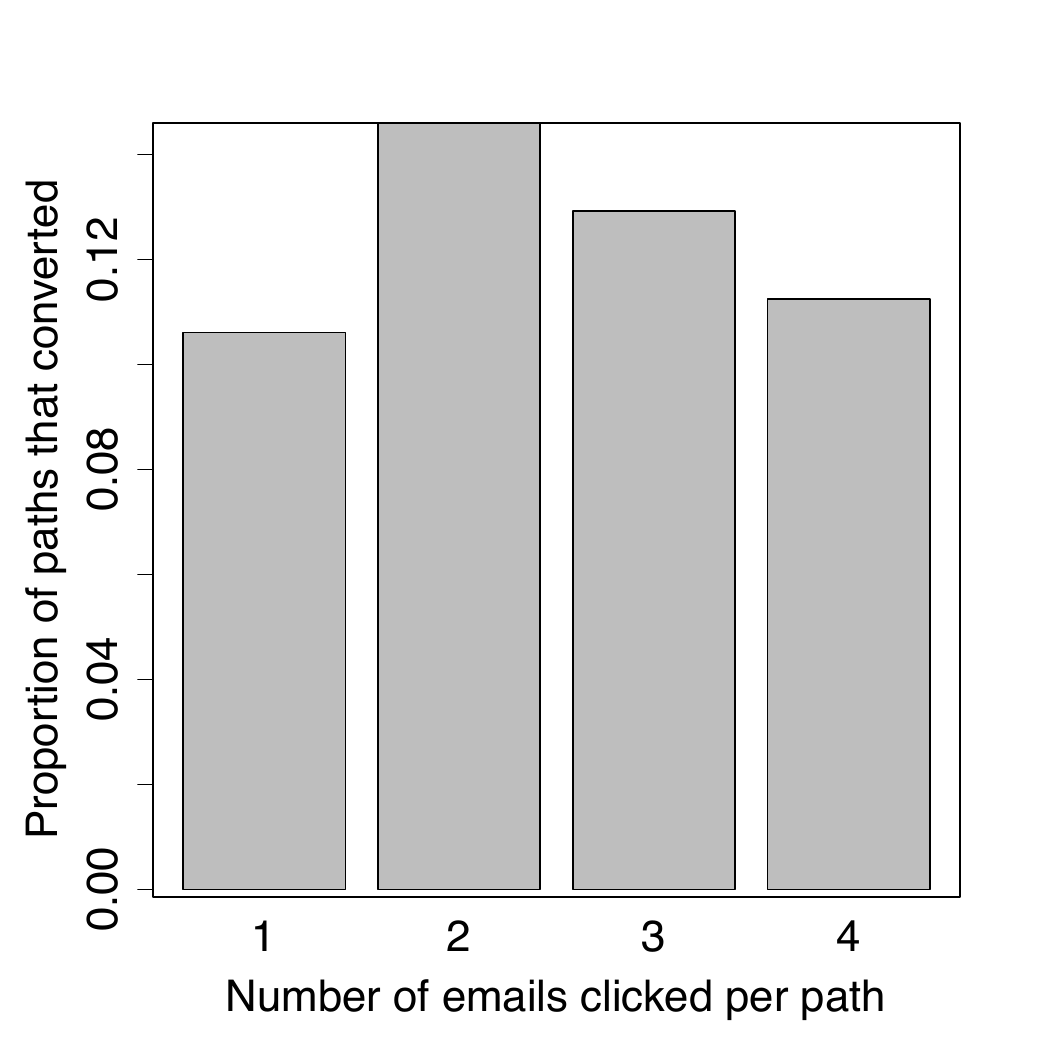}
\caption{\small Clicked}
\label{fig:ConversionVsClick}
\end{subfigure}
\caption{\normalfont Empirical proportion of paths that converted as a
  function of consumer's level of interaction.  These
    exploratory  plots  
    simply summarize the observed data.  For
    example, out of the paths that received exactly 3 emails,
    approximately $0.005$ fraction converted.  Note that ``received'' tag
    only counts the number of emails received --  
    some of these might have been opened or clicked.}
\label{fig:ConversionVsInteraction}
\end{figure}

Figure \ref{fig:nEmails} shows some summary statistics of emails received,
opened, and clicked per consumer path.  
The decaying curves in Figures~\ref{fig:nEmailsOpened} and~\ref{fig:nEmailsClicked} are not
surprising,  and consistent with the 
traditional conversion funnel.
We expect conversion rates to vary as a function of a consumer's
interaction level. We display this in Figure
\ref{fig:ConversionVsInteraction}.  The y-axis scale increases from 0.03
(Figure \ref{fig:ConversionVsSent}) to 0.06 (Figure
\ref{fig:ConversionVsOpen}) to 0.12 (Figure \ref{fig:ConversionVsClick}),
which suggests a conversion is more likely as a consumer is more engaged.
Of course, as mentioned in \S\ref{sec:intro},  a key challenge here is
that there can be non-trivial carryover, spillover, and temporal effects
and characterizing them is non-trivial.  With this in mind, we proceed to understand consumer-level behavior. 

\subsection{Understanding Consumer-Level Behavior} \label{sec:behavior}
Given the daily email frequency,  we index time in the number of days
$t \in \{1,2,\ldots,T\}$ with $T=14$ days as
above.  Day 1 ($t=1$) corresponds to the consumer signing up.  The firm
possibly sends an email to the consumer on day 1. If so,  the consumer can
exhibit any of the following four mutually exclusive behaviors:  (1)
ignore the email and not convert (\emph{ignore}), (2) open the email but
not click on any link in it and not convert (\emph{open}),  (3) open the
email and click on a link but not convert (\emph{click}),  and (4)
convert. If she converts, her path ends. Else, the process repeats on day
2, with the end of horizon being $T=14$. Our goal here is to be able to
explain/predict consumer behavior, i.e., given the historical
interactions by a consumer,  how will she behave if she is sent an email
of type $a \in \{0,1,2,3,4\}$, where 0 denotes no email? 

To map our setup to standard machine learning terminology, denote by $\fx_{nt}$ the
data we have on consumer~$n$ at the start of day $t$ (e.g., her
interaction history from the previous $t-1$ days) and by $y_{nt} \in
\{\text{ignore, open, click, convert}\}$ the consumer $n$ behavior
following the email on day $t$.  Our goal is to predict $y_{nt}$ as a
function of $\fx_{nt}$ for $n \in \{1,\ldots,N\}$ and $t \in
\{1,\ldots,T\}$, where $N$ equals a few million consumer paths.  We have
two choices to make here: (a) the construction of features $\fx_{nt}$ and
(b) the prediction function that maps $\fx_{nt}$ to a prediction.  
For features, in order to capture various temporal,  carryover, and spillover effects, 
we experimented with the following four dimensions: 
\begin{enumerate}
\item \bold{Temporal.} Number of days $t$ since sign-up.
\item \bold{Awareness.} Number of type-specific emails received by consumer $n$ before day $t$:
$\fx_{nt}^{\text{received}} := [x_{nta}^{\text{received}}]_{a=1}^4.$
\item \bold{Engagement.} Number of type-specific emails opened and clicked by consumer $n$ before day $t$: $\fx_{nt}^{\text{opened}} := [x_{nta}^{\text{opened}}]_{a=1}^4$ and $\fx_{nt}^{\text{clicked}} := [x_{nta}^{\text{clicked}}]_{a \in \{1,3\}}.$
\item \bold{Email type.} Email type $a_{nt}$ sent to consumer $n$ on day $t$.   
\end{enumerate}
The first dimension captures temporal effects,  and the second and third
dimensions capture the level of consumer awareness and engagement,
respectively (motivated by Figure~\ref{fig:ConversionVsInteraction}). 
For dimensions~2 and 3, we track type-specific counts\footnote{Recall that click information is available for ``awareness'' and ``ROI'' emails only.}, and hence, allow for
interaction effects across email types, and across time (dimension 1). The
fourth dimension allows us to tailor our prediction to the type of email
that is sent, and its potential interactions with historical data captured
in the first three dimensions. 
We restrict the feature $x_{nta}^{\text{received}}$  that tracks the
number of emails  received 
to 
$\{0,1,2+\}$,  denoting whether the consumer has received $0$, $1$, or more
than $1$ emails.  
Similarly,  we restrict the feature $x_{nta}^{\text{opened}}$ that tracks
the number of emails opened
and clicked 
to
$\{0,1+\}$.   
This bucketing proves helpful in controlling the size of the MDP state
space (\S\ref{sec:model}), and as we show, it suffices to explain consumer
behavior to high accuracy. 
For the prediction function, we follow the prior
literature~\citep{yoganarasimhan2020design,  kadiyala2021data} and
give ourselves the flexibility to capture possibly non-linear interactions
between the features by fitting boosted trees via the \texttt{xgboost}
package \citep{Chen2016}. We discuss the implementation details in
\S\ref{sec:xgboostdetails}. 

We experimented with $12$ different models, depending on the input features
used, as shown in the first two columns of
Table~\ref{table:predictionmodels}. The first $6$ models do not include
email type 
as a feature whereas the last $6$ do. For each model, we fit a multiclass
boosted tree (since the output variable can take on $4$ values) via
\texttt{xgboost}.  
We use 5-fold cross-validation for parameter tuning (details in
\S\ref{sec:xgboostdetails}) and report two out-of-sample\footnote{\label{ft:OutOfSample}Here, out-of-sample refers to the data corresponding to the ``test'' fold in cross-validation. } evaluation
metrics in the last two columns of Table \ref{table:predictionmodels}: (1)
multiclass area under the curve (AUC) and (2) multiclass log-loss (``mlogloss'' in \texttt{xgboost}).  AUC always lies between 0 and 1, with a higher value indicating a better performance, whereas a lower value of log-loss is desirable.  We elaborate on both the metrics in \S\ref{sec:xgboostdetails}.

\begin{table}
\small
\centering
\begin{tabular}{c | c | c | c}
 &  & \multicolumn{2}{c}{\bold{Out-of-sample performance}} \\
\bold{Model} & \bold{Input features} & \bold{AUC} & \bold{Log-loss}  \\
\hline 
1 & Temporal & 0.66 (0.00) & 0.0589 (0.0004) \\
2 & Awareness & 0.79 (0.00) & 0.0542 (0.0003) \\
3 & Engagement & 0.70 (0.00) & 0.0566 (0.0003) \\
4 & Temporal + awareness & 0.86 (0.00) & 0.0503 (0.0004) \\
5 & Temporal + engagement & 0.80 (0.00) & 0.0499 (0.0003) \\
6 & Temporal + awareness + engagement & 0.88 (0.00) & 0.0465 (0.0003) \\
\hline 
7 & Temporal + email type & 0.85 (0.00) & 0.0476 (0.0004) \\
8 & Awareness + email type & 0.93 (0.00) & 0.0422 (0.0004) \\
9 & Engagement + email type & 0.88 (0.00) & 0.0433 (0.0003) \\
10 & Temporal + awareness + email type & 0.94 (0.00) & 0.0417 (0.0004) \\
11 & Temporal + engagement + email type & 0.90 (0.00) & 0.0414 (0.0003) \\
12 & Temporal + awareness + engagement + email type & 0.96 (0.00) & 0.0376 (0.0003) 
\end{tabular}
\caption{\normalfont Summary of the 12 models for understanding
  consumer behavior. For each model, we list
  the input features and report
  the out-of-sample average (and standard deviation) of the evaluation
  metrics (AUC and log-loss) over 5 folds of cross-validation.  Note that
  the in-sample scores were similar to out-of-sample scores, suggesting no
  overfitting. For example, the in-sample AUC and log-loss for model $12$
  were 0.96 and 0.0372, respectively. 
  } 
\label{table:predictionmodels}
\end{table}

Our key finding is we can explain consumer behavior with extremely
  high accuracy (e.g., an out-of-sample AUC of over 0.95!) if we capture
  micro-level information as model~$12$ does. This is quite remarkable and
  possible because we encode very rich information (temporal,
  type-specific awareness and engagement, and email type) in our features
  and allow for non-linear interactions between them (via boosted
  trees). Further, there is a clear value in using such features as the
  out-of-sample evaluation metrics improve with more information.
  For example, the AUC increases from 0.66 to 0.80 to 0.88 to 0.96 when 
  we sequentially add temporal (model 1), engagement (model 5), 
  awareness (model 6), and email type (model 12) information.
  We note that this finding is not peculiar to our dataset as others have reported similar
  results (see \cite{rafieian2022optimizing} for example).  
  Of course, as in \cite{rafieian2022optimizing}, one can capture more
  features such as time since the last email, and possibly improve the model.  
  All of this makes a strong case to account for such micro-level
  information when modeling consumer behavior and as such, the consumer
  behavior model we present next is general enough to capture an arbitrary
  construction of input features. 

\section{Consumer Behavior Model and Conversion Funnel Optimization} \label{sec:model}
We now formally define the problem.  First, we discuss our model for consumer behavior (\S\ref{sec:consumermodel})
and then, formulate the conversion funnel optimization problem (\S\ref{sec:optimization}). 
Note that we leverage the model of \cite{singal2022shapley}, who used such a model for attribution (rather than optimization). 

\subsection{Model for Consumer Behavior: The Conversion Funnel MDP} \label{sec:consumermodel}

Motivated by our data analysis in \S\ref{sec:data}, 
we model the consumer behavior using a state-based model, i.e.,
at each point in time,  the consumer is in a state (possibly a function
of her history). The firm observes\footnote{In the traditional conversion
  funnel models, the state is latent or hidden.  However, given 
  the ability to collect very granular consumer data, 
  and our ability to 
  fairly accurately predict
  consumer behavior using
  such data (as shown in \S\ref{sec:data}), we assume the state 
  is given by observed consumer features.
  We acknowledge that 
  observed consumer features may not completely define
  the underlying true latent state of the 
  consumer; 
  however, approximating the 
  true state by 
  the observable features greatly simplifies 
  optimization/learning problem since learning partially observable MDPs
  is much more challenging. We anticipate that the richness of the
  observed consumer features will be adequate for identifying good
  decision, and this borne out in our numerical experiments.}
  the state of the consumer and performs an intervention.  
As a result, the consumer transitions to a new state that is  a stochastic
function of the firm's interventions and her current state.  The process
ends when the consumer converts or quits. If the consumer converts, the
firm earns a reward. 
As our model is an MDP \citep{bertsekas1995dynamic},
we define the five underlying components of the MDP next: (1) state space
$\bS$, (2) action space $\bA$, (3) transition probabilities $\cP$, (4)
initial state probabilities $\blambda$, and (5) reward $r$. We refer to
this MDP as the \textit{conversion funnel} and denote it by $\cM \equiv
(\bS, \bA, \cP, \blambda, r)$. 

\bold{State Space.} We define $\bS := \{1, \ldots, S\}$ as the set
of states of an ``active'' consumer, 
i.e., a consumer who has not converted or quit. In addition, there are two
aborbing states $\{q, c\}$ with $\bS^+ := \bS \cup \{q,
c\}$. State $c$ refers to conversion (consumer buys the product) and $q$
to the quit state (consumer leaves the system).   
At each point of time, a consumer is in one of the states and the firm observes this information. 
For the application discussed in~\S\ref{sec:data},  
one definition for the state \(s_{nt}\) of consumer $n$ at the beginning
of day $t$ is $s_{nt} = (t,\fx_{nt}^{\text{received}}, \fx_{nt}^{\text{opened}}, \fx_{nt}^{\text{clicked}})$.
The state does not include the email type $a_{nt}$ that will be shown to
her on day $t$, which forms the action, as we discuss next.  

\bold{Action Space.} The set of interventions available to the firm is
defined as $\bA := \{0, 1, \ldots, A\}$ with $0$ denoting no
intervention.  We can allow the action space to depend on the state, i.e., $\bA_s$ for all $s \in \bS$, but to keep the notation simple, we use $\bA$ and note that all results hold for state-specific action space too.  We assume that the  firm takes actions 
at discrete time points $[t_k]_{k \geq 1}$ where we 
allow the gap $\delta_k := t_{k+1} - t_k$ to be unequal.
For the setting in \S\ref{sec:data},  the actions
correspond to the types of email (including no email), and the 
discrete time points $[t_k]_{k \geq 1}$ correspond to days.

\bold{Transition Probabilities.} If a consumer is in state $s \in \bS$ and
the firm takes action $a \in \bA$, the consumer transitions to state $s'
\in \bS^+$ w.p.\ $p_{sas'} \in [0,1]$. Note that $s'$ can equal $s$
(self-loop). We allow for an arbitrary transition structure (as long as
Assumption~\ref{ass:CFOAbsorption} stated below holds). Trivially,
$\sum_{s' \in \bS^+} p_{sas'} = 1 \ \forall (s,a) \in \bS \times \bA$ and
$p_{cac} = p_{qaq} = 1 \ \forall a \in \bA$ (absorbing states).  
Furthermore, $p_{sac}$ denotes the \textit{one-step conversion
  probability} corresponding to $(s,a) \in \bS \times \bA$. 
We denote by $\cP$ the collection of all transition
probabilities. 
In the setting discussed in \S\ref{sec:data},
consumer~$n$ in state $s_{nt} =
(t,\fx_{nt}^{\text{received}}, \fx_{nt}^{\text{opened}},
\fx_{nt}^{\text{clicked}})$ at the beginning of day $t$ transitions to
$s_{n,t+1} = (t+1,\fx_{n,t+1}^{\text{received}},
\fx_{n,t+1}^{\text{opened}}, \fx_{n,t+1}^{\text{clicked}})$ where
$(\fx_{n,t+1}^{\text{received}}, \fx_{n,t+1}^{\text{opened}},
\fx_{n,t+1}^{\text{clicked}})$ depend on the email $a_{nt}$ sent to her
on day $t$ and her interaction with it (ignore, open, click, or
convert). If she converts, then $s_{n,t+1} = c$. 
The only restriction we
impose is that every path \textit{eventually} terminates, i.e., either
converts or quits.   
In our \S\ref{sec:data} application,  every path terminates after 14 days
given the finite time horizon. 

\begin{assumption}[\bold{Absorption}] \label{ass:CFOAbsorption}
Under any policy, every consumer eventually converts or quits.
\end{assumption}

\bold{Initial State Probabilities.} We denote by $\lambda_s$ the initial
state probability corresponding to state $s \in \bS$, i.e., the
probability a consumer starts in state $s$. Trivially,
$\sum_{s \in \bS} \lambda_s = 1$. We define $\blambda := [\lambda_s]_{s
  \in \bS}$ and we allow the consumers to start in different initial
states. Given $\blambda$, consider the set of initial states
$\bS_{\blambda} := \{s \in \bS : \lambda_s > 0\}$. We assume without loss
of generality (wlog) that each state in $\bS$ is reachable (via possibly
multiple transitions) via \textit{some} state in $\bS_{\blambda}$ under
\textit{some} policy. This assumption is wlog since if there exists a
state $s \in \bS$ that violates this assumption, then we can discard that
state and re-define $\bS \leftarrow \bS \setminus \{s\}$ (as no consumer
will ever visit state $s$). We will refer to this assumption as
``connectedness'' and note that it is weaker than the notion of a Markov chain being ``irreducible''.
For the application in \S\ref{sec:data},  with $s_{nt} = (t,\fx_{nt}^{\text{received}},
\fx_{nt}^{\text{opened}}, \fx_{nt}^{\text{clicked}})$, we can set the
initial state to $s_{n1} = (1, \pmb{0}, \pmb{0},\pmb{0})$ for all $n$
since every path starts with a sign-up. 

\bold{Reward Structure.} The firm earns an immediate reward of $r \in [0, \infty)$ when a consumer converts, i.e., transitions from a state in $\bS$ to the conversion state $c$.  All other transitions result in an immediate reward of 0.
Since $r$ is a constant, we assume it to equal 1 wlog. We emphasize that such a \textit{terminal nature} of the reward is specific to our context and hence, our conversion funnel model belongs to a special class of MDPs. We will exploit this application-specific structure when developing our algorithm in \S\ref{sec:MFABL}.

Having defined our model, we emphasize it is general enough to capture an arbitrary construction of input features, which is powerful given such features explain consumer-level behavior with high accuracy.  
As such,  our model encodes a multitude of temporal, carryover, and spillover effects.
Before stating the optimization problem, we define some preliminaries,
which prove useful later on. All these preliminaries are part of a
standard MDP setup \citep{bertsekas1995dynamic}. 

\bold{Policy.}  A policy $\pi := [\pi_{sa}]_{(s,a) \in \bS \times \bA}$ is a mapping from states to action probabilities, i.e., $\pi_{sa}$ denotes the probability firm takes action $a \in \bA$ when a consumer is at state $s \in \bS$.

\bold{Value Function of a Policy.}  Given the firm employs policy ${\pi}$, state-value function $V^{{\pi}}_s$ denotes the expected reward the firm reaps from a consumer who is in state $s \in \bS$. 
With $r=1$ wlog, $V^{{\pi}}_s$ equals the \textit{eventual conversion probability} from state $s$ under policy $\pi$.
Using the Bellman equation and the terminal reward structure of the conversion funnel, we get
$
V^{{\pi}}_s = \sum_{a \in \bA} \pi_{sa} \left\{ \sum_{s' \in \bS} p_{sas'} V^{{\pi}}_{s'} \ + \   p_{sac} \right\} \ \forall s \in \bS.
$
Given $\pi$, define the action-value function 
\begin{align}
Q^{{\pi}}_{sa} := \underbrace{\sum_{s' \in \bS} p_{sas'} V^{{\pi}}_{s'}}_{\text{long-run value}} \ + \   \underbrace{p_{sac}}_{\text{one-step value}} \ \forall (s,a) \in \bS \times \bA,
\label{eq:CFOQpi}
\end{align} 
which denotes the eventual conversion probability of a consumer in state $s \in \bS$ with the firm taking action $a \in \bA$ in the current period and following policy ${\pi}$ from then on. 

\bold{Optimal Policy and Optimal Value Function.}  A policy $\pi$ is optimal iff $V^{{\pi}}_s \ge V^{{\pi'}}_s \ \forall s \in \bS, \ \forall \pi'.$
We use the notation $\pi^*$ to denote an optimal policy and define the optimal state-value function $V^*_s := V^{\pi^*}_s \ \forall s \in \bS$ and the optimal action-value function $Q^*_{sa} := Q^{\pi^*}_{sa} \ \forall (s,a) \in \bS \times \bA$.  We use the notation $\mathbf{Q}^* := [Q^*_{sa}]_{(s,a) \in \bS \times \bA}$.

\subsection{Online Optimization for the Conversion Funnel MDP} \label{sec:optimization}
We now state the \textit{conversion funnel optimization} (CFO) 
problem the firm wishes to solve. 
Suppose there are $N$ sequential consumers.
Denote by $z_n \in \{0,1\}$ whether or not consumer $n \in \{1, \ldots, N\}$ converts. 
The objective of the firm is to maximize the average expected reward over all consumers, i.e, 
$\frac{r}{N} \sum_{n=1}^N\mathbb{E}[z_n], $
which is equivalent to maximizing the average conversion probability 
$\frac{1}{N} \sum_{n=1}^N\mathbb{E}[z_n]$
since $r$ is a constant. (Hence,  $r=1$ is wlog.)
If the firm knew the true MDP $\mathcal{M}$, 
solving the CFO problem is the same as identifying the
optimal policy, and this can be done  using standard 
value/policy iteration techniques
\citep{sutton2018reinforcement}.
However,  the main challenge is the firm does not know the true model
$\cM$. 

Constructing a ``good'' state space $\bS$ and estimating the corresponding transition probabilities $\cP$ can be challenging since they must explain consumer behavior to high accuracy.
When the firm has 
  sufficient 
  data, as is the case in the application discussed in \S\ref{sec:data},
  a reliable model for 
  consumer behavior 
  can be estimated.
  However, for a new product or when there 
  is a
  concept/covariate shift,  constructing $\bS$ and estimating $\cP$ is challenging,
  especially so when the ``true''
  state space is high dimensional. For example, when the consumer behavior is dictated by
  temporal, awareness, and engagement dimensions as in \S\ref{sec:data},
  the number of states $S \approx 10,000$, even after bucketing the awareness and
  engagement counts to $\{0,1,2+\}$ and $\{0,1+\}$, respectively. 
  Consequently, 
  $\cP \equiv [p_{sas'}]_{(s,a,s')}$ 
  has 
  around $10^4
  \times A \times 4$ non-zero parameters\footnote{The count $10^4
  \times A \times 4$ corresponds to the number of feasible $(s,a,s')$
  tuples. There are around $10^4$ states. At each state, there are $A$
  possible actions ($A+1$ to be precise) and for each state-action pair,
  there are 4 possibilities in terms of the consumer behavior (ignore,
  open, click, or convert).}, which is over a hundred thousand for 
  $A \ge 3$. 
  Hence,
  we do not assume any apriori knowledge over
  $\cP$.
We do assume that the firm has decided on a definition of $\bS$, which can possibly correspond to the observable data as in \S\ref{sec:data}.  Our framework allows the firm to choose an \emph{arbitrary} $\bS$ that it seems appropriate and we note that state space construction is a topic of recent interest \citep{singh2003learning,  bennouna2021learning}.
The ``true'' consumer behavior can either be low- or high-dimensional, but the firm does not know that apriori. If it incorrectly assumes the behavior to be low dimensional, then it might suffer from suboptimality in its marketing actions (as we illustrate in our \S\ref{sec:scalability} numerics).  Accordingly, it is desirable to start with a high dimensional state space (e.g., model 12 in \S\ref{sec:data} with $S \approx 10,000$) and let the decision-making algorithm learn which dimensions are important.  This highlights the need for \emph{scalability}, and we revisit this discussion in our \S\ref{sec:numerics} and \S\ref{sec:scalability} numerics.

In order to make intervention decisions, the firm can use information
collected by interacting with the consumers. As before, let $s_{nt}$ denote the state of consumer $n$ at time $t$ and $a_{nt}$ the corresponding intervention.  As a result, consumer $n$ transitions to state $s_{n,t+1}$. 
Let $\cH_t :=\{(s_{n\tau}, a_{n\tau},s_{n,\tau+1}): \tau < t,  n \le N\}$ denote
the set of all interactions the firm has had with all the consumers
until time $t$. To make an intervention decision at time $t$, the firm is allowed to use the information in $\cH_t$.

When
  the number of states 
  is
  large (e.g.,  $S \approx 10,000$
  for the application in \S\ref{sec:data}),
  this is a 
  large-scale
   learning problem.
  Consequently,  as we numerically illustrate in \S\ref{sec:numerics}, 
  model-based learning approaches 
  that
  learn/estimate the $S^2 A$-dimensional
  $\cP$ 
  and compute the
  corresponding optimal policy after each update of $\cP$
  are impractical.
  One alternative is
    to use deep RL.  Although RL is scalable and can 
  have very good empirical performance, 
  it lacks 
  \emph{interpretability}. With this in mind, we 
  focus on designing a scalable and interpretable algorithm that despite
  being simple, comes with an optimality guarantee. To do so, we leverage
  the terminal reward structure of the conversion funnel MDP and the
  Thompson sampling algorithm for multi-armed bandit. 
  
  \section{Model-Free Approximate Bayesian Learning} \label{sec:MFABL}
We now propose an algorithm to solve the problem.  We build intuition by presenting TS for multi-armed bandit in
  \S\ref{sec:CFOAlgoMotivation} and use it to build our approach in
  \S\ref{sec:CFOOurAlgo}, followed by a discussion of its properties in
  \S\ref{sec:MFABLproperties}. 
  
\subsection{Motivation: Thompson Sampling (TS) for Multi-Armed Bandit} \label{sec:CFOAlgoMotivation}
To discuss the TS algorithm, we briefly define the multi-armed bandit
model \citep{sutton2018reinforcement}. The multi-armed bandit can be seen
as a special case of our conversion funnel MDP. In particular, if we
restrict the number of states in $\mathbb{S}$ to 1 and allow the firm
to interact only once with each consumer (as opposed to sequential
interactions), the conversion funnel reduces a multi-armed bandit. The
multiple actions available to the firm in the action space $\mathbb{A}$
represent the ``arms'' of the bandit. The consumer dynamics in a
multi-armed model are as follows: 
\begi
\item Consumer 1 arrives and firm takes action $a \in \mathbb{A}$.
\item Consumer converts w.p.\ $p_a$.
\item If the consumer converts, the firm earns a reward equal to $r$ (equals 1 wlog).
\item Then, consumer 2 arrives. And so on.
\endi

The firm's objective is to maximize its expected reward over all consumers and the challenge is that it does not know the parameters $[p_a]_{a \in \mathbb{A}}$.
TS~\citep{thompson1933likelihood}  is a well-known algorithm to tackle the multi-armed bandit problem (see Algorithm~\ref{alg:CFOTSBandits}).  We explain its working via Example \ref{example:bandit}.

\begin{algorithm}[ht]
\begin{algorithmic}[1]
\small
\REQUIRE Prior counts $(\alpha_{a}, \beta_{a}) \ \forall a \in \mathbb{A}$
\FOR{$n=1$ to $N$}
\STATE{$q_a \sim \text{Beta}(\alpha_a, \beta_a) \ \forall a \in \mathbb{A}$ \hfill \% generate samples}
\STATE{$a^* = \argmax_{a \in \mathbb{A}} q_a$ \hfill \% play action with the highest sample value}
\IF{consumer $n$ converts}
\STATE{$\alpha_{a^*} \leftarrow \alpha_{a^*} + 1$  \hfill \% update belief using Beta-Bernoulli conjugacy}
\ELSE
\STATE{$\beta_{a^*} \leftarrow \beta_{a^*} + 1$  \hfill \% update belief using Beta-Bernoulli conjugacy}
\ENDIF
\ENDFOR
\end{algorithmic}
\caption{Thompson Sampling for Multi-Armed Bandit with $N$ Consumers}
\label{alg:CFOTSBandits}
\end{algorithm}

\begin{figure}[ht]
\centering
\begin{subfigure}[b]{0.45\textwidth}
\centering
\begin{tikzpicture}
   \node[state, line width=0.5mm, minimum size=20pt, fill=gray!20!white] (1) {1};
   \node[state, line width=0.5mm, minimum size=20pt, fill=gray!20!white, above right=1cm of 1] (c) {$c$};
   \node[state, line width=0.5mm, minimum size=20pt, fill=gray!20!white, below right=1cm of 1] (q) {$q$};
   \draw[every loop, >=latex, line width=0.5mm]
    (1) edge[auto=right, color=blue] node {$0.3$} (c)
    (1) edge[auto=left, color=blue] node {$0.7$} (q)
    (1) edge[dashed, auto=right, bend right, color=red] node {$1$} (q);
\end{tikzpicture}
\caption{\small Model of consumer behavior}
\label{fig:ExampleOneState}
\end{subfigure}
\begin{subfigure}[b]{0.45\textwidth}
\centering
\includegraphics[scale=0.4]{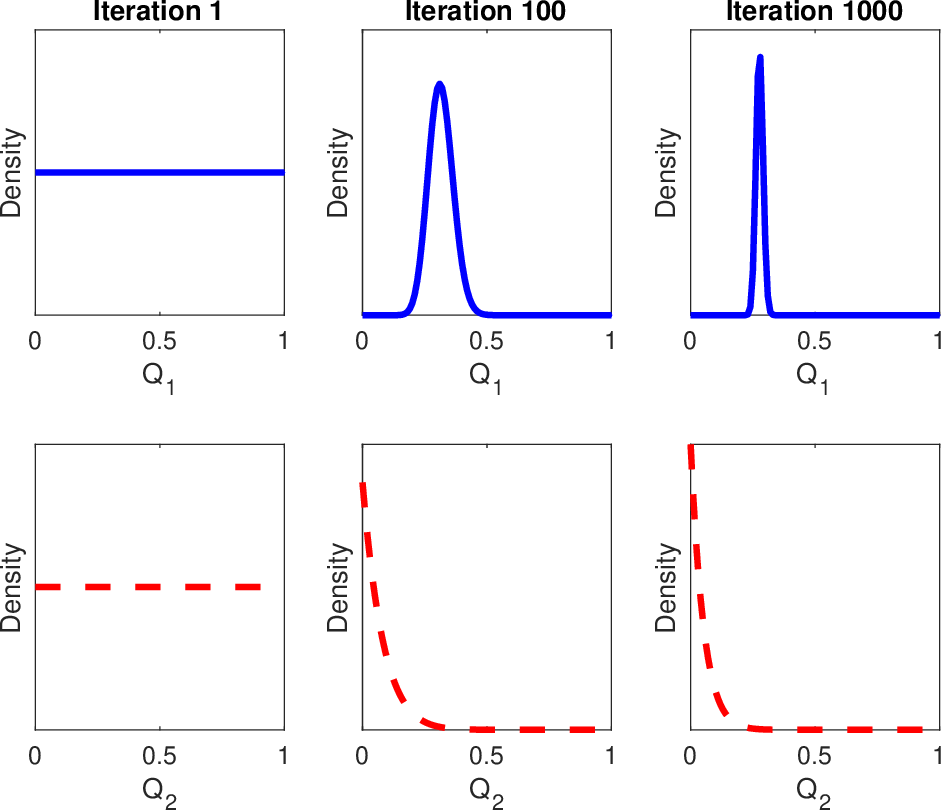}
\caption{\small Performance of Thompson sampling}
\label{fig:ExampleBandit}
\end{subfigure}
\caption{\normalfont Illustration of Example \ref{example:bandit}. In subplot (a), we show the model for consumer behavior. If firm
  takes action $a \in \{1,2\}$, the consumer converts ($c$) w.p.\ $p_a$.  Solid blue action ($a=1$) corresponds to a
  ``call-to-action'' type ad with a higher one-step conversion probability
  than the dashed red action ($a=2$), e.g., $p_1 = 0.3$ and $p_2 = 0$.  
  In subplot (b),
  we illustrate the performance of TS. We display
  the evolution of the Beta belief over $[p_a]_{a=1}^2$ (denoted by $[Q_a]_a$) as a function of
  number of consumers (iteration). The initial count is set as
  $(\alpha_a, \beta_a) = 
  (1,1)$ for $a \in \{1,2\}$,  and hence, the initial belief (iteration 1)
  is uniform(0,1). As TS interacts with more consumers (see iterations 100
  and 1000),  action 1 (the optimal action) is chosen more often than action
  2, and hence, the belief over action 1 concentrates around the true value
  of $0.3$.  We see the
  belief over action 2 essentially remains the same after the 100th
  consumer,  suggesting TS 
  learned the optimal action by interacting with 100 consumers. 
}
\label{fig:ExampleOneStateMain}
\end{figure}
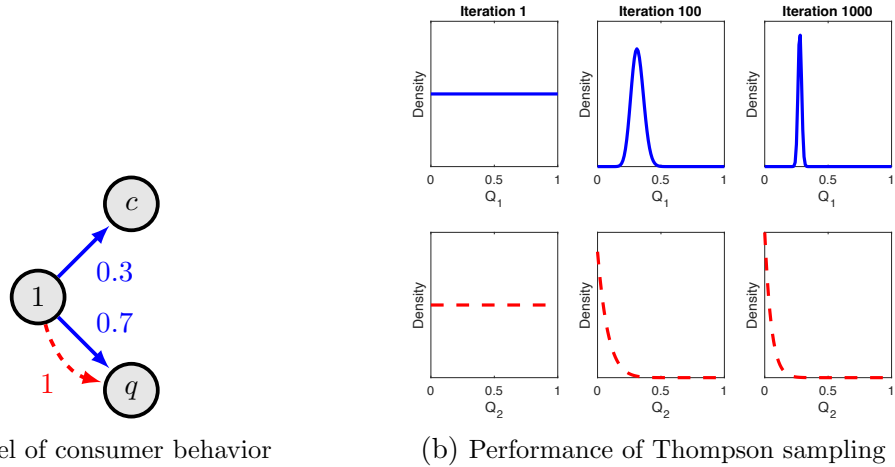

\begin{example}[Bandit] \label{example:bandit}
  Consider a setting in which a consumer is only in one state (Figure
  \ref{fig:ExampleOneState}).  The firm needs to decide
  between two possible ad actions: $a \in \{1, 2\}$ with conversion
  probabilities $[p_a]_{a=1}^2$ obeying $p_1 > p_2$.  
  The firm wishes to maximize its expected reward over all
  consumers and the challenge is that  
  the  
  parameters $p_a$, $a \in \{1,2\}$, are unknown apriori.
  TS  is simple and interpretable:
  it maintains a Beta$(\alpha_a,\beta_a)$ belief (denoted by $Q_a$) on the value $p_a$ of action $a \in
  \{1,2\}$. 
  With no prior information,  we initialize the prior parameters to
  $(\alpha_{a}, \beta_{a}) = (1,1)$, which 
  results in 
  a uniform prior over
  $p_a$.  
  The algorithm can be split into two modules: (1) optimization and
  (2) attribution:
  \begin{enumerate}
  \item The optimization module (lines 2 and 3) corresponds
    to action selection.   
    When a consumer arrives, the algorithm samples from Beta$(\alpha_a,\beta_a)$ for $a
    \in \{1,2\}$ (line 2) and chooses the action with the higher sampled
    value (line 3), i.e.,  TS takes action $a$ with probability
    it is optimal.
  \item The attribution module (lines 4 to 8) updates the Beta
    beliefs by  
    attributing positive 
    credit to the played action if the consumer converts (line 5) or
    attributing negative credit if she does not (line 7). 
    This update is exact as it obeys the Bayes rule (Beta-Bernoulli
    conjugacy).
  \end{enumerate}
  Despite its simplicity, TS
  exhibits strong theoretical guarantees when consumer behavior is
  dictated by a bandit \citep{agrawal2012analysis} and works well in
  marketing applications \citep{chapelle2011empirical}. We illustrate the performance of TS for this
  example in Figure \ref{fig:ExampleBandit}, where TS finds the optimal
  action after interacting with only 
  approximately
  100 consumers.
\end{example}

Although TS performs well for a bandit setting, it is not designed to handle
multiple states,  and therefore, cannot be applied in 
our application. Handling multiple states requires one to maintain a belief over not just the one-step conversion probability but the total value of each state-specific action (one-step plus long-run value as in \eqref{eq:CFOQpi}).  Next,  we develop an algorithm to capture the total value while maintaining the
scalable and interpretable Beta-Bernoulli structure of TS.  

\subsection{The Proposed Algorithm: Model-Free Approximate Bayesian Learning (MFABL)} \label{sec:CFOOurAlgo}
\textit{Model-free\footnote{\label{ft:CFOmodelfree}As noted in
    \S\ref{sec:intro} (Footnote~\ref{ft:CFOModelFreeOne}), ``model-free''
    does not mean that there does not exist an underlying model. Instead,
    it means that there exists an underlying model $\mathcal{M}$ (the
    conversion funnel MDP) but our algorithm does not learn/estimate
    $\mathcal{M}$. In particular,  MFABL never attempts to learn/estimate
    the transition probabilities $\mathcal{P}$ but only attempts to learn $\mathbf{Q}^*$. Approaches that estimate/learn the transition
    probabilities $\mathcal{P}$ are called ``model-based''.} approximate
  Bayesian learning} (MFABL) extends the simple Beta-Bernoulli structure
of TS to the conversion funnel MDP and is presented as
Algorithm~\ref{alg:CFOOurAlgo}. Due to the terminal reward structure in
our conversion funnel, the $Q$-value of an action at a given state
represents the \emph{eventual} conversion probability (as discussed in
\S\ref{sec:consumermodel}). Hence, similar to TS, we assign a
$\text{Beta}(\alpha_{sa}, \beta_{sa})$ belief to the ``value'' $Q_{sa}$ of taking
action $a \in \mathbb{A}$ at state $s \in \mathbb{S}$. 

\begin{algorithm}[ht]
\begin{algorithmic}[1]
\small
\REQUIRE Prior counts $(\alpha_{sa}, \beta_{sa}) \ \forall (s,a) \in \mathbb{S} \times \mathbb{A}$, $\epsilon$
\STATE{$(\alpha_{ca}, \beta_{ca}) = (\infty, 1)$ and $(\alpha_{qa}, \beta_{qa}) = (1, \infty)$ for all $a \in \mathbb{A}$ \hfill \% prior counts for states $c$ and $q \ \ $}
\FOR{$t=1,2,\ldots$ \hfill \% until there exists an ``active'' consumer (has not converted or quit)}
\FOR{$n \in \mathbb{N}_t$ \hfill \% $\mathbb{N}_t \subseteq \{1, \ldots,N\}$ denotes the set of active consumers at time $t$}
\STATE{Observe state $s = s_{nt}$ of consumer $n$ at time $t$}
\STATE{$q_{sa} \sim \text{Beta}(\alpha_{sa}, \beta_{sa}) \ \forall a \in \mathbb{A}$ \hfill \% generate samples $ \ $}
\STATE{$a^* = \argmax_{a \in \mathbb{A}} q_{sa}$ with $\epsilon$-greedy \hfill \% highest sample value with $\epsilon$-greedy $ \ $}
\STATE{Consumer transitions to state $s' = s_{n,t+1} \in \mathbb{S}^+$}
\STATE{$f_{s'} \sim \text{Bernoulli}\left( \max_{a' \in \mathbb{A}} \frac{\alpha_{s'a'}}{\alpha_{s'a'} + \beta_{s'a'}} \right)$ \hfill \% generate feedback $ \ $}
\IF{$f_{s'} = 1$}
\STATE{$\alpha_{sa^*} \leftarrow \alpha_{sa^*} + 1$ \hfill \% approximate posterior update mimicking Beta-Bernoulli $ \ $}
\ELSE
\STATE{$\beta_{sa^*} \leftarrow \beta_{sa^*} + 1$ \hfill \% approximate posterior update mimicking Beta-Bernoulli $ \ $}
\ENDIF
\ENDFOR
\ENDFOR
\RETURN{$\mathbf{Q} := [Q_{sa}]_{(s,a) \in \mathbb{S} \times \mathbb{A}} \text{ where } Q_{sa} \stackrel{d}{=} \text{Beta}(\alpha_{sa}, \beta_{sa})$}
\end{algorithmic}
\caption{MFABL with $N$ Consumers}
\label{alg:CFOOurAlgo}
\end{algorithm}

\bold{Optimization Module.} MFABL mimics TS (lines 5 and 6 in Algorithm~\ref{alg:CFOOurAlgo}) in its action selection. When a consumer is in state $s \in \mathbb{S}$, it samples $q_{sa} \sim \text{Beta}(\alpha_{sa}, \beta_{sa})$ for $a \in \mathbb{A}$ and plays the action with the highest sample value (with $\epsilon$-greedy), i.e.,
\begin{align*}
a^* = \begin{cases}
\argmax_{a \in \mathbb{A}} q_{sa} & \text{ w.p.\ } 1-\epsilon \\ 
\texttt{UniformAtRandom}(\mathbb{A}) & \text{ w.p.\ } \epsilon.
\end{cases}
\end{align*} 
$\epsilon$-greedy is used  to ensure 
each state-action pair is visited infinitely often in order to ensure the convergence of MFABL to $\mathbf{Q}^*$
(Theorem~\ref{theorem:CFOConvergence}). In
theory,  $\epsilon$ can be arbitrarily small but strictly positive.   

\bold{Challenge in Attribution.} Belief update in  the conversion
funnel is non-trivial because, 
unlike in a bandit,  
one has to 
capture the long-run value in addition to the
one-step value.
For example, suppose the consumer path is  
$
s \stackrel{a}{\to} s' \stackrel{a'}{\to} c,
$
i.e., the
consumer starts in state $s \in \mathbb{S}$, 
the firm takes action $a \in \mathbb{A}$,
causing a 
transition to state $s' \in \mathbb{S}$, where the firm takes action $a'
\in \mathbb{A}$, 
leading
to conversion. Given this path, how does one attribute the positive credit to $(\alpha_{sa}, \beta_{sa})$ and $(\alpha_{s'a'}, \beta_{s'a'})$? Does one increase both $\alpha_{sa}$ and $\alpha_{s'a'}$ by 1? As another example, consider the path
$
s \stackrel{a}{\to} s' \stackrel{a'}{\to} q.
$
Given the consumer quit after visiting $(s,a)$ and $(s',a')$, does one increase both $\beta_{sa}$ and $\beta_{s'a'}$ by 1? But what if taking action $a$ at state $s$ was optimal and it was the action $a'$ at state $s'$ that was the ``culprit''? Should $(s,a)$ be still ``penalized''? The challenge here is to identify the ``contributions'' of various state-specific actions to the final convert/quit outcome. 

\bold{Attribution Module.} To overcome this challenge, 
we propose two attribution strategies that extend the attribution employed
by TS.
We start with the first strategy (lines 8
to 13 in Algorithm~\ref{alg:CFOOurAlgo}).
Suppose taking action $a \in
\mathbb{A}$ at state $s \in \mathbb{S}$ transitions the consumer to state
$s' \in \mathbb{S}^+$. Intuitively speaking, if the belief on the value at state $s'$ is ``high'', the state-action pair
$(s,a)$ should receive positive attribution. 
For example, in the extreme case when $s' = c$, we should increase
$\alpha_{sa}$ by 1. 
On the other hand, if the belief on the value at state $s'$ is ``low'', then $(s,a)$ should receive
negative attribution.  
For example,  when $s' = q$, we should increase
$\beta_{sa}$ by 1. 
To operationalize this intuition, MFABL generates a
binary ``feedback'' $f_{s'}$ from state $s'$ as follows: 
\begin{align}
f_{s'} \sim \text{Bernoulli}\left( \max_{a' \in \mathbb{A}} \frac{\alpha_{s'a'}}{\alpha_{s'a'} + \beta_{s'a'}}  \right).
\label{eq:CFOFeedback}
\end{align}
If $f_{s'} = 1$,
MFABL positively attributes $(s,a)$ by 
updating
$\alpha_{sa} \leftarrow \alpha_{sa} + 1$. 
Else, if 
$f_{s'} = 0$, 
MFABL negatively attributes $(s,a)$ by 
updating 
$\beta_{sa} \leftarrow \beta_{sa} + 1$. 
The term $\frac{\alpha_{s'a'}}{\alpha_{s'a'} + \beta_{s'a'}}$ in
\eqref{eq:CFOFeedback} equals the expected value of the belief
$Q(s',a')$. Hence, the feedback is generated using the action
with the highest expected value at $s'$. 

The second attribution strategy (not shown in pseudo-code) is even simpler
and resembles the widely used \emph{uniform attribution} rule.  Suppose we
use the above-mentioned optimization module for action selection and
observe the following consumer path: 
$
s_1 \stackrel{a_1}{\to} s_2 \stackrel{a_2}{\to} \ldots
\stackrel{a_{\ell-1}}{\to} s_{\ell} \stackrel{a_{\ell}}{\to} c. 
$
Given the consumer converted,  we uniformly attribute a positive credit to
each state-action pair that appeared in the path by increasing the
corresponding $\alpha_{sa}$ values by 1. 
Else, if we observe a path that does not convert,  we
uniformly attribute a negative credit to each state-action pair that
appeared in the path by increasing $\beta_{sa}$ values by 1. To
distinguish this attribution rule from the first, we refer the corresponding algorithm as \emph{pathwise MFABL} (pMFABL). 

Both MFABL variants reduce to
TS when the MDP is simplified to a bandit.  
Despite similarities,  we note that there are fundamental
differences between TS for multi-armed bandit
(Algorithm~\ref{alg:CFOTSBandits}) and MFABL for conversion funnel MDP
(Algorithm~\ref{alg:CFOOurAlgo}). The belief update in TS is exact, i.e.,
satisfies Bayes' rule, whereas the belief update in MFABL is approximate.  
In fact, it is not clear if one can maintain an exact \textit{model-free} belief in the conversion funnel MDP in a tractable manner. Accordingly, the belief we maintain in MFABL can be seen as an approximation to the true posterior.  A natural question to ask is whether it converges to $\mathbf{Q}^*$, which we explore next along with other properties such as scalability and interpretability.

\subsection{Properties of MFABL} \label{sec:MFABLproperties}
We start by establishing the asymptotic optimality of MFABL, which is non-trivial given the approximate Bayes' update.  We provide a proof sketch here and a formal proof is in \S\ref{sec:CFOConvergenceProof}. 

\begin{restatable}[\bold{Asymptotic Convergence}]{theorem}{CFOConvergence}
\label{theorem:CFOConvergence}
Let $\mathbf{Q}^N$ denote the output of Algorithm~\ref{alg:CFOOurAlgo} with $N$ consumers. Then, for any $\epsilon > 0$ and prior counts $\alpha_{sa}
> 0$ and $\beta_{sa} > 0$ for $(s,a) \in \mathbb{S} \times \mathbb{A}$, 
\begin{align*}
\mathbf{Q}^N \to \mathbf{Q}^* \text{ w.p.\ 1 as } N \to \infty.
\end{align*}
\end{restatable}

\textit{Proof Sketch.} The key insight is to analyze how the belief in MFABL
evolves \textit{in expectation}. 
Let
$\mathbf{Q}(i) :=
[Q_{sa}(i)]_{(s,a)}$ denote the belief in MFABL
at the start of iteration $i \in \{1, 2, \ldots\}$ where $Q_{sa}(i)
\stackrel{d}{=} \text{Beta}(\alpha_{sa}(i), \beta_{sa}(i)) \ \forall (s,a)
\in \mathbb{S} \times \mathbb{A}$. (By an ``iteration'' $i$, we refer to a
$(t,n)$ pair in Algorithm~\ref{alg:CFOOurAlgo}.) We use
$\mathbf{\overline{Q}}(i) := [\overline{Q}_{sa}(i) =
\mathbb{E}[Q_{sa}(i)]]_{(s,a)}$ to denote
the expected value of $\mathbf{Q}(i)$. 
We first show how the MFABL update in the $\mathbf{Q}$-space translates to an update in the $\mathbf{\overline{Q}}$-space. Second, we establish that the
update process in the $\mathbf{\overline{Q}}$-space is an asynchronous
stochastic approximation scheme to the Bellman optimality equations
corresponding to $\mathbf{Q}^*$. Third, we leverage the stochastic
approximation theory to prove that $\mathbf{\overline{Q}}(i)$ converges to
$\mathbf{Q}^*$. Finally, we show that the variance of the Beta belief
$\mathbf{{Q}}(i)$ goes to zero and hence, $\mathbf{{Q}}(i)$ converges to
$\mathbf{Q}^*$. 
\hfill $\Halmos$

\leaveline

Theorem~\ref{theorem:CFOConvergence} combined with MFABL's action selection (lines 5 and 6 of Algorithm~\ref{alg:CFOOurAlgo})
implies  that, asymptotically, MFABL picks an optimal action w.p.\ at least
$1-\epsilon$. This establishes the asymptotic optimality of MFABL with
high probability. In other words, as the firm interacts with more
consumers, MFABL learns the optimal personalized interventions. Note 
that this result holds irrespective of the prior counts supplied to
Algorithm~\ref{alg:CFOOurAlgo} as an input and we prove it only for MFABL
(but not pMFABL).   The key difficulty in extending our proof
  to 
  pMFABL is that
  we are not able to establish that the noise term, i.e., the difference
  between random $Q$-value and its expected value, 
  is
  zero mean (see condition 3 of Proposition \ref{prop:CFOASAConvergence} in
  \S\ref{sec:CFOConvergenceProof}).

Next, we analyze the convergence rate of MFABL as a function of $N$, i.e., how fast $\mathbf{Q}^N$ converges to
  $\mathbf{Q}^*$ as a function of $N$.  
  As before,
  we first  understand how MFABL evolves in expectation and then,
  bound the error due to the noisy updates. Let $\mathbf{\overline{Q}}^N =
  \mathbb{E}[\mathbf{Q}^N]$ and $\lVert \cdot \rVert$ denote an arbitrary norm. Then,
  \begin{align}
    \lVert \mathbf{Q}^N - \mathbf{Q}^* \rVert = \lVert \mathbf{Q}^N -
    \mathbf{\overline{Q}}^N + \mathbf{\overline{Q}}^N  - \mathbf{Q}^*
    \rVert  \le \underbrace{\lVert \mathbf{Q}^N - \mathbf{\overline{Q}}^N
    \rVert}_{(\star)} + \underbrace{\lVert \mathbf{\overline{Q}}^N  -
    \mathbf{Q}^* \rVert}_{(\square)}. 
\label{eq:ConRateDecomposition}
\end{align}
Recall that, 
$\mathbf{\overline{Q}}^N$ corresponds to an asynchronous stochastic
approximation scheme for solving  the Bellman optimality equations
associated with $\mathbf{Q}^*$.  This connection proves useful while
analyzing the $(\square)$ term in \eqref{eq:ConRateDecomposition}. In
particular,  it allows us to use off-the-shelf convergence rates for
Q-learning (e.g., \cite{even2003learning}) and provide a meta-result of
the following form: given \emph{any} convergence rate for Q-learning, we
convert that into one for MFABL. Of course,  doing so requires analyzing
the $(\star)$ term, which we do carefully by invoking standard
concentration inequalities. We state the formal result in Theorem
\ref{theorem:ConRate} below and expand upon the off-the-shelf convergence
rate next.

In the proof for Theorem \ref{theorem:ConRate} in \S\ref{sec:ConRateProof},  we establish that  $\mathbf{\overline{Q}}$
mimics asynchronous  Q-learning with a linear\footnote{As we discuss in \S\ref{sec:ConRateProof} (Remark \ref{rem:polystep}),  we can modify MFABL so that the corresponding $\mathbf{\overline{Q}}$ process mimics Q-learning with a ``polynomial'' learning-rate.  Then, we can use the rate from Theorem 4 of \cite{even2003learning}.} learning-rate. Therefore,  we can leverage the corresponding Q-learning convergence rates to analyze $(\square)$. One such rate is established in Theorem 5 of \cite{even2003learning}.  Thus, for our purposes, we assume access to the following generic convergence rate for Q-learning (further details in \S\ref{sec:ConRateProof}): there exists a strictly positive vector $\pmb{v}$ (of the same dimension as $\mathbf{Q}$) such that for all $\varrho > 0$,
\begin{align}
\mathbb{P}\left\{\lVert \mathbf{\overline{Q}}^N  - \mathbf{Q}^* \rVert_{\pmb{v}} < \varrho \right\} >  1 - \delta(N,  \varrho), 
\label{eq:QLearningOffTheShelfRate}
\end{align}
where $\lVert \pmb{x} \rVert_{\pmb{v}} := \max_i \frac{|x_i|}{v_i}$ is the weighted maximum norm with $\pmb{v} > \pmb{0}$, and $\lim_{N \to \infty}\delta(N,  \varrho) = 0$ for all $\varrho > 0$.  For instance,  in Theorem 5 of \cite{even2003learning}, $\delta(N, \varrho)$ exhibits an $e^{-N}$ (ignoring constants) dependence on $N$ (easy to verify via straightforward algebra). We are now ready to state Theorem \ref{theorem:ConRate} (proof in
\S\ref{sec:ConRateProof}) that provides a probabilistic bound on the error $\lVert \mathbf{Q}^N - \mathbf{Q}^* \rVert_{\pmb{v}}$ as a function of $N$.

\begin{restatable}[\bold{Convergence Rate}]{theorem}{ConRate}
\label{theorem:ConRate}
Let $\mathbf{Q}^N$ denote the output of Algorithm~\ref{alg:CFOOurAlgo} with $N$ consumers.  Given a generic off-the-shelf convergence rate for Q-learning as in \eqref{eq:QLearningOffTheShelfRate} (and the corresponding $\pmb{v} > 0$),  we have
\begin{align*}
\mathbb{P}\left\{ \lVert \mathbf{Q}^N - \mathbf{Q}^* \rVert_{\pmb{v}} < \varrho \right\} > 1 - \delta(N,  \varrho) - \mathcal{O}\left( \frac{1}{\varrho^2 N} \right) - \mathcal{O}\left(e^{-N}\right),
\end{align*}
for all $\varrho > 0$.  (As in Theorem \ref{theorem:CFOConvergence}, this result holds for any $\epsilon > 0$ and  $(\alpha_{sa}, \beta_{sa}) > 0$ for $(s,a) \in \mathbb{S} \times \mathbb{A}$.)
\end{restatable}

Given the  $e^{-N}$ dependence on $N$ of $\delta(N,\varrho)$ (discussed above), it follows that the term that dominates is $\frac{1}{\varrho^2 N}$. 
Having established the theoretical desirability, we next discuss some
important practical considerations such as scalability, interpretability,
prior information, and concept shift.

\bold{Scalability.} Scalability is critical in a large-scale system as in
\S\ref{sec:data}. MFABL stores $\{\alpha_{sa}, \beta_{sa}\}_{a \in
  \bA}$ for only those states that it has encountered so far, i.e.,  in the
worst case approximately $2SA$
parameters.
  In terms of computational cost, taking an intervention decision for a
consumer at a given time requires MFABL to generate a sample from $A+1$
Beta distributions and pick the maximum among them (lines 5 and 6 of
Algorithm~\ref{alg:CFOOurAlgo}), which can be done in real-time.   
In fact, MFABL only operates on the states that are visited, which is in
contrast with model-based approaches that require one to optimize over all
the states. For example, if the firm incorrectly assumes the state space
to be large and most of the states are not observed, then MFABL
simply ignores those redundant states and learns the values of the
relevant states. 
The value of such high scalability will become evident in our \S\ref{sec:numerics} and \S\ref{sec:scalability} numerics.
Deep RL-based approaches approximate the value function, and hence, reduce the dependence on $S$ to the dimensionality of the approximator. However, the approximation might not be exact and such methods lack interpretability,  a property we discuss next.  

\bold{Interpretability.} MFABL is interpretable in both its modules:
action selection and belief update.  For action selection, as in Figure
\ref{fig:ExampleBandit}, MFABL picks an action with the highest
probability of being optimal (according to the current beliefs).  To
update the beliefs, MFABL employs a simple attribution strategy that maps
to a transparent closed-form update. 
Such simplicity naturally has important practical implications in terms of
algorithmic adoption.  Not only it is easy to explain to managers,  but it
is easy to implement as well, only requiring the firm to track the
$(\pmb{\alpha}, \pmb{\beta})$ counts and update them via simple
interpretable rules.  
We further showcase the interpretability of MFABL/pMFABL when we present the numerical results in \S\ref{sec:numerics}.

\bold{Prior Information.} A firm might have useful prior information regarding
the consumer behavior. For instance, using third-party data or its experience with
an earlier version of the product, the firm might know that the consumers
who are in a certain state are highly likely to convert if they are shown
a specific intervention. 
MFABL allows the firm to encode such state-action specific prior knowledge by setting appropriate prior counts $(\pmb{\alpha}, \pmb{\beta})$ for the ``value'' of each state-action pair.
It is worth mentioning that even if the firm
provides an incorrect prior, the convergence of MFABL to optimality
(Theorem~\ref{theorem:CFOConvergence}) holds. 

\bold{Concept Shift. } Concept shift refers to the consumer behavior
changing over time, say from $\mathcal{P}_1$ to $\mathcal{P}_2$.
The firm does not know $\mathcal{P}_1$ and $\mathcal{P}_2$ apriori but might (or might not) know the time of shift.
As discussed in \citet{simester2020targeting}, an algorithm that
\emph{estimates} the model from phase 1 data and uses it to make phase 2 decisions will generally result in suboptimal policies.  
Given its \emph{learning-based} nature, MFABL is robust to concept shift.  
It initially makes progress towards learning $\mathbf{Q}^*_1$.
As soon as there is a shift, the exploratory nature of MFABL
\emph{automatically} forces it to learn $\mathbf{Q}^*_2$.
One way to see this is the belief maintained by MFABL at the end of phase 1 serves as the prior for phase 2 and
Theorem~\ref{theorem:CFOConvergence} holds for an arbitrary prior.  There is nothing special about the two-phase setting and this reasoning extends to a multi-phase setting. Of course,  such robustness is not special to MFABL but all learning-based algorithms. Nonetheless, it is a desirable property worth mentioning and we numerically understand its value in \S\ref{sec:numerics}.

\section{Numerical Experiments} \label{sec:numerics}
We now shift our focus to numerically evaluating the proposed algorithm. We do so via large-scale simulations calibrated to the real data discussed in \S\ref{sec:data}. We elaborate on the simulation setup in \S\ref{sec:setup},  present the results in \S\ref{sec:results}, and end this section by illustrating how MFABL adapts under concept shift in \S\ref{sec:conceptshift}.

\subsection{Simulation Setup} \label{sec:setup}
We use the data presented in \S\ref{sec:data} to estimate the ground truth
model of consumer behavior.  As discussed in \S\ref{sec:model}, though
consumer behavior is complicated due to various carryover and spillover
effects, it can be explained via micro-level temporal, awareness, and
engagement features. In fact,  as shown in Table
\ref{table:predictionmodels},  model 12 explains consumer-level behavior
with extremely high accuracy (out-of-sample AUC of over 0.95).  
Therefore, in our simulations, we assume that the ground truth consumer behavior is given by model 12. 
In terms of the conversion funnel MDP of \S\ref{sec:consumermodel},  this
corresponds to the state \(s_{nt}\) of consumer $n$ at the beginning of
day $t$ being $s_{nt} = (t,\fx_{nt}^{\text{received}},
\fx_{nt}^{\text{opened}}, \fx_{nt}^{\text{clicked}})$, which results in a
large-scale state space ($S = 11,060$).  The actions correspond to the
types of email (including no email), and the action frequency is daily.  
The transition probabilities $\cP$ are governed by the fitted prediction
function (boosted trees) via cross-validation (details in
\S\ref{sec:xgboostdetails}).   
The initial state is $s_{n1} = (1, \pmb{0}, \pmb{0},\pmb{0})$ for all $n$
since every path starts with a sign-up and the conversion reward equals 1
wlog. 
Having a handle on a ground truth model lets us simulate consumer behavior
but none of the decision-making algorithms know the transition
probabilities $\cP$ and are required to learn in an online manner.
Accordingly,  given an input algorithm $\ALG$, the simulation proceeds
sequentially over $N$ consumers as shown in Algorithm \ref{alg:sim}.
(Seed $r$ is used to simulate consumer behavior and we average over
$R=100$ seeds as explained below.) 

\begin{algorithm}[ht]
\begin{algorithmic}[1]
\small
\REQUIRE Sequential targeting algorithm $\ALG$ and seed $r$ 
\FOR{$n=1$ to $N$}
\STATE{Sample initial state $s_{n1}$ from the ground truth model of consumer behavior}
\FOR{$t=1$ to $T$}
\STATE{Send email $a_{nt}$ (possibly no email) as dictated by $\ALG$}
\STATE{Sample next state $s_{n,t+1}$ from the ground truth model of consumer behavior}
\STATE{Update parameters of $\ALG$}
\ENDFOR
\ENDFOR
\end{algorithmic}
\caption{Simulation Setup with $N$ Consumers}
\label{alg:sim}
\end{algorithm}

Given the Bayesian nature of MFABL,  we benchmark it 
against two Bayesian approaches: (1) Thompson sampling (TS)
\citep{thompson1933likelihood} and (2) posterior sampling for
reinforcement learning (PSRL) \citep{strens2000bayesian}.   
Both are known to be state-of-the-art approaches for their respective
use cases: TS for bandits~\citep{chapelle2011empirical,agrawal2012analysis}
and PSRL for RL~\citep{osband2013more}. 
In contrast to MFABL, TS only captures
the ``one-step value'' in \eqref{eq:CFOQpi} whereas PSRL is
model-based. Hence, benchmarking with TS and PSRL enables us to understand
how much value MFABL adds by capturing the long-run dynamics of consumer
behavior and by being model-free, respectively.  The latter highlights the
value of scalability, which we discuss in \S\ref{sec:scalability} as
well. 
We note that all these approaches reduce to TS when the conversion funnel is
simplified to a bandit. 
We also benchmark against QL-UCB \citep{jin2018q},  which is the only model-free approach that comes with a regret bound.
The detailed description of the benchmark algorithms is provided in
\S\ref{sec:BenchmarksAppendix}.

We evaluate each algorithm via two metrics: (1) performance ratio and (2) compute time. The performance ratio quantifies the quality of the decisions made by algorithm $\ALG$ and is defined as 
$
\PR^{\ALG} := \frac{\sum_{n=1}^N \bE_{\ALG}[z_n]}{N v^*}.
$
Recall that $z_n \in \{0,1\}$ denotes whether or not consumer $n \in \{1, \ldots, N\}$ converts. We define $v^*$ to be the optimal conversion rate under the ground truth MDP and hence,  $\PR^{\ALG} \in [0,1]$, with a higher value denoting better performance.
We estimate $\PR^{\ALG}$ using the following unbiased Monte-Carlo
estimator: 
\begin{align}
\label{eq:PR}
\PR^{\ALG} \approx \frac{1}{R} \sum_{r=1}^R \frac{\sum_{n=1}^N z_{nr}^{\ALG}}{ N v^*},
\end{align}
where $z^{\ALG}_{nr} \in \{0,1\}$ denotes whether or not consumer $n$ in the $r$-th run associated with
$\ALG$ converted and we use $R = 100$ seeds to average over the randomness. 
The compute time metric quantifies the computational cost of each algorithm.  For each algorithm, we parallelize over the $R=100$ seeds using 100 cores of a high-performance computing cluster: \texttt{HP Enterprise XL170r} with an \texttt{E5-2650v4} CPU. None of our jobs required more than 16 GB of RAM.

\subsection{Results and Discussion} \label{sec:results}
\begin{figure}
\centering
\begin{subfigure}[]{0.45\textwidth}
\centering
\includegraphics[scale=0.4]{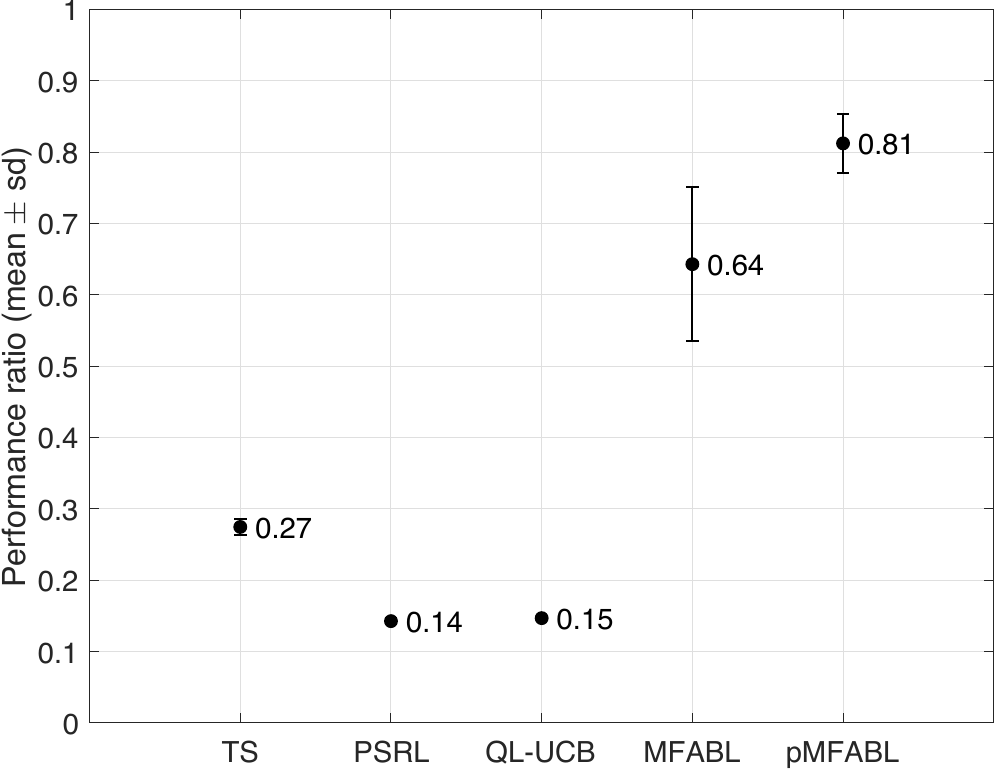}
\caption{\small PR after $N = 500,000$ consumers}
\label{fig:PRsnapshot}
\end{subfigure}
\begin{subfigure}[]{0.45\textwidth}
\centering
\includegraphics[scale=0.4]{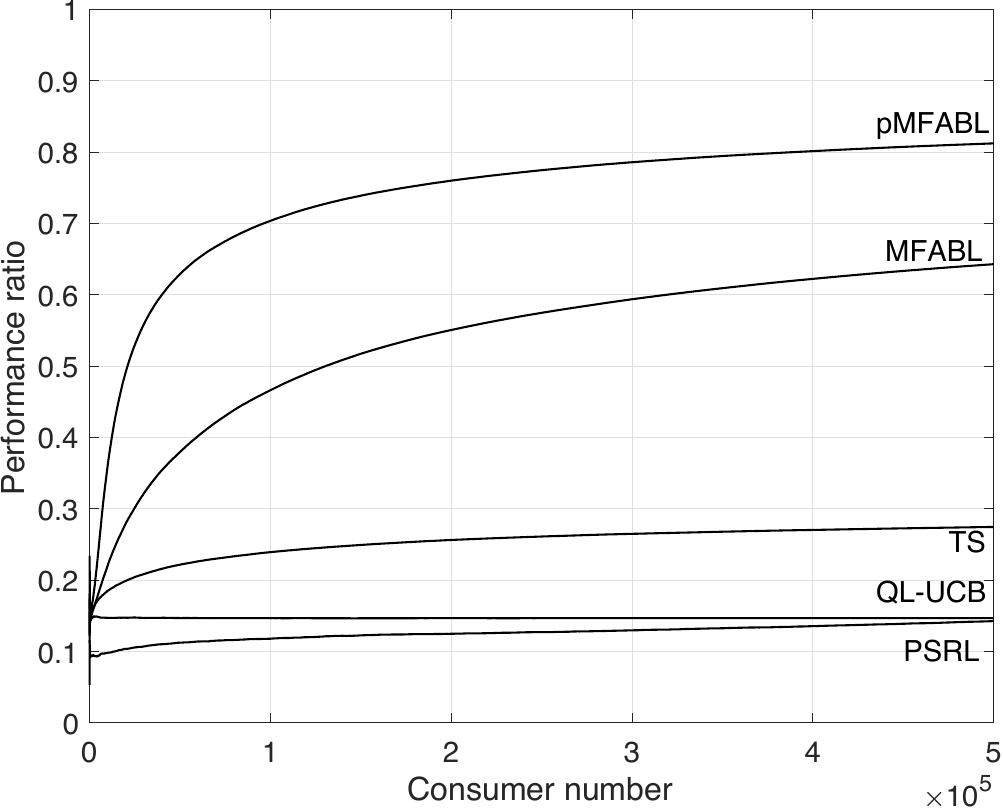}
\caption{\small Evolution of PR}
\label{fig:PRdynamic}
\end{subfigure}
\caption{\normalfont Benchmarking MFABL/pMFABL with TS, PSRL, and QL-UCB in terms of the performance ratio. In subplot (a), we show the mean and standard deviation of the PR over $R=100$ seeds with $N = 500,000$ consumers.  In subplot (b), we show the evolution of PR as we increase $N$ from 1 to 500,000 (averaged over $R=100$ seeds).} 
\label{fig:PR}
\end{figure}

In Figure \ref{fig:PR}, we show the performance ratio of the five algorithms with $N=500,000$.  TS maintains a belief over the one-step conversion probabilities and hence,  converges to a myopically optimal policy, achieving an average PR of 0.27.  However,  the myopically optimal policy ignores the long-run value and is suboptimal.
Intuitively, myopic approaches ``fail'' when the long-run value of an action dominates its myopic value.  In 
our funnel, we expect such instances to occur in earlier stages 
since there might be an intervention that does a good job of pushing the
consumer deeper into the funnel but has a lower one-step conversion
probability.   Given the large state space, the model-based PSRL is barely
able to learn after $500,000$ consumers. This is because PSRL needs to
learn $S \times A \times 4 \approx 200,000$ parameters. This results in it
having an average PR of 0.14, even lower than TS.  
QL-UCB's numerical performance is similar to PSRL. In fact, we found that 
QL-UCB essentially kept playing a completely random policy, suggesting a
failure to learn. 
Given their attribution-based model-free nature, MFABL and pMFABL learn
rapidly, beating TS, PSRL, and QL-UCB with an average PR of 0.64 and 0.81,
respectively.  In fact, as we show in Figure \ref{fig:PRdynamic}, pMFABL
achieves a PR of 0.6 in less than 50,000 consumers. This is remarkable
since we have over 50,000 state-action pairs and it shows that the
proposed approach learns over the relevant states and ignores the
redundant states (as discussed in the ``scalability'' paragraph of
\S\ref{sec:MFABLproperties}).   
Note that MFABL/pMFABL outperform 
  both PSRL and QL-UCB in our numerical experiments, even though these
  methods  are known to be near-optimal (from 
  a regret perspective). 
  This 
  is because PSRL and QL-UCB are optimal in an
  asymptotic  sense; whereas 
  our numerics are non-asymptotic, highlighting the value MFABL/pMFABL in
  the non-asymptotic setting. 
We observe similar results across a wide set of input parameters, as we
illustrate in our sensitivity analysis (\S\ref{sec:SensAppendix}). 
Also, we zoom in on the first few thousand consumers in \S\ref{sec:SmallN}.

The significant performance gap between MFABL and pMFABL is worth
  noting. We believe that this gap is driven by the fact that pMFABL uses
  ``real'' feedback (i.e.,  consumer converting or not) to update the
  $(\pmb{\alpha}, \pmb{\beta})$ counts by rolling back the observed
  outcome (conversion or not) to the underlying state-action pairs on the
  consumer path.  On the other hand,  MFABL updates the $(\pmb{\alpha},
  \pmb{\beta})$ counts using a``synthetic'' 
  feedback
  $f_{s'}$ generated by sampling from the belief 
  associated with
  action with the highest
  expected value in state $s'$. 
  It is important to realize that this
  $f_{s'}$ sample is not a true data point but an artificially created
  data point that enables MFABL to learn.  Given it is generated using the
  belief at state $s'$ (which is likely to be incorrect in the initial
  stages of learning), the generated feedback might not be very
  informative in the first few rounds, leading to slower learning.
  \cite{peng1994incremental} report similar numerical findings regarding
  the fluid variants of MFABL and pMFABL.

Clearly, pMFABL is able to learn faster than MFABL. However, MFABL comes with convergence guarantees (Theorems \ref{theorem:CFOConvergence} and \ref{theorem:ConRate}), whereas pMFABL does not (discussed further in \S\ref{sec:conc}).  In terms of the practical usage of these algorithms,  can one get the best of both worlds, i.e., fast initial learning \emph{and} convergence guarantees?  Interestingly, this is possible via a hybrid approach, i.e., start with pMFABL and eventually shift to MFABL. Since MFABL's convergence guarantees hold for all prior counts,  such a hybrid approach obeys Theorems \ref{theorem:CFOConvergence} and \ref{theorem:ConRate}.\footnote{As pMFABL did converge to optimality in our numerics, we do not expect to see the value of this hybrid algorithm in our dataset.} Such a win-win approach has a strong practical appeal, but requires the practitioner to decide on when to shift to MFABL.

As discussed in \S\ref{sec:MFABLproperties},  MFABL/pMFABL are
  highly interpretable. We illustrate this in Figure
  \ref{fig:interpret} (\S\ref{sec:AdditionalFigs}), where we show how the belief under pMFABL evolves
  (analogous to Figure \ref{fig:ExampleBandit} for TS). 
  We expect 
  MFABL/pMFABL to be appealing in practice, since they are as interpretable as TS.

We compare the compute times of various approaches in Figure
\ref{fig:TimeVsPR} (\S\ref{sec:AdditionalFigs}).  QL-UCB and TS have the two lowest compute
times. Though MFABL and pMFABL have compute times of the same order as TS,
they significantly boost the PR by a factor of over 2 and 3, respectively,
highlighting the value of capturing long-run consumer dynamics.   
The relatively low compute times of TS, MFABL, and pMFABL are driven by
their simple design. For optimization, one only needs to sample from $A+1$
Beta distributions and pick the highest value and for attribution,  one
needs to update a handful of parameters via interpretable rules. This is
in contrast to the model-based PSRL, which involves solving a large-scale
MDP in its optimization step. 
As such, PSRL has a compute time that is two orders of magnitude higher than the other four approaches,  but still does not learn the optimal
targeting policy, highlighting the importance of scalability in
large-scale conversion funnel optimization.  

It is well-known that PSRL will eventually converge to the optimal policy
\citep{osband2013more}. In fact, when we increase $N$ from $500,000$ to
$3,000,000$,  the average PR of PSRL increases from 0.14 to 0.63. However,
by then, MFABL and pMFABL achieve an even higher average PR (0.77 and
0.88, respectively).\footnote{The average PR of TS remains
  approximately the same to what it was in Figure \ref{fig:PRsnapshot},
  i.e.,  $0.27$.  This is because TS had already converged to the
  myopically optimal policy by the end of $500,000$ consumers.}
Furthermore, it is possible that consumer behavior might change during
this period of interacting with so many consumers, which motivates
understanding the performance of these algorithms under concept shift. 

\subsection{Concept Shift} \label{sec:conceptshift}

As discussed in \S\ref{sec:MFABLproperties},  concept shift refers to the possibility that consumer behavior
(transition probabilities $\cP$) changes 
over time. We conduct the following simple
experiment to understand how our
algorithm operates under concept shift. We set $N=1,000,000$ and split the consumer behavior into two phases.  
In phase 1,  the underlying consumer behavior (transition probabilities $\cP_1$) is the same as in the setup above
and in phase 2,  it changes to $\cP_2$.
To generate $\cP_2$, we randomly
permute the actions in order to ensure
the optimal conversion probability remains the
same -- just the optimal action sequence gets permuted.  Phase 1 corresponds to the first $N/2$ consumers and phase 2
corresponds to consumers $N/2 + 1$ to $N$. 
As before, we perform $R=100$ runs ($\mathcal{P}_2$ in run $r$ can be 
different from $\mathcal{P}_2$ in run $r'$ due to the random
permutation). The decision-making algorithms do not know that a switch
from $\mathcal{P}_1$ to $\mathcal{P}_2$ will happen, or when.  As such, the posterior belief at the end of phase 1 becomes the prior belief at the beginning of phase 2.   Had the algorithms known the time of phase shift,  it would have made sense to reset the beliefs to the original priors at the beginning of phase 2.  Then, the performance in phase 2 would be similar to that in phase 1.  Therefore, the more interesting case is to not let the algorithms know when the phase shift happens.
(We note that our two-phase treatment of concept shift is a
  simplification of the non-stationary RL setting with sliding windows
  \citep{cheung2023nonstationary} or periodic restarting
  \citep{mao2021near}. 
  That being said,
  we apply this simplification to all
  algorithms. Furthermore,  in \S\ref{sec:GradualConcept}, we supplement
  our numerics via additional results pertaining to a ``gradual'' concept
  shift.)

\begin{figure}
\centering
\includegraphics[scale=0.4]{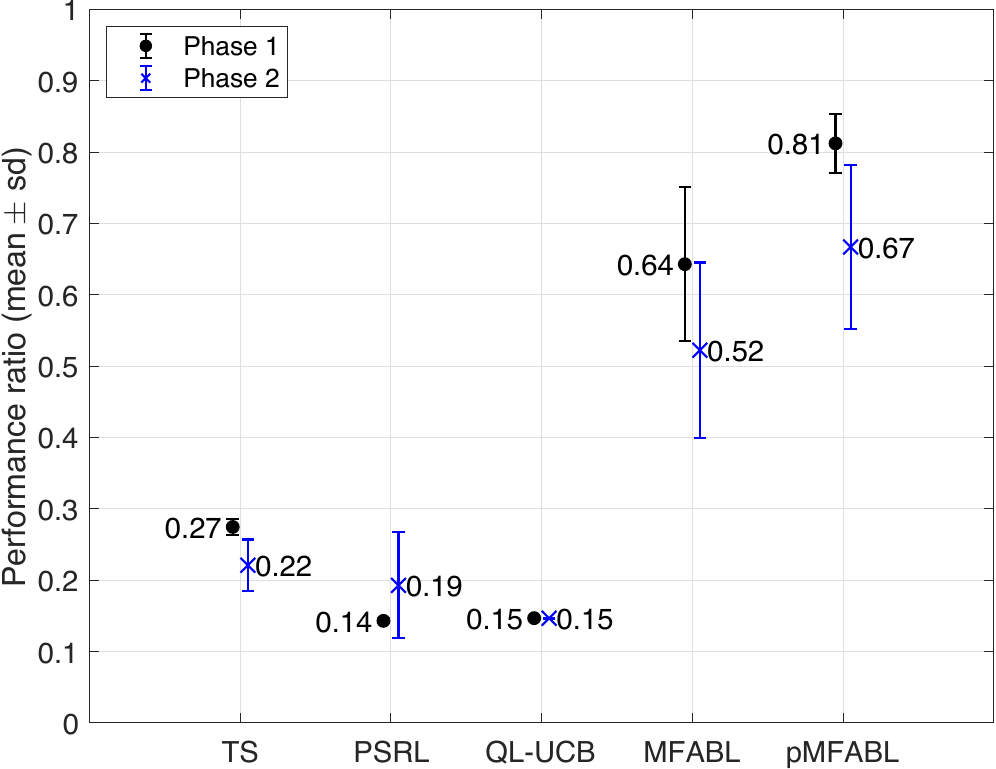}
\caption{\normalfont Phases 1 and 2 performance ratios under various algorithms.  Phase 1 is identical to the setup of Figure \ref{fig:PRsnapshot} whereas phase 2 corresponds to an additional 500,000 consumers after phase 1. }
\label{fig:PRconcept}
\end{figure}

The performance ratios for all algorithms under both phases are shown in Figure~\ref{fig:PRconcept}.  Clearly, MFABL/pMFABL exhibit a
``win-win'' behavior in the sense that they have a higher PR in both phases.  The drop in the PR values of MFABL/pMFABL between the two phases is expected as by the end of phase 1, the posterior beliefs under these algorithms converge to the optimal phase 1 values with very little variance. As a result, they start phase 2 with a very strong but incorrect belief, which takes time to adapt. Similar logic holds for TS, but as before, it converges to the myopically optimal policy in phase 2 as well, which is suboptimal in general.  
Furthermore,  PSRL barely learns in either of the two phases because of
its lack of scalability, a topic we discuss next.

\section{Value of Scalability} \label{sec:scalability}

As noted in \S\ref{sec:BenchmarksAppendix}, to manage the compute time of PSRL, we re-optimized once every 1000 consumers.  Even then, PSRL is much more expensive than other approaches (as seen in Figure \ref{fig:TimeVsPR}).  This lack of scalability of PSRL is driven by the large state space of the underlying MDP (recall $S = 11,060$), which captures the temporal, awareness, and engagement dimensions. 
However, as seen in Table \ref{table:predictionmodels},  it is possible to explain consumer behavior with high accuracy via lower dimensional models. For example,  if we ignore the engagement or the awareness dimensions, then we get an out-of-sample AUC of 0.94 (model 10) or 0.90 (model 11), respectively.
If we ignore both, we still get an out-of-sample AUC of 0.85 (model 7).  Compared to $11,060$ states under model 12, models 7, 10, and 11 have 14, 810, and 420 states, respectively.  Thus, under these models, PSRL can possibly exhibit manageable compute times with a higher re-optimization/learning frequency (e.g., once every 10 or 100 consumers).
Note that as discussed in \S\ref{sec:optimization},  the firm might not know the ``true'' state space apriori and working with a lower dimensional state space can result in model misspecification.
As such,  we evaluate the performance of PSRL when it learns over such lower dimensions (models 7, 10, and 11) but the ``true'' consumer behavior is governed by the higher dimensional model 12.  On the surface, this seems promising as these lower dimensional models explain consumer behavior almost as well as model 12 but can enable PSRL to learn faster (via a higher re-optimization frequency and since there are fewer parameters to learn).   However,  as we illustrate next,  our proposed MFABL approach still outperforms PSRL, highlighting the value MFABL brings by being scalable to large-scale models.

\begin{figure}
\centering
\includegraphics[scale=0.4]{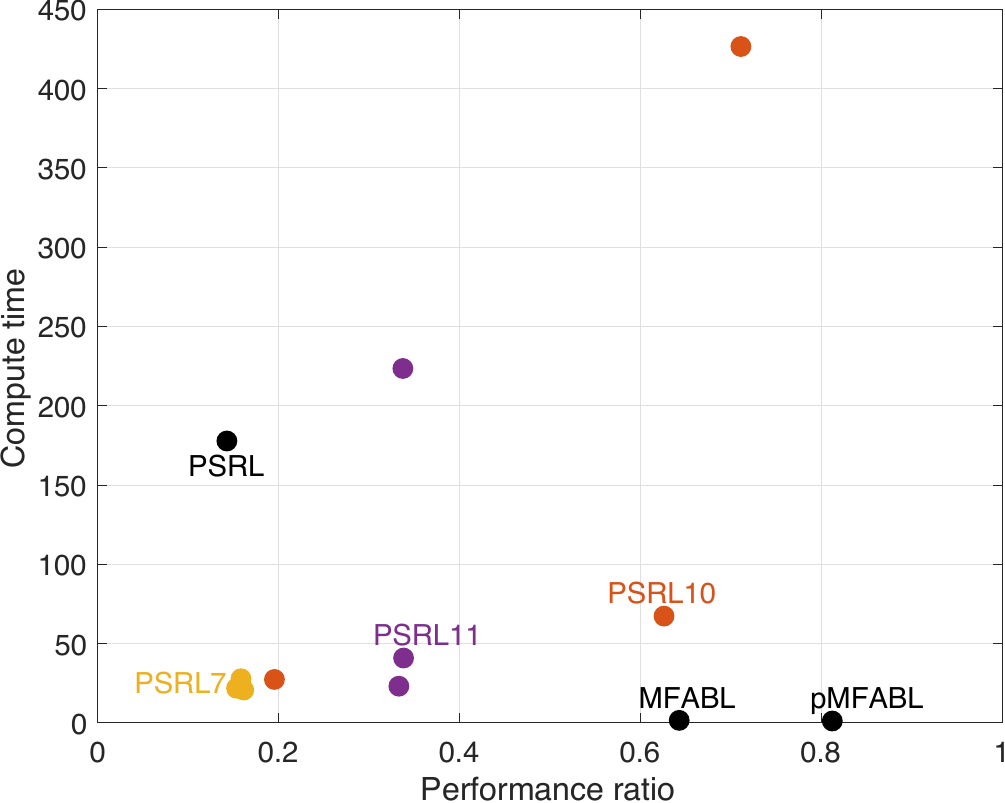}
\caption{\normalfont Compute time versus performance ratio for $N = 500,000$ for various possibilities under PSRL.  There are three black dots (PSRL, MFABL, and pMFABL) that are identical to the dots in Figure \ref{fig:TimeVsPR}.  Compared to Figure \ref{fig:TimeVsPR},  there are nine new dots: three in yellow (PSRL7),   three in orange (PSRL10), and three in purple (PSRL11). PSRL10 corresponds to when PSRL learns over a state space corresponding to model 10 (from Table \ref{table:predictionmodels}) but the consumer behavior being driven by model 12. 
The three dots under PSRL10 correspond to different re-optimization frequencies: once every 10, 100, and 1000 consumers.  The more frequently we re-optimize,  the faster PSRL learns but at the expense of a higher compute time.
PSRL7 and PSRL11 have similar interpretations.}
\label{fig:TimeVsPRMisfit}
\end{figure}

We show the results in Figure \ref{fig:TimeVsPRMisfit}, which is analogous to Figure \ref{fig:TimeVsPR}. The three black dots (PSRL, MFABL, and pMFABL) are as in Figure \ref{fig:TimeVsPR} and there are nine new dots,  with three each for models 7 (PSRL7), 10 (PSRL10), and 11 (PSRL11) (detailed description in the caption).  
We discuss the results corresponding to the three models sequentially.
First, consider PSRL7, i.e., the three yellow dots.  PSRL7 denotes the case when PSRL learns over the state space of model 7 ($S = 14$) but the underlying consumer behavior is governed by the much higher dimensional model 12 ($S = 11,060$). Though model 7 explains consumer behavior reasonably well (AUC of 0.85 versus AUC of 0.96 under model 12),  the 0.11 difference in AUC results in a major model misspecification.  In particular, PSRL converges to a suboptimal policy with an average PR of 0.16.  MFABL and pMFABL achieve a PR of approximately 4x and 5x respectively, and that too with less compute time.   Note that the re-optimization frequency (once every 10, 100, or 1000 consumers) does not play a role in PSRL7 since the number of parameters is small enough such that a frequency of 1000 enables PSRL to converge. This can be seen in Figure \ref{fig:TimeVsPRMisfit} as all of the three yellow dots are clustered around a PR value of approximately 0.16. Furthermore, given the small value of $S=14$, the compute time does not change drastically as we vary the re-optimization frequency since the cost of optimization is much smaller than the other costs related to storage and parameter updates.

Second, consider PSRL11, i.e., the three purple dots in Figure \ref{fig:TimeVsPRMisfit}.  Model 11 corresponds to the state space capturing all dimensions except awareness with $S = 420$ states.  
In addition to the temporal and email type dimensions captured in model 7,  model 11 captures the engagement dimension, which enables it to better explain consumer behavior (AUC of 0.90 versus 0.85). This 0.05 improvement in AUC improves the PR by a factor of over 2 (from 0.16 to 0.34), illustrating how small improvements in the consumer behavior model can result in big improvements in conversion rates.
However,  despite its high AUC and similar to model 7, model 11 also suffers from model misspecification as PSRL converges to a suboptimal policy, with MFABL and pMFABL improving the PR by a factor of over 1.9 and 2.4, respectively.
As with PSRL7, the re-optimization frequency does not play a role in the PR of PSRL11 as the number of parameters is small enough such that a frequency of 1000 enables PSRL to converge. This can be seen in Figure \ref{fig:TimeVsPRMisfit} as all of the three purple dots map to a PR value of approximately 0.34.   However, given the not-so-small value of $S=420$, the compute time increases as we re-optimize more frequently.

Third,  consider PSRL10, i.e., the three orange dots in Figure \ref{fig:TimeVsPRMisfit}.  Model 10 corresponds to the state space capturing all dimensions except engagement with $S = 810$ states, which is higher than the 420 states under model 11.  Model 10 explains consumer behavior even better than model 11 (AUC of 0.94 versus 0.90).  Despite this improved AUC,  PSRL10 achieves a much smaller PR than PSRL11 when the re-optimization frequency is 1000 (PR of 0.2 versus 0.34). This is because under model 10, there are much more parameters to be learned.  However, as we re-optimize more frequently, PSRL10 beats PSRL11. 
In particular, PSRL10 achieves a PR of 0.63 and 0.71 when we re-optimize once every 100 and 10 consumers, respectively. Of course, this comes at a high computational cost.  Furthermore, pMFABL outperforms PSRL10, and MFABL is comparable to PSRL10\footnote{Observe from Figure \ref{fig:PRsnapshot} that 0.71 is within one standard deviation of MFABL's PR.}, both at a fraction of the compute time.  
Of course, this is not an apples-to-apple comparison since model 10 is almost as good as model 12, and we let PSRL learn over the much smaller state space of model 10 as compared to MFABL/pMFABL learning over the state space of model 12. A better comparison might be to run MFABL/pMFABL on the state space of model 10, but the point to emphasize here is that MFABL/pMFABL seamlessly scales to large-scale state spaces,  giving the firm a high degree of flexibility in modeling consumer behavior.

For completeness, we also ran MFABL/pMFABL on the smaller state spaces
implied by models 7, 10, and 11. In these models, 
MFABL achieved a PR of 0.22, 0.55, and 0.31 respectively. 
This performance is 
comparable to that of PSRL; however,
the compute time of MFABL is significantly lower than that of PSRL. 
The PRs being comparable is not surprising since we expect
PSRL to perform well for not-so-large state spaces.   
Nonetheless, it seems that MFABL requires an accurate model specification to maintain its high performance, and real-world problems are likely to suffer from model misspecification.  This can be seen as a potential limitation of MFABL\footnote{Though model misspecification is likely to be the case in practice, MFABL's model-free nature allows it to operate on much larger state spaces and hence,  allows one to reduce the ``misspecification gap'' (contrast this with the model-based PSRL).} but we note that handling model misspecification is a non-trivial challenge.
Surprisingly, pMFABL significantly outperformed PSRL, achieving a PR of
0.75, 0.78, and 0.82 with models 7, 10, and 11, respectively.  
We posit that this robustness of pMFABL is driven by its model-free nature
coupled with its ability to learn from ``real'' data (recall the 
discussion around ``synthetic'' vs.\ ``real'' feedback in
\S\ref{sec:results}). 

The key message here is MFABL/pMFABL bring a lot of value by being scalable to large-scale models.  As discussed in \S\ref{sec:optimization},  the ``true'' consumer behavior is unknown apriori and can be either low or high dimensional.   Incorrectly assuming the behavior to be low dimensional (e.g., PSRL7 and PSRL11) can lead to suboptimal marketing actions, which highlights the need for scalability.

\section{Concluding Remarks} \label{sec:conc}
We study the problem of optimal sequential personalized
interventions for a firm promoting a product under a
fairly general conversion funnel model for consumer behavior. 
Our state-based MDP explains the journey of a
consumer from her initial interactions with the
firm till the time she makes a decision to purchase or not. 
Consumer behavior is driven by the sequential interventions the
firm makes. The effect of each intervention on consumer behavior is allowed
to be state-specific and the firm does not know such effects apriori.
Though our framework is general, we demonstrate its application 
on a real-world email marketing dataset.  
We show how one can use observable consumer-level data to construct 
the state space of our model,  and that such micro-level data explains 
consumer behavior to very high accuracy,  especially when one captures 
non-linear interaction effects via boosted trees.
Recognizing that the obtained state space results in a large-scale learning
problem, we exploit the terminal reward structure of the MDP to design a novel
attribution-based decision-making algorithm, which is simple and highly scalable.
We establish the asymptotic optimality of our algorithm and analyze its convergence rate. We also demonstrate its
value by benchmarking it with state-of-the-art algorithms
on large-scale simulations calibrated to our email marketing dataset.

We believe there is immense potential for learning-based
approaches to personalized marketing and we highlight some
research directions. In our conversion funnel model, consumer behavior
is captured via an observable 
state and the firm's decisions are state-dependent.  
Although our framework is general enough
to capture an arbitrary (finite) state space, constructing an appropriate
state space is an interesting research avenue. 
In our application, we showed how a data-driven approach that maps observable
consumer-level data to a state space explains consumer behavior with
high accuracy.  However,  such an approach can lead to a very high dimensional
state space.  Though our decision-making algorithm scales gracefully, it is naturally of
interest to find the lowest dimensional state space that explains consumer behavior.
We note that this would very much vary between firms
and industries and constructing such context-specific state space is an
active area of research \citep{singh2003learning,  bennouna2021learning}. 
  
Our work focuses on the limited consumer features regime.  An extension 
to a setting where consumers can have
high-dimensional features (and the firm can observe such information)
would be of interest.  Works such as
\cite{agrawal2013thompson}
handle high-dimensional consumer features, but they optimize for a myopic
reward function. There appears to be some progress in modeling
high-dimensional contextual information along with long-run rewards (see
\cite{hallak2015contextual}). We believe the idea of ``value function
approximation'' from the reinforcement learning literature
\citep{sutton2018reinforcement} can be leveraged here,
especially if the value function is parameterized using domain expertise
from marketing. Such approximations can allow efficient learning across
consumers and some very recent works in sequential marketing 
leverage such techniques \citep{liu2022dynamic}.

Given the available data, we limited our action set to $4$ classes of
emails.  We do not address the question of the optimal number of classes of emails and the optimal design of emails in each class.  The design of the action space is an interesting question,  especially for the email marketing application.  
In this work, we address the problem of state-dependent action selection once the action is defined.

From a theoretical perspective,  in addition to understanding the convergence rate, it is important to understand the regret of our algorithm. Given the connection of our algorithm to Q-learning (which we used while proving its asymptotic convergence and deriving the convergence rate), recent works such as \cite{wainwright2019stochastic} and \cite{li2020sample} that establish theoretical guarantees for Q-learning can possibly be leveraged.  It is also worthwhile to study whether there is another way to prove the convergence.

Finally,  we would like to provide convergence guarantees
for pMFABL, which outperforms MFABL in our numerics.
However, analyzing pMFABL is likely to be
challenging.  This is because pMFABL
does not correspond to the standard stochastic approximation scheme that
model-free approaches such as Q-learning do
\citep{tsitsiklis1994asynchronous}.  We are able to establish the
convergence of MFABL by connecting its ``fluid'' variant to Q-learning. As
such, a natural step to understand pMFABL is to analyze its ``fluid''
variant, which is similar to modifications of Q-learning such as Watkin's
Q($\lambda$) \citep{watkins1989learning} and incremental multi-step
Q-learning \citep{peng1994incremental}. To the best of our
knowledge\footnote{See Remark 12.10 in
    \cite{sutton2018reinforcement} and \S6 in
    \cite{peng1994incremental}.}, the convergence of these modifications has
not been proven.   In fact, \cite{peng1994incremental} report similar
numerical results regarding such modifications performing better than Q-learning.   

\ACKNOWLEDGMENT{We thank the POM review team (DE, SE, and 3 referees) for their careful reading of our work and enabling us to write a better paper, and the following colleagues for useful comments: Asim Ansari, Raghuram Iyengar, and Praveen Kopalle.}

\bibliographystyle{informs2014}
\bibliography{biblio}

\ECSwitch
\begin{center} 
\bold{\large Model-Free Approximate Bayesian Learning \\ for Large-Scale Conversion Funnel Optimization (E-Companion)} \\
\bold{Garud Iyengar} and \bold{Raghav Singal}
\end{center}

\section{Implementation Details for \texttt{xgboost}} \label{sec:xgboostdetails}
We use the \texttt{xgboost} package \citep{Chen2016} in the \texttt{R}
programming language to fit the predictive models discussed in
\S\ref{sec:behavior}. 
We discuss the implementation details here. Our dataset consists of a few million consumer paths, and each path contains 14 data points (one data point per day and recall $T=14$).  To handle the scale of this data, we randomly
create a $10\%$ subset, which still contains a few million data points.  
Recall the output variable $y_{nt}$ denotes consumer $n$ behavior following the email on day $t$ and it can take 4 possible values:
ignore, open, click, convert.
Out of the few million data points,  approximately 46000, 1600, and 1100
correspond to the consumer behavior of opening, clicking, and converting,
respectively.   

For each of the 12 models discussed in \S\ref{sec:behavior}, we use the
cross-validation functionality \texttt{xgb.cv} with the following input
parameters. We set \texttt{nrounds} = 500, which denotes the number of
rounds for boosting. This is the parameter we tune in cross-validation. We
set \texttt{nfold} = 5 (number of folds to use for cross-validation),
\texttt{max\_depth} = 10 (tree depth),  \texttt{eta} = 0.1 (learning
rate), \texttt{objective} = ``multi:softprob'' (since the output is
multiclass), and \texttt{num\_class} = 4 (since there are 4 output
classes). Moreover, to prevent overfitting, we set \texttt{maximize} =
``FALSE'' and \texttt{early\_stopping\_rounds} = 1 with ``mlogloss''
(multiclass log-loss) as the \texttt{metrics} (defined below). That
is,  we let the model 
train until the multiclass log-loss (on test data) did not improve in 1
round. 
We used cross-validation 
to select the 
value (lowest multiclass
log-loss in test data) for \texttt{nrounds}.  It varied between 50 and 150
depending on the model,  and hence, our choice of 500 above was big
enough.   
This act of selecting the best \texttt{nrounds} may bias the
metrics reported in Table \ref{table:predictionmodels}. To overcome this
bias, we held out another $5\%$ of the data (in addition to the
$10\%$ subset mentioned above) and evaluated the two metrics (AUC and
log-loss) on this ``holdout'' set for each of the 12 models (fitted on
the $10\%$ subset using the best \texttt{nrounds}).  For each model,  we
found the AUC on the holdout set to be within $\pm 1\%$ of the
corresponding AUC reported in Table \ref{table:predictionmodels} and the
same for log-loss as well.  This suggests 
that overfitting is not an issue here. 
Note that when we simulate the ground truth consumer behavior (via model 12) for the
numerical experiments discussed in 
\S\ref{sec:numerics}, 
we use the optimal
\texttt{nrounds} to fit the model on all of the few million data points (the $10\%$ subset)
via the \texttt{xgboost} function (with the same values for other
parameters). 

For completeness, 
we
include
the definitions
for the two metrics: (1) multiclass AUC and (2) multiclass log-loss.  In
the case of a binary prediction, AUC and log-loss are standard
metrics.  However,  
we have
multiple classes (4 in particular), and this
introduces complexity. We use the 
multiclass extensions of
AUC and log-loss that come as the default with \texttt{xgboost}.
For
multiclass AUC,  \texttt{xgboost} computes the AUC of each of the 4
classes separately and takes a weighted average of the 4 AUCs (weighted
by the frequency of the class prevalence). Even though it takes the
weighted average, we often found the 4 AUCs to be close to each
other. For instance, for model 12, we report a multiclass AUC of 0.96 in
Table \ref{table:predictionmodels}, which is a weighted average of 4
AUCs, with the lowest being 0.93. This implies that our models do a good
job of explaining \emph{each} of the 4 possible consumer actions
(ignore, open, click,  and convert).

Let $z_{ij} \in \{0,1\}$ denote the indicator variable for whether data point
$i$ is in class $j$, 
and let $p_{ij}$ denote the corresponding predicted probability. Then  multiclass
log-loss in \texttt{xgboost}  is defined as follows: 
\begin{align*}
-\frac{1}{N} \sum_{i} \sum_{j} z_{ij} \log(p_{ij}).
\end{align*}

\section{Proof of Theorem~\ref{theorem:CFOConvergence}} \label{sec:CFOConvergenceProof}
We prove Theorem~\ref{theorem:CFOConvergence} using ideas from 
stochastic approximation.  We provide a brief primer in
\S\ref{sec:CFOTsitsiklis}, establish a few supporting lemmas in
\S\ref{sec:CFOConvergenceLemmas}, and invoke them to prove
Theorem~\ref{theorem:CFOConvergence} in
\S\ref{sec:CFOConvergenceProofSub}.  

\subsection{Primer on Asynchronous Stochastic Approximation} \label{sec:CFOTsitsiklis}
The contents of this subsection are based on
\citet{tsitsiklis1994asynchronous}. To help the reader connect these results to our setup, we
alter some of the notations in \citet{tsitsiklis1994asynchronous} so that
it matches our notations.  Here, we only
present the results 
that are
relevant to us, and 
refer the reader to
\citet{tsitsiklis1994asynchronous} for further details. 

Observe that the optimal $Q$-value vector $\mathbf{Q}^* := [Q^*_{sa}]_{(s,a) \in \mathbb{S} \times \mathbb{A}}$ for the conversion funnel is \textit{the} unique solution to the following system of equations \citep{bertsekas1995dynamic, puterman2014markov}:
\begin{align}
Q^*_{sa} = \sum_{s' \in \mathbb{S}^+} p_{sas'} \Big(\max_{a' \in \mathbb{A}} Q^*_{s'a'}\Big) \quad \forall (s,a) \in \mathbb{S} \times \mathbb{A},
\label{eq:CFOQStarEquations}
\end{align}
where we set $Q^*_{ca} = 1$ and $Q^*_{qa} = 0$ for all $a \in \mathbb{A}$. For ease of notation, we will denote the RHS of \eqref{eq:CFOQStarEquations} by $F_{sa}(\mathbf{Q}^*)$ and hence, we get:
\begin{align}
Q^*_{sa} = F_{sa}(\mathbf{Q}^*) \quad \forall (s,a) \in \mathbb{S} \times \mathbb{A}.
\label{eq:CFOQStarEquationsCompact}
\end{align}

An \textit{asynchronous stochastic approximation} scheme to solve the system of equations~\eqref{eq:CFOQStarEquationsCompact} is defined as
follows.  We initialize $X_{sa}(1)$ arbitrarily in $[0,1]$ for all $(s,a) \in \mathbb{S}
\times \mathbb{A}$, and set $X_{ca}(1) = 1$ and $X_{qa}(1) = 0$ for all $a \in
\mathbb{A}$. 
The scheme iteratively generates approximate solutions as follows:
\begin{align}
X_{sa}(t+1) \leftarrow \begin{cases}
X_{sa}(t) &\text{ if } t \notin \mathbb{T}_{sa} \\
X_{sa}(t) + \kappa_{sa}(t) \left( F_{sa}(\mathbf{X}(t)) - X_{sa}(t) +
  w_{sa}(t)  \right) &\text{ if } t \in \mathbb{T}_{sa},
\end{cases}
\label{eq:CFOASAUpdate}
\end{align}
where $\mathbb{T}_{sa} \subseteq \{1,2,\ldots\}$ denotes the set of
iterations when value $X_{sa}(t)$ corresponding to $(s,a) \in
\mathbb{S} \times \mathbb{A}$ is updated, $\kappa_{sa}(t) \in [0,1]$ is the
step size 
and $w_{sa}(t)$ is a noise term. For $t \in \{1, 2, \ldots\}$, the information set $\mathcal{F}(t)$ captures the history of the algorithm till the time
the stepsizes $\kappa_{sa}(t)$ are selected, but does not include the noise
information $w_{sa}(t)$. 
Theorem 3 in \citet{tsitsiklis1994asynchronous} implies the following
result. 
\begin{restatable}{proposition}{CFOASAConvergence} \label{prop:CFOASAConvergence}
Under Assumption~\ref{ass:CFOAbsorption}, $\mathbf{X}(t) \to \mathbf{Q}^*$ w.p.\ 1 as $t \to \infty$ if the following conditions hold:
\begin{enumerate}
\item For every $(s,a) \in \mathbb{S} \times \mathbb{A}$, $t \in \{1,2,\ldots\}$, $w_{sa}(t)$ is $\mathcal{F}(t+1)$-measurable.
\item For every $(s,a) \in \mathbb{S} \times \mathbb{A}$, $t \in \{1,2,\ldots\}$, $\kappa_{sa}(t)$ is $\mathcal{F}(t)$-measurable.
\item For every $(s,a) \in \mathbb{S} \times \mathbb{A}$, $t \in \{1,2,\ldots\}$, $\mathbb{E}[w_{sa}(t) | \mathcal{F}(t)] = 0$.
\item There exist deterministic constants $A$ and $B$ such that for every $(s,a) \in \mathbb{S} \times \mathbb{A}$, $t \in \{1,2,\ldots\}$, we have
$$\mathbb{E}[w_{sa}^2(t) | \mathcal{F}(t)] \le A + B \max_{(s',a')} \max_{\tau \le t} |Q_{s'a'}(\tau)|^2.$$
\item For every $(s,a) \in \mathbb{S} \times \mathbb{A}$, 
\begin{align*}
\sum_{t=0}^{\infty} \kappa_{sa}(t) &= \infty \text{ w.p.\ 1} \\
\sum_{t=0}^{\infty} \kappa_{sa}^2(t) &< \infty \text{ w.p.\ 1.}
\end{align*}
\end{enumerate}
\end{restatable}
Theorem 3 in
\citet{tsitsiklis1994asynchronous} relies on Assumptions
1, 2, 3, and 5 as stated in \citet{tsitsiklis1994asynchronous}. 
Assumption~1 in 
\citet{tsitsiklis1994asynchronous}  is trivially satisfied by our
asynchronous stochastic approximation. 
The
conditions
in Proposition~\ref{prop:CFOASAConvergence} cover Assumptions 2 and 3 of
\citet{tsitsiklis1994asynchronous}.  Finally, our absorption assumption
implies Assumption~5 of \citet{tsitsiklis1994asynchronous} (see the
discussion above Theorem 4 in \citet{tsitsiklis1994asynchronous}). 

\subsection{Supporting Lemmas} \label{sec:CFOConvergenceLemmas}
We now establish a few supporting lemmas. In particular, we show that the \textit{expected} update in our MFABL algorithm is an instance of the asynchronous stochastic approximation scheme above. 

Consider an arbitrary iteration\footnote{By ``iteration'', we refer to specific $(t,n)$ pair in Algorithm~\ref{alg:CFOOurAlgo}.} $i$ in Algorithm~\ref{alg:CFOOurAlgo} such that the corresponding consumer was in state $s \in \mathbb{S}$, the firm took action $a \in \mathbb{A}$, and the consumer transitioned to state $s' \in \mathbb{S}^+$. Denote by $Q_{sa}(i) \stackrel{d}{=} \text{Beta}(\alpha_{sa}(i), \beta_{sa}(i))$ the belief over the value of $(s,a)$ \textit{before} the update and by $Q_{sa}(i+1) \stackrel{d}{=} \text{Beta}(\alpha_{sa}(i+1), \beta_{sa}(i+1))$ the belief \textit{after} the update. Denote by $\overline{Q}_{sa}(i)$ and $\overline{Q}_{sa}(i+1)$ the expected values of $Q_{sa}(i)$ and $Q_{sa}(i+1)$, respectively. Finally, define $n_{sa}(i) := \alpha_{sa}(i) + \beta_{sa}(i)$ and $n_{sa}(i+1) := \alpha_{sa}(i+1) + \beta_{sa}(i+1)$.
Recall that $f_{s'}$  denotes the feedback generated from state $s'$:
\begin{align*}
f_{s'} \sim \text{Bernoulli}\left( \max_{a' \in \mathbb{A}} \frac{\alpha_{s'a'}(i)}{\alpha_{s'a'}(i) + \beta_{s'a'}(i)}  \right).
\end{align*}
The following lemma characterizes the \textit{expected} update.

\begin{restatable}{lemma}{CFOExpectedUpdate} \label{lemma:CFOExpectedUpdate}
The expected update obeys the following equation:
\begin{align*}
\overline{Q}_{sa}(i+1) = \frac{n_{sa}(i)}{n_{sa}(i) + 1} \overline{Q}_{sa}(i) + \frac{1}{n_{sa}(i) + 1} f_{s'}.
\end{align*}
\end{restatable}

\bold{Proof.} We split the proof into two parts: (1) $f_{s'} = 0$ and (2) $f_{s'} = 1$.

\noindent \bold{Case 1.} If $f_{s'} = 0$, MFABL increases $\beta_{sa}$ by 1, i.e., 
\begin{align*}
\alpha_{sa}(i+1) &= \alpha_{sa}(i) \\
\beta_{sa}(i+1) &= \beta_{sa}(i) + 1.
\end{align*}
Before the update, the expected value of $Q_{sa}(i) \stackrel{d}{=} \text{Beta}(\alpha_{sa}(i), \beta_{sa}(i))$ equals
\begin{align*}
\overline{Q}_{sa}(i) = \frac{\alpha_{sa}(i)}{n_{sa}(i)}.
\end{align*} 
After the update, the expected value of $Q_{sa}(i+1) \stackrel{d}{=} \text{Beta}(\alpha_{sa}(i+1), \beta_{sa}(i+1))$ equals
\begin{align*}
\overline{Q}_{sa}(i+1) &= \frac{\alpha_{sa}(i+1)}{n_{sa}(i+1)} \\
   &= \frac{\alpha_{sa}(i)}{n_{sa}(i) + 1}  \\
   &= \frac{n_{sa}(i)}{n_{sa}(i) + 1} \frac{\alpha_{sa}(i)}{n_{sa}(i)} + \frac{1}{n_{sa}(i) + 1} 0 \\
   &= \frac{n_{sa}(i)}{n_{sa}(i) + 1} \overline{Q}_{sa}(i) + \frac{1}{n_{sa}(i) + 1} 0 \\
   &= \frac{n_{sa}(i)}{n_{sa}(i) + 1} \overline{Q}_{sa}(i) + \frac{1}{n_{sa}(i) + 1} f_{s'}.
\end{align*}

\noindent \bold{Case 2.} If $f_{s'} = 1$, MFABL increases $\alpha_{sa}$ by 1, i.e., 
\begin{align*}
\alpha_{sa}(i+1) &= \alpha_{sa}(i) + 1 \\
\beta_{sa}(i+1) &= \beta_{sa}(i).
\end{align*}
Before the update, the expected value of $Q_{sa}(i) \stackrel{d}{=} \text{Beta}(\alpha_{sa}(i), \beta_{sa}(i))$ equals
\begin{align*}
\overline{Q}_{sa}(i) = \frac{\alpha_{sa}(i)}{n_{sa}(i)}.
\end{align*} 
After the update, the expected value of $Q_{sa}(i+1) \stackrel{d}{=} \text{Beta}(\alpha_{sa}(i+1), \beta_{sa}(i+1))$ equals
\begin{align*}
\overline{Q}_{sa}(i+1) &= \frac{\alpha_{sa}(i+1)}{n_{sa}(i+1)} \\
   &= \frac{\alpha_{sa}(i) + 1}{n_{sa}(i) + 1}  \\
   &= \frac{\alpha_{sa}(i)}{n_{sa}(i) + 1} + \frac{1}{n_{sa}(i) + 1} \\
   &= \frac{n_{sa}(i)}{n_{sa}(i) + 1} \frac{\alpha_{sa}(i)}{n_{sa}(i)} + \frac{1}{n_{sa}(i) + 1} \\
   &= \frac{n_{sa}(i)}{n_{sa}(i) + 1} \overline{Q}_{sa}(i) + \frac{1}{n_{sa}(i) + 1} f_{s'}.
\end{align*}
This completes the proof.
\hfill $\Halmos$

\leaveline

In iteration $i$, MFABL updates belief $Q_{sa}(i)$ to $Q_{sa}(i+1)$ by
updating the corresponding parameters $\alpha_{sa}(i)$ and
$\beta_{sa}(i)$ to $\alpha_{sa}(i+1)$ and $\beta_{sa}(i+1)$. The belief
over the value of all other state-action pairs $(s',a') \neq (s,a)$
remains unchanged. This process repeats for multiple iterations and the
corresponding process is denoted by $\{\mathbf{Q}(i)\}_{i}$ where
$\mathbf{Q}(i) := [Q_{sa}(i)]_{(s,a) \in \mathbb{S} \times \mathbb{A}}$ for
all $i$. Let $\overline{\mathbf{Q}}(i) = \mathbb{E}[\mathbf{Q}(i)]$ for all $i
\geq 1$.

\begin{restatable}{lemma}{CFOExpectedUpdateIsASA} \label{lemma:CFOExpectedUpdateIsASA} 
The process $\{\overline{\mathbf{Q}}(i)\}_{i}$ is an asynchronous
stochastic approximation scheme with respect to the system of equations
\eqref{eq:CFOQStarEquationsCompact}. 
\end{restatable}

\bold{Proof.} Given any prior counts as an input to MFABL, $\overline{Q}_{sa}(1) \in [0,1]$ for all $(s,a) \in \mathbb{S} \times \mathbb{A}$. Line 1 of Algorithm~\ref{alg:CFOOurAlgo} ensures $\overline{Q}_{ca}(1) = 1$ and $\overline{Q}_{qa}(1) = 0$  for all $a \in \mathbb{A}$. Lemma~\ref{lemma:CFOExpectedUpdate} implies \eqref{eq:CFOASAUpdate} is satisfied with the stepsize parameter equal to
$
\kappa_{sa}(i) = \frac{1}{n_{sa}(i) + 1},
$
which is in [0,1] since $n_{sa}(i) \ge 0$ for all $(s,a) \in \mathbb{S} \times \mathbb{A}$ and for all $i$. To see this, observe that
\begin{align*}
\overline{Q}_{sa}(i+1) &= \frac{n_{sa}(i)}{n_{sa}(i) + 1} \overline{Q}_{sa}(i) + \frac{1}{n_{sa}(i) + 1} f_{s'} \\
   &= \overline{Q}_{sa}(i) + \frac{1}{n_{sa}(i) + 1} \left( f_{s'} - \overline{Q}_{sa}(i) \right) \\
   &= \overline{Q}_{sa}(i) + \frac{1}{n_{sa}(i) + 1}  \left( F_{sa}(\mathbf{\overline{Q}}(i)) - \overline{Q}_{sa}(i)  + f_{s'} - F_{sa}(\mathbf{\overline{Q}}(i)) \right) \\
   &= \overline{Q}_{sa}(i) + \frac{1}{n_{sa}(i) + 1}  \left( F_{sa}(\mathbf{\overline{Q}}(i)) - \overline{Q}_{sa}(i)  + w_{sa}(i)  \right),
\end{align*}
where $w_{sa}(i)  := f_{s'} - F_{sa}(\mathbf{\overline{Q}}(i))$ represents the noise term:
\begin{align*}
\mathbb{E}\left[w_{sa}(i) | \mathcal{F}(i) \right] &= \mathbb{E}\left[f_{s'} - F_{sa}(\mathbf{\overline{Q}}(i)) | \mathcal{F}(i) \right]  \\
   &=  \mathbb{E}\left[f_{s'} | \mathcal{F}(i) \right] -  \mathbb{E}\left[F_{sa}(\mathbf{\overline{Q}}(i)) | \mathcal{F}(i) \right]  \\
   &= \mathbb{E}_{s'} \left[ \text{Bernoulli}\left( \max_{a' \in \mathbb{A}} \frac{\alpha_{s'a'}(i)}{\alpha_{s'a'}(i) + \beta_{s'a'}(i)} \right) \Bigg| a \right] - \sum_{s' \in \mathbb{S}^+} p_{sas'} \max_{a' \in \mathbb{A}} \overline{Q}_{s'a'}(i) \\
   &= \sum_{s' \in \mathbb{S}^+} p_{sas'} \max_{a' \in \mathbb{A}} \overline{Q}_{s'a'}(i) - \sum_{s' \in \mathbb{S}^+} p_{sas'} \max_{a' \in \mathbb{A}} \overline{Q}_{s'a'}(i) \\
   &= 0.
\end{align*}
Note that $\mathcal{F}(i)$ includes the information that MFABL played action $a$ in iteration $i$, but it does not include the information the consumer transitioned to state $s'$. This completes the proof.
\hfill $\Halmos$

\subsection{Proof of Theorem~\ref{theorem:CFOConvergence}} \label{sec:CFOConvergenceProofSub}
To prove Theorem~\ref{theorem:CFOConvergence}, we first establish that the
expectation counterpart $\overline{\mathbf{Q}}(i)$ converges to
$\mathbf{Q^*}$ w.p.\ 1 as $i$ goes to infinity. Then, we show that the
variance of the Beta belief $\mathbf{Q}(i)$ goes to zero as $i$ goes to
infinity, and hence, $\mathbf{Q}(i)$ converges to $\mathbf{Q^*}$. The
following lemma claims the first part. 

\begin{restatable}{lemma}{CFOExpectedUpdateIsConverges} \label{lemma:CFOExpectedUpdateIsConverges}
Under Assumption~\ref{ass:CFOAbsorption}, $\overline{\mathbf{Q}}(i)$
converges to $\mathbf{Q}^*$ w.p.\ 1 as $i \to \infty$. 
\end{restatable}

\bold{Proof.} Given Lemma~\ref{lemma:CFOExpectedUpdateIsASA}, it suffices to show that the conditions in Proposition~\ref{prop:CFOASAConvergence} are satisfied. For condition 1, as shown in the proof of Lemma~\ref{prop:CFOASAConvergence}, the noise term equals $w_{sa}(i) = f_{s'} - F_{sa}(\mathbf{\overline{Q}}(i))$, which is $\mathcal{F}(i+1)$-measurable for every $(s,a) \in \mathbb{S} \times \mathbb{A}$, $i \in \{1,2,\ldots\}$. From the proof of Lemma~\ref{prop:CFOASAConvergence}, the stepsize equals $\kappa_{sa}(i) = \frac{1}{n_{sa}(i) + 1}$, which is $\mathcal{F}(i)$-measurable for every $(s,a) \in \mathbb{S} \times \mathbb{A}$, $i \in \{1,2,\ldots\}$. We verified condition 3 (noise is mean-zero) in the proof of Lemma~\ref{prop:CFOASAConvergence}. For condition 4, note that $A=1$ and $B=0$ works since for all $(s,a) \in \mathbb{S} \times \mathbb{A}$, $i \in \{1,2,\ldots\}$, and $s' \in \mathbb{S}^+$, we have
\begin{align*}
w_{sa}^2(i) &= \left( f_{s'} - F_{sa}(\mathbf{\overline{Q}}(i)) \right)^2 \\
    &= \left( f_{s'} - \sum_{s' \in \mathbb{S}^+} p_{sas'} \max_{a' \in \mathbb{A}} \overline{Q}_{s'a'}(i) \right)^2 \\
    &\le 1.
\end{align*}
Final inequality is true because $f_{s'} \in [0,1]$ and $\overline{Q}_{s'a'}(i) \in [0,1]$. Finally, condition 5 is true because the stepsize sequence corresponding to $\kappa_{sa}(i) = \frac{1}{n_{sa}(i) + 1}$ forms a harmonic series and each state-action pair is visited infinitely often due to the $\epsilon$-greedy construction of MFABL and the ``connectedness'' assumption (recall the discussion when defining initial state probabilities in \S\ref{sec:model}).
\hfill  $\Halmos$

\leaveline

\bold{Proof of Theorem~\ref{theorem:CFOConvergence}.} Given Lemma~\ref{lemma:CFOExpectedUpdateIsConverges}, it suffices to show that the variance of the Beta belief $\mathbf{Q}(i)$ goes to zero as $i$ goes to infinity. Observe that in each visit to $(s,a) \in \mathbb{S} \times \mathbb{A}$, the ``count'' $n_{sa}$ increases by 1 since either $\alpha_{sa}$ is increased by 1 or $\beta_{sa}$ is increased by 1. Furthermore, due to the $\epsilon$-greedy construction of MFABL and the ``connectedness'' assumption (recall the discussion when defining initial state probabilities in \S\ref{sec:model}), each state-action pair is visited infinitely often and hence $n_{sa}(i) \to \infty$ as $i$ goes to infinity for all $(s,a) \in \mathbb{S} \times \mathbb{A}$. This implies that the variance of $Q_{sa}(i)$ goes to 0 because
\begin{align*}
\text{Var}\left( Q_{sa}(i) \right) &= \text{Var}\left( \text{Beta}(\alpha_{sa}(i), \beta_{sa}(i) \right)  \\
    &= \frac{\alpha_{sa}(i) \times \beta_{sa}(i)}{n_{sa}^2(i) \times (n_{sa}(i) + 1)} \\
    &\le \frac{n_{sa}(i) \times n_{sa}(i)}{n_{sa}^2(i) \times (n_{sa}(i) + 1)} \\
    &= \frac{1}{n_{sa}(i) + 1}.
\end{align*}
The proof is complete.
\hfill  $\Halmos$

\section{Proof of Theorem~\ref{theorem:ConRate}} \label{sec:ConRateProof}

Recall $\mathbf{Q}^N$ denotes the belief maintained by
MFABL after $N$ customers have left (converted or quit). 
With $\mathbf{\overline{Q}}^N = \mathbb{E}[\mathbf{Q}^N]$, observe that for all $\pmb{u} > \pmb{0}$,
\begin{align*}
\lVert \mathbf{Q}^N - \mathbf{Q}^* \rVert_{\pmb{u}} = \lVert \mathbf{Q}^N - \mathbf{\overline{Q}}^N + \mathbf{\overline{Q}}^N  - \mathbf{Q}^* \rVert_{\pmb{u}}  \le \underbrace{\lVert \mathbf{Q}^N - \mathbf{\overline{Q}}^N \rVert_{\pmb{u}}}_{(\star)} + \underbrace{\lVert \mathbf{\overline{Q}}^N  - \mathbf{Q}^* \rVert_{\pmb{u}}}_{(\square)}.
\end{align*}
We decompose our analysis into $(\square)$ and $(\star)$.

\subsection*{Convergence Rate of $(\square)$ via Q-Learning}
Lemma  \ref{lemma:CFOExpectedUpdate} implies $\mathbf{\overline{Q}}$ mimics asynchronous Q-learning with a linear learning-rate (see \S3 of \cite{even2003learning}).
As such, we can leverage the corresponding Q-learning convergence rates to analyze $(\square)$ and our Theorem \ref{theorem:ConRate} is a meta-theorem that converts a given convergence rate for Q-learning into one for MFABL.
We prove Theorem \ref{theorem:ConRate} assuming access to the following generic convergence rate for Q-learning: 
there exists a $\pmb{v} > \pmb{0}$ such that for all $\varrho > 0$,
\begin{align}
\mathbb{P}\left\{\lVert \mathbf{\overline{Q}}^N  - \mathbf{Q}^* \rVert_{\pmb{v}} <
  \varrho\right\} >  1 - \delta(N, \varrho), 
\label{eq:QLearningGenericRate}
\end{align}
where $\lim_{N \to \infty}\delta(N, \varrho) = 0$, for all $\varrho > 0$.
See Theorem 5 in \cite{even2003learning} for an off-the-shelf example of such a rate.
Note that the convergence results in \cite{even2003learning} 
  are
  for the
  maximum norm $\lVert \cdot \rVert_{\infty}$ as opposed to the
  \emph{weighted} maximum norm $\lVert \cdot \rVert_{\pmb{v}}$. 
  We
  use the weighted norm 
  because the cumulative reward in our
  Markov decision process in not discounted. 
  The results in~\cite{even2003learning}  rely on discounting 
  to guarantee an appropriate 
  contraction; consequently, they 
  do not directly apply
  to our setup. 
  That being said, 
  Assumption \ref{ass:CFOAbsorption}
  (absorption) 
  implies that the rate results
  in \cite{even2003learning} 
  continue to hold when the maximum norm is replaced by the weighted
  maximum norm. The details are as follows:
\begin{enumerate}
\item 
  Assumption \ref{ass:CFOAbsorption}
  (absorption) implies
  that $\mathbf{\overline{Q}}^N$ 
  satisfies the contraction property in Definition~7(3) in \S6 of
  \cite{even2003learning} for the 
  weighted maximum norm with respect
  to some $\pmb{v} > 0$
  (see the
  discussion above Theorem 4 in \citet{tsitsiklis1994asynchronous}).  
\item 
  Thus, $\mathbf{\overline{Q}}^N$ is a 
  ``well-behaved
  iterative stochastic algorithm'' as per \cite{even2003learning} under
  the 
  the weighted maximum norm.
\item 
  The key result for establishing the bounds 
  in \cite{even2003learning} is
  Lemma 9, which continues to hold when the stochastic process satisfies
  contraction with respect 
  to the weighted maximum norm. 
\end{enumerate}
In summary, even though \cite{even2003learning} prove convergence rates for
$\lVert \mathbf{\overline{Q}}^N  - \mathbf{Q}^* \rVert_{\infty}$ for
discounted MDPs,  these rates apply to undiscounted absorbing MDPs as well
under \emph{some} weighted norm, i.e.,  \emph{there exists} a $\pmb{v} >
\pmb{0}$ for which $\lVert \mathbf{\overline{Q}}^N  - \mathbf{Q}^*
\rVert_{\pmb{v}}$ obeys the same convergence rate. 

\subsection*{Bounding the Additional Term $(\star)$}
Consider $| Q^N_{sa} - \overline{Q}^N_{sa} |$ for an arbitrary $(s,a) \in
\mathbb{S} \times \mathbb{A}$ and the positive vector $\pmb{v} = [v_{sa}]_{(s,a)}$ from above.  Define $\vmin := \min_{(s,a) \in \mathbb{S} \times \mathbb{A}} v_{sa}$, which is strictly positive as $\pmb{v} > 0$. Since $\overline{Q}^N_{sa} =
\mathbb{E}[Q^N_{sa}]$, we can invoke a standard concentration bound.   In
particular,  Chebyshev's inequality implies the following: 
\begin{align}
\label{eq:ChebInequality}
\mathbb{P}\left\{| Q^N_{sa} - \overline{Q}^N_{sa} | < \varrho \vmin \bigl\vert n_{sa}^N
  \right\} > 1 - \frac{\text{Var}(Q^N_{sa})}{\varrho^2 \vmin^2} \ge 1 -
  \frac{1}{\varrho^2 \vmin^2 n_{sa}^N}. 
\end{align}
where $n_{sa}^N$ denotes the number of visits to $(s,a)$ after $N$ consumers.
Recall that $\text{Var}(Q^N_{sa}) \le \frac{1}{n_{sa}^N}$ 
was established in the proof of Theorem \ref{theorem:CFOConvergence}
(\S\ref{sec:CFOConvergenceProofSub} in particular). Next, we 
quantify
$n_{sa}^N$ as a function of $N$.

Recall that we have $\epsilon$-greedy in MFABL and assume
that our state space is ``connected''. 
Therefore, there exists 
positive probability  
$\eta_{sa}>0$ that $(s,a)$ is visited at least
once for every consumer.  
Let $B(N,\eta_{sa})$ denote a Bernoulli random variable with $N$ trials
and success probability $\eta_{sa}$. Then, we claim that $n^N_{sa}$
stochastically dominates $B(N,\eta_{sa})$, and this is because
$B(N,\eta_{sa})$ only 
counts
the \emph{first} visit of each of the first
$N$ consumers whereas $n_{sa}^N$ 
counts
\emph{all} the visits. 
Since $\mathbb{E}[B(N,\eta_{sa})] = \eta_{sa}N$,
Hoeffding's inequality implies that 
\begin{align*}
  \mathbb{P}\left\{n^{N}_{sa} > k \right\} \ge \mathbb{P}\left\{ B(N,\eta_{sa}) > k \right\}
  \ge 1  -e^{-2N \left(\eta_{sa} -
  \frac{k}{N} \right)^2},
\end{align*}
for $k \le \eta_{sa} N$.
On substituting $k = \eta_{sa} N/2$ 
we get:
\begin{align*}
  \mathbb{P} \left\{n^{N}_{sa} > \eta_{sa}N/2 \right\} \ge \mathbb{P}\left\{ B(N,\eta_{sa}) > \eta_{sa}N/2 \right\}
  \ge 1  -e^{-N\eta_{sa}^2/2}.  
\end{align*}
Next, we have that 
\begin{align}
  \mathbb{P}\left\{| Q^N_{sa} - \overline{Q}^N_{sa} | < \varrho \vmin \right\}
  & \ge \mathbb{P}\left\{ |Q^N_{sa} - \overline{Q}^N_{sa} | <
    \varrho \vmin \bigl\vert n^N_{sa} > \eta_{sa}N/2 \right\} \mathbb{P}\left\{ n^N_{sa} >\eta_{sa}N/2
    \right\} \nonumber \\
  & > \left(1 - \frac{2}{\varrho^2 \vmin^2 \eta_{sa} N} \right) \left( 1 -
    e^{-N\eta_{sa}^2/2}  
    \right) \nonumber \\
    &\ge  1 - \frac{2}{\varrho^2 \vmin^2 \eta_{sa} N} - e^{-N \eta_{sa}^2/2} \nonumber \\ 
    &\ge  1 - \frac{2}{\varrho^2 \vmin^2 \eta N} - e^{-N \eta^2/2} \label{eq:ChebInequalityTwo},       
\end{align}
where $\eta := \min_{(s,a) \in \mathbb{S} \times \mathbb{A}} \eta_{sa} > 0$.
\eqref{eq:ChebInequalityTwo} implies the following bound for $\lVert \mathbf{Q}^N - \mathbf{\overline{Q}}^N \rVert_{\infty}$:
\begin{align}
\label{eq:ChebInequalityNorm}
\mathbb{P}\left\{\lVert \mathbf{Q}^N - \mathbf{\overline{Q}}^N \rVert_{\infty} < \varrho \vmin \right\} > 1 - \frac{2S(A+1)}{\varrho^2 \vmin^2 \eta N} - S(A+1)e^{-N\eta^2/2}. 
\end{align}
Recall that $S$ and $A+1$ denote the size of $\mathbb{S}$ and $\mathbb{A}$, respectively.
\eqref{eq:ChebInequalityNorm} holds because for $E_{sa} :=  \left\{ | Q^N_{sa}
  - \overline{Q}^N_{sa} | < \varrho \vmin \right\}$, we have 
\begin{align*}
  \mathbb{P}\left\{ E_{sa} \cap E_{s'a'} \right\}
  &= \underbrace{\mathbb{P}\{ E_{sa}\}}_{> 1 -  \frac{2}{\varrho^2 \vmin^2 \eta N}
    - e^{-N\eta^2/2}}
    + \underbrace{\mathbb{P}\{ E_{s'a'}
    \}}_{> 1 -  \frac{2}{\varrho^2 \vmin^2 \eta N}
    - e^{-N\eta^2/2}}
    - \underbrace{\mathbb{P}\{ E_{sa}
    \cup E_{s'a'} \}}_{\le 1}.  
\end{align*}
It is straightforward to generalize this to $\mathbb{P}\{ \cap_{(s,a) \in \mathbb{S} \times \mathbb{A}} E_{sa} \}$, which gives
\eqref{eq:ChebInequalityNorm}. 
The maximum norm bound in \eqref{eq:ChebInequalityNorm} implies the following bound for the weighted norm:
\begin{align}
\label{eq:ChebInequalityWeightedNorm}
\mathbb{P}\left\{\lVert \mathbf{Q}^N - \mathbf{\overline{Q}}^N \rVert_{\pmb{v}} < \varrho \right\} > 1 - \frac{2S(A+1)}{\varrho^2 \vmin^2 \eta N} - S(A+1)e^{-N\eta^2/2}. 
\end{align}
This is because $\lVert \mathbf{Q}^N - \mathbf{\overline{Q}}^N \rVert_{\infty} < \varrho \vmin$ implies $\lVert \mathbf{Q}^N - \mathbf{\overline{Q}}^N \rVert_{\pmb{v}} < \varrho$.  

\subsection*{Putting $(\square)$ and $(\star)$ Together: The Meta-Result}
Finally,  merging \eqref{eq:QLearningGenericRate} and \eqref{eq:ChebInequalityWeightedNorm} and using the same logic as we did for $\mathbb{P}\{ E_{sa} \cap E_{s'a'} \}$ above, we get
\begin{align*}
\mathbb{P}\left\{ \lVert \mathbf{Q}^N - \mathbf{Q}^* \rVert_{\pmb{v}} < 2 \varrho
  \right\} > 1 - \delta(N, \varrho) - \frac{2S(A+1)}{\varrho^2 \vmin^2 \eta N} -
  S(A+1)e^{-N \eta^2/2}. 
\end{align*}
The first term ($\delta(N, \varrho)$) is provided by off-the-shelf rate for Q-learning, the second term decays as $\mathcal{O}(\frac{1}{\varrho^2 N})$, and the third term decays as $\mathcal{O}(e^{-N})$. The proof is now complete. \hfill \Halmos

\leaveline

\begin{remark}[Polynomial Learning Rate] 
\label{rem:polystep}
We can modify MFABL so that the corresponding $\mathbf{\overline{Q}}$ process mimics asynchronous Q-learning with a ``polynomial'' learning rate (instead of ``linear''). In that case, we can use the off-the-shelf convergence rate for Q-learning from Theorem 4 (instead of Theorem 5) of \cite{even2003learning}.
To do so, we simply need to change the updates in lines 10 and 12 of Algorithm \ref{alg:CFOOurAlgo} to as follows:
\begin{align*}
\alpha_{sa^*}  &\leftarrow \alpha_{sa^*} + \frac{\alpha_{sa^*} + \beta_{sa^*}}{(n_{sa^*} + 1)^{\omega} - 1} \\
\beta_{sa^*} &\leftarrow \beta_{sa^*} + \frac{\alpha_{sa^*} + \beta_{sa^*}}{(n_{sa^*} + 1)^{\omega} - 1},
\end{align*}
where $\omega \in (1/2, 1]$, with $\omega \in (1/2,1)$ resulting in  a polynomial learning-rate and $\omega=1$ resulting in a linear learning rate.  Note that the notation $n_{sa}$ here tracks the number of visits to $(s,a)$ and does not necessarily equal $\alpha_{sa} + \beta_{sa}$ anymore. Under these updates, it is straightforward to verify that the expected update obeys the following equation (analogous to Lemma \ref{lemma:CFOExpectedUpdate}): 
\begin{align*}
\overline{Q}_{sa}(i+1) = \left( 1 - \left(\frac{1}{n_{sa}(i) + 1}\right)^{\omega} \right) \overline{Q}_{sa}(i) +  \left(\frac{1}{n_{sa}(i) + 1}\right)^{\omega} f_{s'}.
\end{align*}
This mimics Q-learning with a polynomial learning rate (see \S3 of \citet{even2003learning}).  Furthermore, it is easy to verify that plugging in $\omega = 1$ recovers the updates in Algorithm \ref{alg:CFOOurAlgo}.
\end{remark}

\leaveline

\begin{remark}[MFABL with Discounting] 
\label{rem:MFABLdicount}
It is possible to modify MFABL to allow for discouting.
To do so, we need to change the updates in lines 10 and 12 of Algorithm \ref{alg:CFOOurAlgo} to as follows:
\begin{align*}
\alpha_{sa^*} &\leftarrow \alpha_{sa^*} + \frac{(\alpha_{sa^*} + \beta_{sa^*})\left\{\gamma(\alpha_{sa^*} + \beta_{sa^*}) - \alpha_{sa^*}\right\}}{n_{sa^*} \beta_{sa^*} + (1-\gamma) (\alpha_{sa^*} + \beta_{sa^*})} \\
\beta_{sa^*}  &\leftarrow \beta_{sa^*} + \frac{\alpha_{sa^*} + \beta_{sa^*}}{n_{sa^*}},
\end{align*}
where $\gamma \in [0,1)$ denotes the discounts factor.  As in Remark \ref{rem:polystep},  $n_{sa}$ tracks the number of visits to $(s,a)$. Under these updates, it is straightforward to verify that the expected update obeys the following equation (analogous to Lemma \ref{lemma:CFOExpectedUpdate}): 
\begin{align}
\overline{Q}_{sa}(i+1) = \frac{n_{sa}(i)}{n_{sa}(i) + 1} \overline{Q}_{sa}(i) + \gamma \frac{1}{n_{sa}(i) + 1} f_{s'}.
\end{align}
This mimics discounted Q-learning with a linear learning-rate (see \S3 of \citet{even2003learning}).  Furthermore, it is easy to verify that plugging in $\gamma = 1$ recovers the updates in Algorithm \ref{alg:CFOOurAlgo}. For this variant of MFABL, we can directly invoke the maximum norm results of \cite{even2003learning}.
\end{remark}

\section{Additional Details for Numerics} \label{sec:NumericsAppendix}
Additional details corresponding to our numerics of \S\ref{sec:numerics} are presented here. In \S\ref{sec:BenchmarksAppendix}, we discuss the benchmark algorithms in detail and in \S\ref{sec:SensAppendix}, we present sensitivity analysis. 

\subsection{Benchmark Algorithms} \label{sec:BenchmarksAppendix}

\bold{Thompson Sampling (TS).} 
TS \citep{thompson1933likelihood} is a well-known learning algorithm for multi-armed bandits, as explained in \S\ref{sec:CFOAlgoMotivation}.  
We extend it to the conversion funnel MDP as follows. TS maintains a Beta belief over the one-step conversion probability of each state-action pair, i.e., $\text{Beta}(\alpha_{sa}, \beta_{sa})$. When a consumer is in state $s \in \mathbb{S}$, TS generates a sample from $\text{Beta}(\alpha_{sa}, \beta_{sa})$ for all $a \in \mathbb{A}$ and plays the action with the highest sample value. If the consumer transitions to the conversion state (in one step), TS increases $\alpha_{sa}$ by 1 of the corresponding action. Else, it increases $\beta_{sa}$ by 1 of the corresponding action. At initialization, $\alpha_{sa}$ and $\beta_{sa}$ are set to 1 and 9, respectively for all $(s,a) \in \mathbb{S} \times \mathbb{A}$ and we do extensive sensitivity analysis with respect to this choice (\S\ref{sec:SensAppendix}).

\bold{Posterior Sampling for Reinforcement Learning (PSRL).}
PSRL \citep{strens2000bayesian}
generalizes TS in a model-based manner. It maintains a
belief over the entire transition probabilities $\mathcal{P}$. In
particular, given $(s,a) \in \mathbb{S} \times \mathbb{A}$, it maintains a
Dirichlet($\pmb{\alpha}_{sa}$) belief over the one-step transition
probability vector $\pmb{p}_{sa} := [p_{sas'}]_{s' \in \mathbb{S}^+}$
where $\pmb{\alpha}_{sa} := [\alpha_{sas'}]_{s' \in \mathbb{S}^+}$. Before
each consumer arrives, PSRL generates a sample from
Dirichlet($\pmb{\alpha}_{sa}$) for all $(s,a) \in \mathbb{S} \times
\mathbb{A}$. Denoting the corresponding sampled transition probabilities
by $\mathcal{\widehat{P}}$, it computes an optimal policy of the MDP
$\mathcal{\widehat{M}} \equiv (\mathbb{S}, \mathbb{A},
\mathcal{\widehat{P}}, \pmb{\lambda}, r)$ and plays the computed
policy. Using the transition data of the form ``$(s,a,s')$'' observed in
the realized consumer path, it updates the belief over $\mathcal{P}$ using
Dirichlet-multinomial conjugacy. For example, if taking action $a \in
\mathbb{A}$ at state $s \in \mathbb{S}$ transitioned the consumer to state
$s' \in \mathbb{S}^+$, PSRL increases $\alpha_{sas'}$ by 1. This updated
belief is used to generate a sample of $\mathcal{P}$ and re-compute an
optimal policy for the next consumer and so on. Hence, given $N$
consumers, PSRL solves $N$ MDPs. 
For a large-scale MDP (which is the case in our \S\ref{sec:numerics} application),
doing so can be prohibitively expensive.  To manage compute time, we leveraged the sparsity of the underlying transition structure (recall the consumer behavior falls in
one of the four buckets of ignore, open, click, and convert).  The sparsity was encoded by initializing the Dirichlet counts as follows: $\forall (s,a,s') \in \mathbb{S} \times \mathbb{A} \times \mathbb{S}^+$, 
\begin{align*}
\alpha_{sas'} = \begin{cases}
1 &\text{ if $(s,a,s')$ is a feasible transition} \\
0 &\text{ otherwise.}
\end{cases}
\end{align*}
Hence, the sampled $\mathcal{\widehat{P}}$ was always sparse and we used sparse data structures in \texttt{MATLAB} to manipulate $\mathcal{\widehat{P}}$ (e.g.,  taking matrix inverse to compute the optimal MDP policy via policy iteration).
Despite this (and using a high-performance computing cluster), we found the PSRL compute time to be extremely high. To further reduce the compute time,  instead of re-optimizing after every consumer, we re-optimized after every 1000 consumers. As we use $N = 500,000$ consumers in our numerics,  this meant we re-optimized for a total of $500,000 / 1000 = 500$ times in PSRL.  Even with this trick, PSRL's runtime was of the order of 100x as compared to the other approaches (TS and MFABL), as discussed in \S\ref{sec:results}.

\bold{Q-Learning with Upper Confidence Bounds (QL-UCB).} 
QL-UCB \citep{jin2018q, dong2019q} is a variant of QL that uses upper confidence bounds (instead of Thompson sampling),  and 
  comes with 
  strong 
  theoretical guarantees.
  We refer the
reader to \cite{jin2018q} and \cite{dong2019q} for its formal
description. We use the version presented in \cite{dong2019q} since it
allows for a variable horizon (i.e., length of a consumer path) whereas
the version in \cite{jin2018q} assumes a deterministic horizon. We note
that both \cite{jin2018q} and \cite{dong2019q} assume the presence of a
discount factor $\gamma \in [0,1)$. In our MDP, we do not discount the
terminal reward and hence, MFABL implicitly sets $\gamma = 1$. For
comparison purposes, in the algorithm of \cite{dong2019q}, we set $\gamma$ to be close to 1 (as setting it to 1 results in some of their parameters being undefined). There are 2 other parameters: $\epsilon$ and $\delta$. The physical meaning of $\delta$ is that the theoretical guarantee of \cite{dong2019q} holds w.p.\ $1-\delta$ (hence, $\delta$ should be close to 0) and $\epsilon$ quantifies the degree of suboptimality (hence, $\epsilon$ should be close to 0 as well). 
We tried 8 possible combinations in the set $\{(\epsilon, \gamma, \delta) : \epsilon \in \{0.01, 0.05\}, \gamma \in \{0.95, 0.99\}, \delta \in \{0.01, 0.05\}\}$ and obtained very similar results for each combination.  The results we report in the paper are for $(0.01, 0.99, 0.01)$. 
Finally,  we initialize $Q_{sa}$ for all $(s,a) \in \mathbb{S} \times \mathbb{A}$ as suggested in \cite{dong2019q}, $Q_{qa} = 0$, and $Q_{ca} = 1$ for all $a \in \mathbb{A}$. 

\subsection{Sensitivity Analysis} \label{sec:SensAppendix}

There are two sets of input parameters that we perform sensitivity analysis on.  The first set corresponds to the prior counts used in TS, MFABL, and pMFABL and the second corresponds to the $\epsilon$ value used in MFABL and pMFABL. We report the corresponding sensitivity analysis in Tables \ref{table:alphabeta} and \ref{table:eps}.  For both sets of analysis, we use $N = 500,000$ and average over $R=100$ seeds, as we did in the \S\ref{sec:numerics} results. 

Our findings in \S\ref{sec:numerics} are robust to such perturbations.  In terms of Table \ref{table:alphabeta}, TS achieves an average PR of 0.27 to 0.28 as we vary the prior counts and pMFABL's average PR consistently lies between 0.80 and 0.82.  Though MFABL's average PR seems to be sensitive around lower values of $\beta_0$, it still outperforms the three benchmarks (by a factor of almost 2 or more).  
It is interesting to observe that MFABL is more sensitive to the
  prior as compared to pMFABL.  
  Our explanation for this is similar to the
  one discussed in \S\ref{sec:results} 
  to explain
  why pMFABL achieves a higher PR than MFABL. 
  In particular,  MFABL synthetically generates the feedback $f_{s'}$ by
  sampling from the belief over the action with the highest expected value
  at state $s'$.  
  If this belief is incorrect (which is likely in the initial stages of
  MFABL), the generated feedback might not be very informative. 
On the other hand, pMFABL never generates synthetic feedback in its attribution module but simply rolls back the observed data (conversion or not). As such,  MFABL inherently depends on the prior belief in its attribution module (since it generates feedback using it) whereas pMFABL does not. 

\begin{table}
\small
\centering
\begin{tabular}{c | c | c | c | c}
\multicolumn{2}{c | }{\bold{Prior counts}} &  \multicolumn{3}{c}{\bold{Performance ratio}} \\
$\alpha_0$ & $\beta_0$  & \bold{TS} & \bold{MFABL} & \bold{pMFABL}  \\
\hline 
1 & 4 & 0.27 & 0.49 & 0.82  \\   
1 & 5 & 0.28 & 0.55 & 0.80  \\   
1 & 6 & 0.27 & 0.59 & 0.81  \\   
1 & 7 & 0.27 & 0.62 & 0.81  \\   
1 & 8 & 0.27 & 0.64 & 0.82  \\   
1 & 9 & 0.27 & 0.64 & 0.81  \\   
1 & 10 & 0.28 & 0.63 & 0.82  \\   
1 & 11 & 0.28 & 0.66 & 0.81  \\   
1 & 12 & 0.28 & 0.65 & 0.81  \\   
1 & 13 & 0.28 & 0.64 & 0.82  \\   
1 & 14 & 0.28 & 0.66 & 0.81  \\   
1 & 15 & 0.28 & 0.64 & 0.81  \\   
1 & 16 & 0.28 & 0.66 & 0.81  \\   
1 & 17 & 0.28 & 0.63 & 0.82  \\   
1 & 18 & 0.28 & 0.63 & 0.81  \\   
1 & 19 & 0.28 & 0.64 & 0.81  \\   
\end{tabular}
\caption{\normalfont Sensitivity of performance ratio with respect to prior counts $(\alpha_0, \beta_0)$. 
In TS, MFABL,  and pMFABL, we initialize the Beta belief of each state action pair via the prior counts $(\alpha_0, \beta_0)$,  i.e.,  $(\alpha_{sa}, \beta_{sa}) = (\alpha_0, \beta_0)$ for all $(s,a) \in \mathbb{S} \times \mathbb{A}$.
We report the performance ratio of TS, MFABL,  and pMFABL as we vary $\beta_0 \in \{4, \ldots,19\}$ while holding $\alpha_0 = 1$.  Note that the prior belief's expected value equals $\frac{\alpha_0}{\alpha_0 + \beta_0}$. Hence,  $(\alpha_0,\beta_0) = (1,4)$ corresponds to a high prior mean of 0.2 whereas $(\alpha_0,\beta_0) = (1,19)$ corresponds to a more realistic prior mean of 0.05, since such marketing campaigns have low conversion rates. 
In the results we reported in Figure \ref{fig:PR}, we used $(\alpha_0,\beta_0) = (1,9)$, i.e., a prior mean of 0.1.}
\label{table:alphabeta}
\end{table}

\begin{figure}
\centering
\begin{subfigure}[]{0.45\textwidth}
\centering
\includegraphics[scale=0.4]{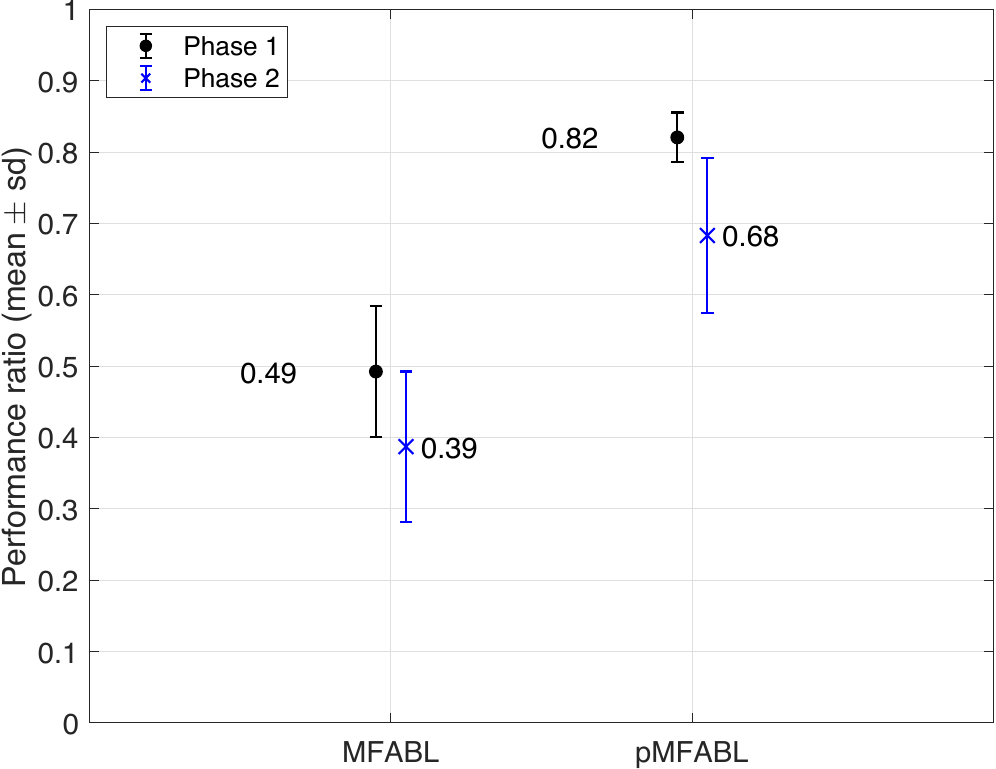}
\caption{\small $(\alpha_0,\beta_0) = (1, 4)$}
\label{fig:ConceptPrior1-4}
\end{subfigure}
\begin{subfigure}[]{0.45\textwidth}
\centering
\includegraphics[scale=0.4]{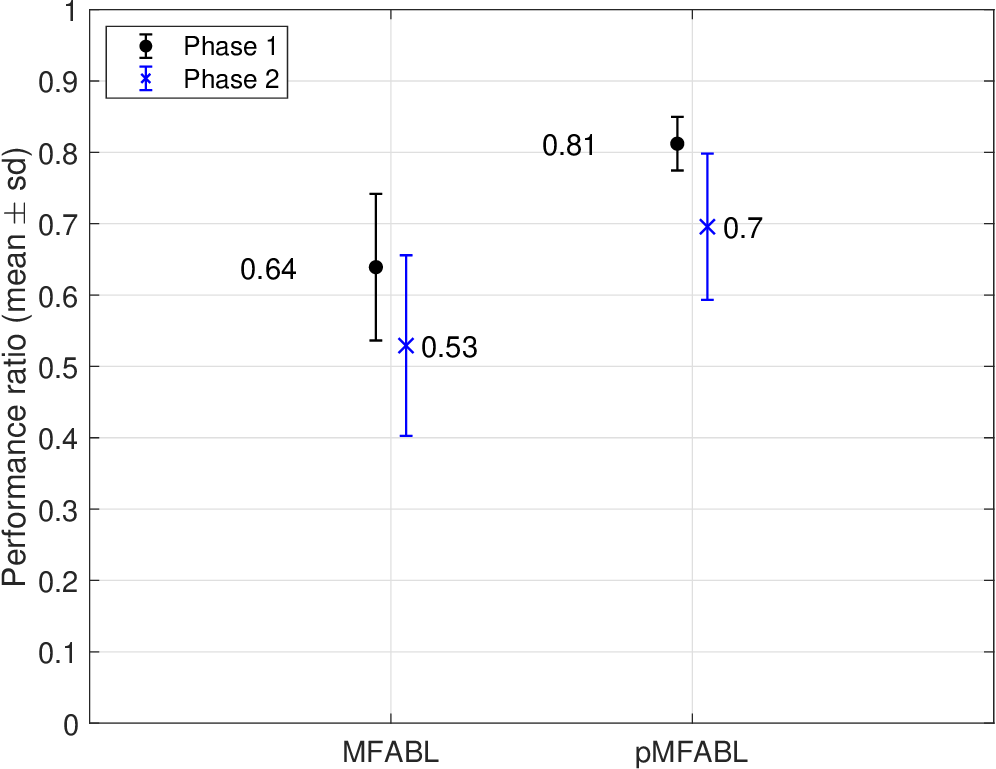}
\caption{\small $(\alpha_0,\beta_0) = (1, 19)$}
\label{fig:ConceptPrior1-19}
\end{subfigure}
\caption{\normalfont Performance of MFABL and pMFABL under concept shift for two different settings of $(\alpha_0,\beta_0)$. }
\label{fig:ConceptPrior}
\end{figure}

As a side note, it is also of interest to explore how the prior affects the performance of MFABL and pMFABL under concept shift (recall the setup in \S\ref{sec:conceptshift}). To do so, we perform the same computations as we did in \S\ref{sec:conceptshift} but by setting the prior counts $(\alpha_0,\beta_0) \in \{(1,4), (1,19)\}$ instead of $(1,9)$. The corresponding results are shown in Figure \ref{fig:ConceptPrior}. The phase 1 results are identical to those reported in Table \ref{table:alphabeta} with MFABL achieving a PR of 0.49, 0.64,  and 0.64 and pMFABL achieving 0.82, 0.81,  and 0.81 for $\beta_0 =$ 4, 9,  and 19, respectively.\footnote{The concept shift results for $\beta_0 = 9$ are shown in Figure \ref{fig:PRconcept} (\S\ref{sec:conceptshift}).}
In terms of phase 2,  MFABL exhibits a similar sensitivity to the prior choice as it does in phase 1 (especially for a low value of $\beta_0$) whereas pMFABL shows robustness. In particular, the phase 2 PR under MFABL is 0.39, 0.52, and 0.53 whereas under pMFABL is 0.68, 0.67, and 0.7 for $\beta_0 =$ 4, 9,  and 19, respectively. 

In terms of Table \ref{table:eps},  both MFABL and pMFABL exhibit a decline in PR as we increase $\epsilon$. 
Though a higher value of $\epsilon$ encourages exploration in the earlier stages of learning, it also corresponds to the probability of sending a random email even when the algorithm has found the optimal policy.  Hence, a higher value of $\epsilon$ results in more random (and possibly suboptimal) emails being sent (asymptotically), which explains the decrease of PR as we increase $\epsilon$.  Irrespective,  even for a high value of $\epsilon = 0.1$, both MFABL and pMFABL comfortably outperform the three benchmarks.
It is interesting to note that $\epsilon = 0$ achieves the best
  performance, suggesting the performance of MFABL/pMFABL can be improved
  without $\epsilon$-greedy. 
  A possible explanation is that 
  MFABL has 
  exploration built-in since it samples from the Beta distributions (similar to
  Thompson sampling for bandits),  and therefore,  
  additional exploration in the
  form of $\epsilon$-greedy might not be needed. 
However, we are unable to prove that this built-in exploration results in each state-action pair being visited infinitely often, which is a ``minimal requirement'' for convergence to $\mathbf{Q}^*$ (quoted from \S6.5 of \cite{sutton2018reinforcement}). 

\begin{table}
\small
\centering
\begin{tabular}{c | c | c }
 &  \multicolumn{2}{c}{\bold{Performance ratio}} \\
$\epsilon$ & \bold{MFABL} & \bold{pMFABL}  \\
\hline 
0.00 & 0.65 & 0.82 \\ 
0.01 & 0.64 & 0.81 \\   
0.02 & 0.63 & 0.80 \\   
0.03 & 0.61 & 0.80 \\   
0.04 & 0.61 & 0.79 \\   
0.05 & 0.60 & 0.78 \\   
0.06 & 0.61 & 0.77 \\   
0.07 & 0.59 & 0.75 \\   
0.08 & 0.57 & 0.75 \\   
0.09 & 0.57 & 0.74 \\   
0.10 & 0.56 & 0.73 \\   
\end{tabular}
\caption{\normalfont Sensitivity of performance ratio with respect to  $\epsilon$. 
In MFABL and pMFABL, we sample a random action w.p.\ $\epsilon$.
We report the performance ratio of MFABL and pMFABL as we vary $\epsilon \in \{0.00, 0.01,  0.02, \ldots, 0.10\}$. 
In the results we reported in Figure \ref{fig:PR}, we used $\epsilon = 0.01$.}
\label{table:eps}
\end{table}

\subsection{Results for Small Number of Consumers} \label{sec:SmallN}
\begin{figure}[H]
\centering
\includegraphics[scale=0.4]{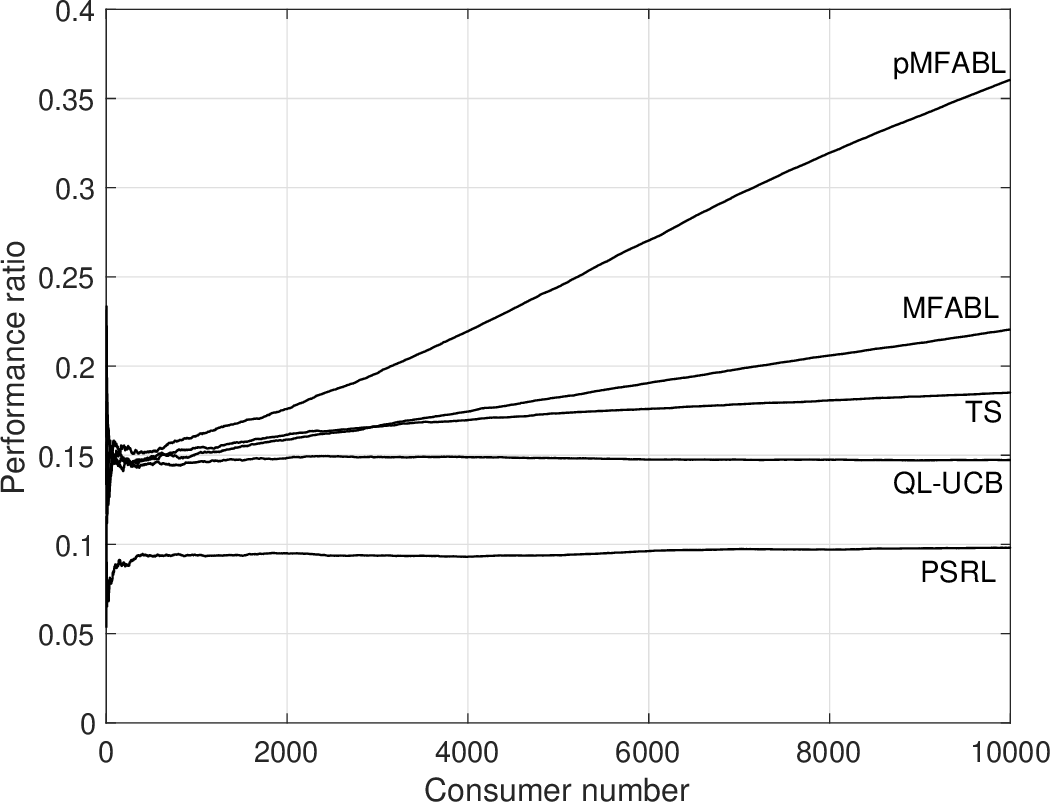}
\caption{\normalfont Evolution of PR as we increase $N$ from 1 to 10,000. This figure zooms in on Figure \ref{fig:PRdynamic}.}
\label{fig:PRdynamic10000}
\end{figure}

Figure \ref{fig:PRdynamic} in \S\ref{sec:results} showed the evolution of PR as we increase $N$ from 1 to 500,000. Given this scale, it made it challenging to understand the evolution when $N$ is small. As such, in Figure \ref{fig:PRdynamic10000}, we do so for $N$ up to 10,000. The results are directionally similar to the ones in Figure \ref{fig:PRdynamic} where MFABL and pMFABL outperform TS, PSRL, and QL-UCB. In particular, for $N=10,000$, MFABL improves PR by a factor of $1.19$, $2.25$, and $1.50$ when compared to TS, PSRL, and QL-UCB, respectively.  pMFABL boosts it by an additional factor of $1.63$ (over MFABL).

We also note that when the number of consumers is less than
10,000, the performance ratios are less than 0.4 (for all algorithms).
This can potentially make the algorithms difficult to operationalize in real-world settings where the number of potential consumers
may not be very large. 
The key challenge here is one is required to learn over a high-dimensional
space, which requires interacting with a large number of consumers.  
That being said, 
there  are 
many
firms that operate on 
the scale of
millions of potential consumers (one example being the firm we got data from).
In addition, one possibility to speed up the convergence is to embed prior knowledge (if available) in the form of prior counts (alpha and beta) that serve as inputs to MFABL / pMFABL. We briefly discuss this in \S4.3 under paragraph ``Prior information''.\footnote{Given the simplicity of MFBAL (as is the case with Thompson sampling for bandits),  embedding such prior information has a remarkably straightforward interpretation. In particular, if a state-action pair $(s,a)$ is initialized at the prior counts $(\alpha_{sa}, \beta_{sa})$, then it is equivalent to saying that a priori, the expected conversion rate from $(s,a)$ equals $\frac{\alpha_{sa}}{\alpha_{sa} + \beta_{sa}}$ and the variance equals the variance of Beta($\alpha_{sa}, \beta_{sa}$) distribution.}
Naturally, if the prior knowledge is consistent with reality,  embedding it will result in faster convergence.

\subsection{Supplementary Figures} \label{sec:AdditionalFigs}
\begin{figure}
\centering
\begin{subfigure}[]{0.3\textwidth}
\centering
\includegraphics[width=1\linewidth]{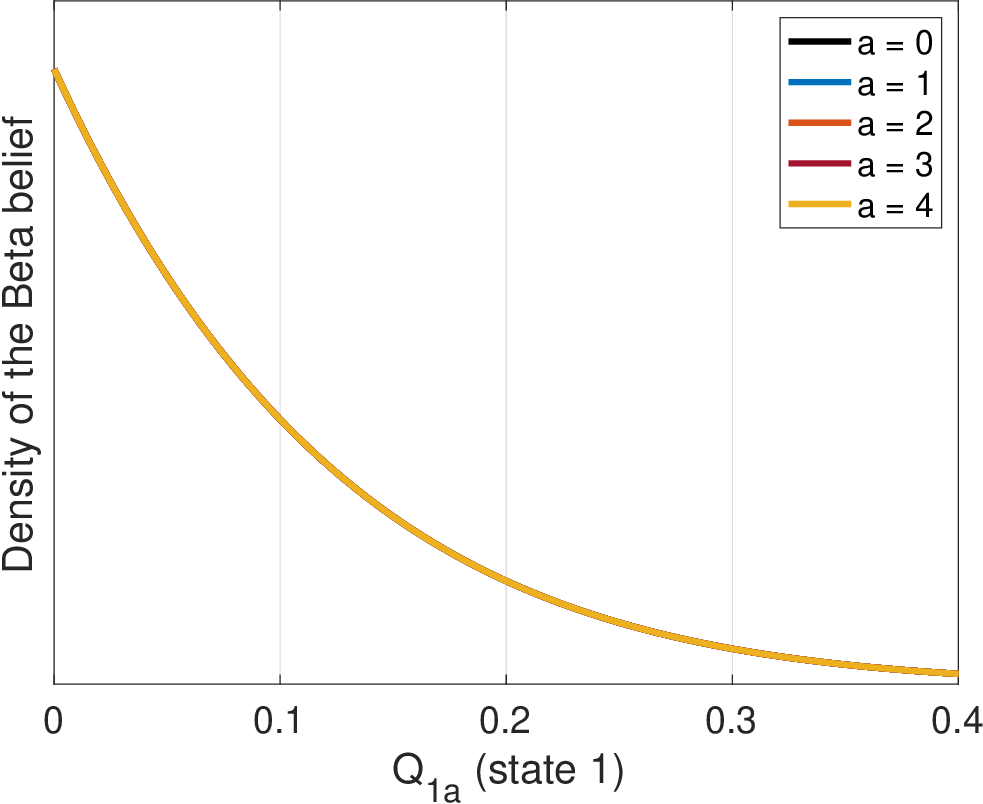}
\caption{\small $n=1$}
\label{fig:interpret1}
\end{subfigure}
\begin{subfigure}[]{0.3\textwidth}
\centering
\includegraphics[width=1\linewidth]{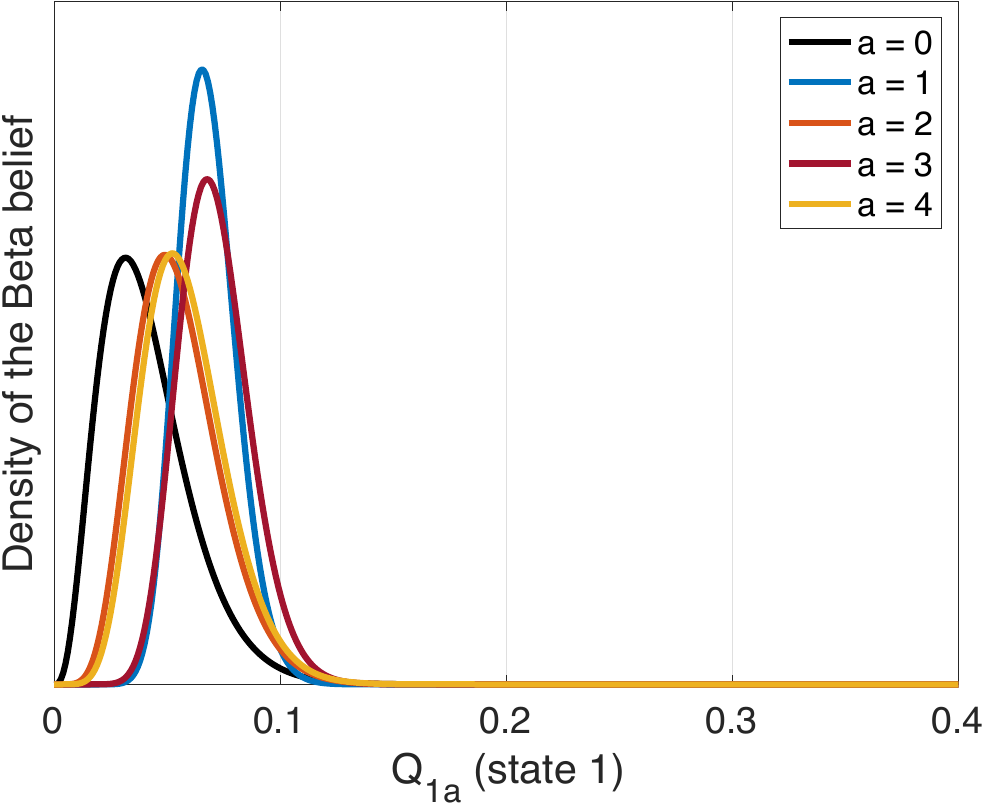}
\caption{\small $n=1000$}
\label{fig:interpret1000}
\end{subfigure}
\begin{subfigure}[]{0.3\textwidth}
\centering
\includegraphics[width=1\linewidth]{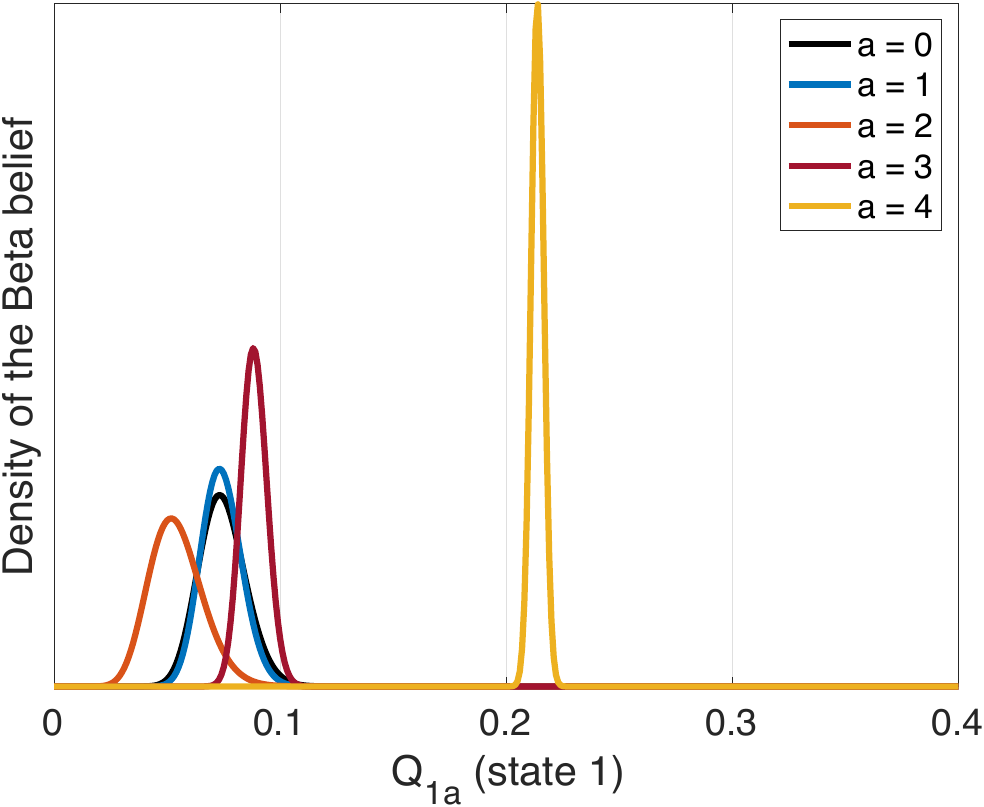}
\caption{\small $n=25000$}
\label{fig:interpret25000}
\end{subfigure}
\caption{\normalfont Evolution of the belief maintained by pMFABL
    over all possible actions (including no email $a=0$) in state 1 (for a 
    given
    seed). To understand the evolution, we picked consumers $n \in
    \{1, 1000, 25000\}$.  For $n=1$, all the actions have the same belief
    (i.e., the prior). Though the belief is somewhat informative after
    interacting with 1000 consumers, there is ambiguity in terms of which
    email is optimal at state 1.  Email \#4 stands out as the winner
    (w.h.p.) after $n=25000$ consumers, with the eventual conversion
    probability of slightly above 0.2. (Though not shown for brevity, we
    can visualize MFABL in a similar manner.)}
\label{fig:interpret}
\end{figure}

\begin{figure}
\centering
\includegraphics[scale=0.4]{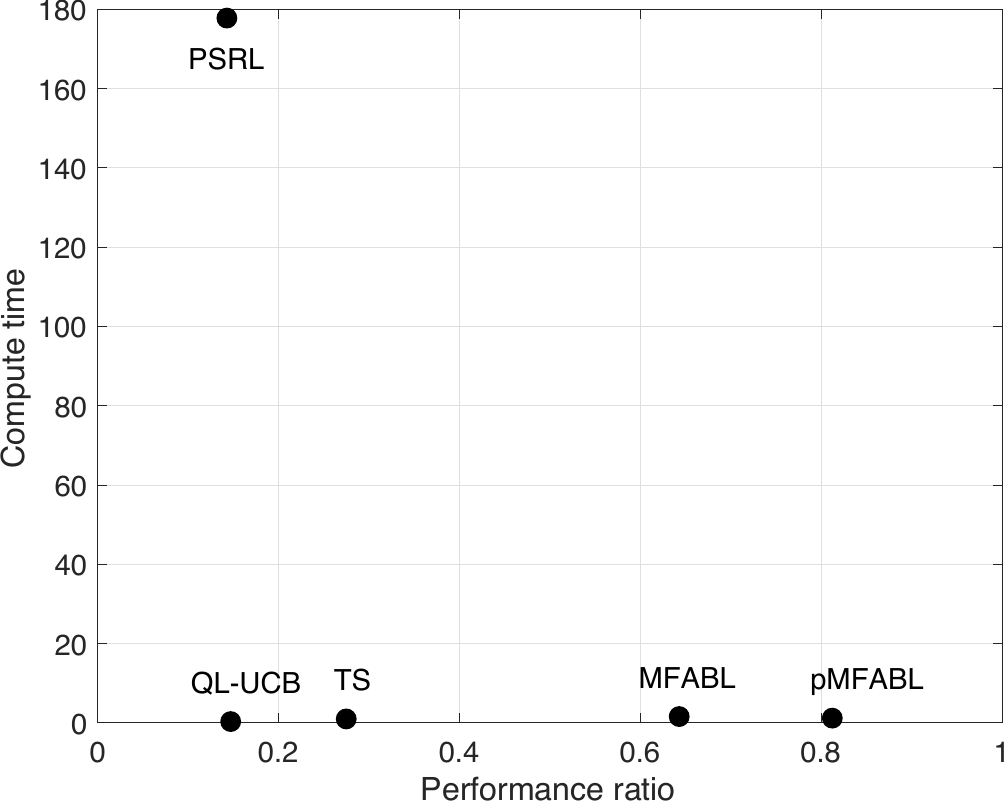}
\caption{\normalfont Compute time versus performance ratio~(PR) for $N =
  500,000$.  The plotted PR values are the average values
  in Figure \ref{fig:PRsnapshot} (averaged over $R=100$ seeds).  The compute times in seconds (averaged over $R=100$ seeds) were $5.49 \times 10^4$ (PSRL),  $308.56$ (TS),  $105.70$ (QL-UCB), $495.66$ (MFABL), and
  $372.81$ (pMFABL). 
  These times are plotted as a ratio
  with respect to the average compute time for TS, i.e., the 
  $y$ values are 177.8 (PSRL),  1 (TS), 0.34 (QL-UCB), 1.6 (MFABL), and 1.2 (pMFABL).  Note that compute time for a learning algorithm
  is the time spent in executing 
  lines 4 and 6 of Algorithm~\ref{alg:sim} but not time spent in
  simulating consumer behavior (line 5).}
\label{fig:TimeVsPR}
\end{figure}

See Figures \ref{fig:interpret} (interpretability) and \ref{fig:TimeVsPR} (compute times).

\subsection{Gradual Concept Shift} \label{sec:GradualConcept}
As mentioned in \S\ref{sec:conceptshift}, our two-phase treatment of concept shift can possibly be enhanced. We now do so via additional simulations pertaining to a ``gradual'' concept shift, which we model as follows. As in  \S\ref{sec:conceptshift}, we simulate $N=1,000,000$ sequential consumers. However, now, we introduce two parameters $N_1$ and $N_2$ to denote the start and end of the concept shift.  In particular, consumers 1 to $N_1$ are in phase 1 whereas consumers $N_2+1$ to $N$ are in phase 2, where phases 1 and 2 are as in \S\ref{sec:conceptshift}.  The consumers in between, i.e., consumers $N_1 + 1$ to $N_2$, undergo a gradual shift from phase 1 to 2. In particular,  consumer $n \in \{N_1 + 1, \ldots, N_2\}$ belongs to phase 1 w.p.\ $1 - \frac{n - N_1}{N_2 - N_1}$ and to phase 2 w.p.\ $\frac{n - N_1}{N_2 - N_1}$. For example, with $(N_1, N_2) = (400000, 600000)$, the first 400000 consumers belong to phase 1 and the last 400000 consumers to phase 2. The middle 200000 consumers undergo a gradual concept shift with consumer 500000 belonging to phase 1 w.p.\ 0.5 and to phase 2 w.p.\ 0.5.

\begin{figure}
\centering
\begin{subfigure}[]{0.32\textwidth}
\centering
\includegraphics[width=1\linewidth]{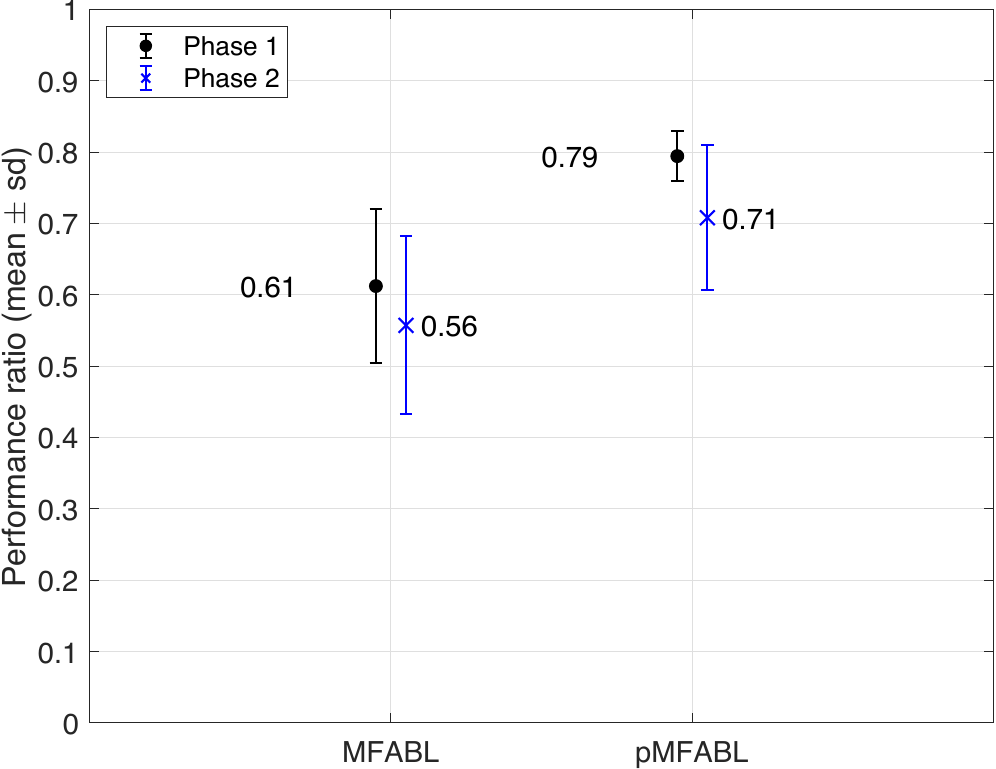}
\caption{\small $(N_1,N_2)=(400000, 600000)$}
\label{fig:ConceptGradual200000}
\end{subfigure}
\begin{subfigure}[]{0.32\textwidth}
\centering
\includegraphics[width=1\linewidth]{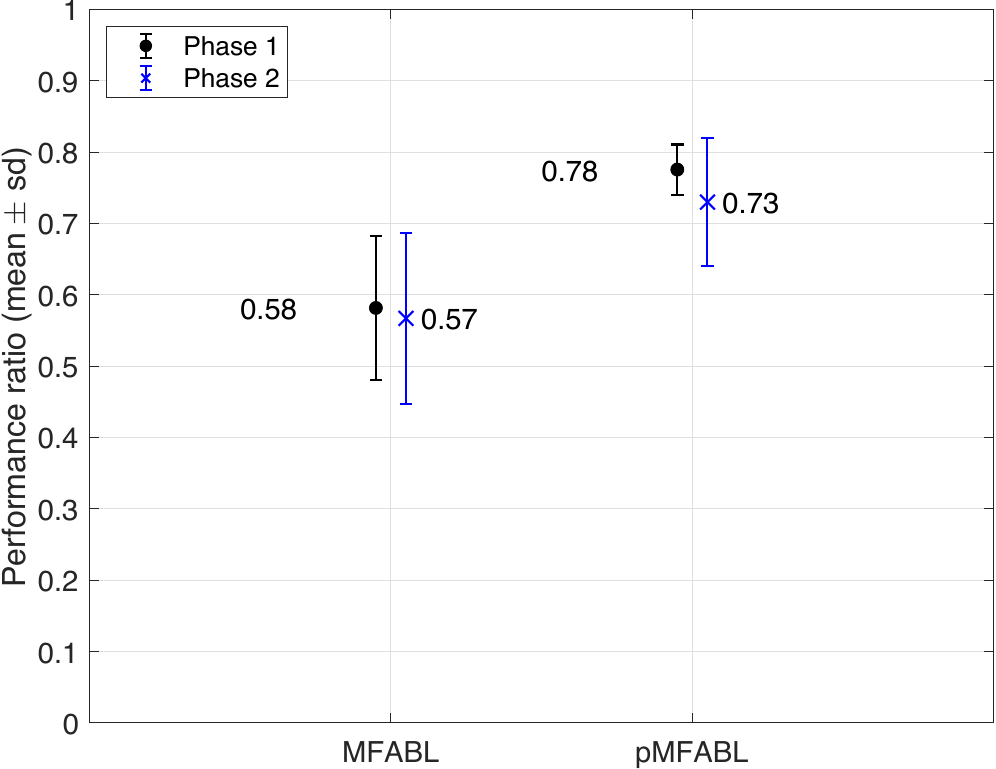}
\caption{\small $(N_1,N_2)=(250000, 750000)$}
\label{fig:ConceptGradual500000}
\end{subfigure}
\begin{subfigure}[]{0.32\textwidth}
\centering
\includegraphics[width=1\linewidth]{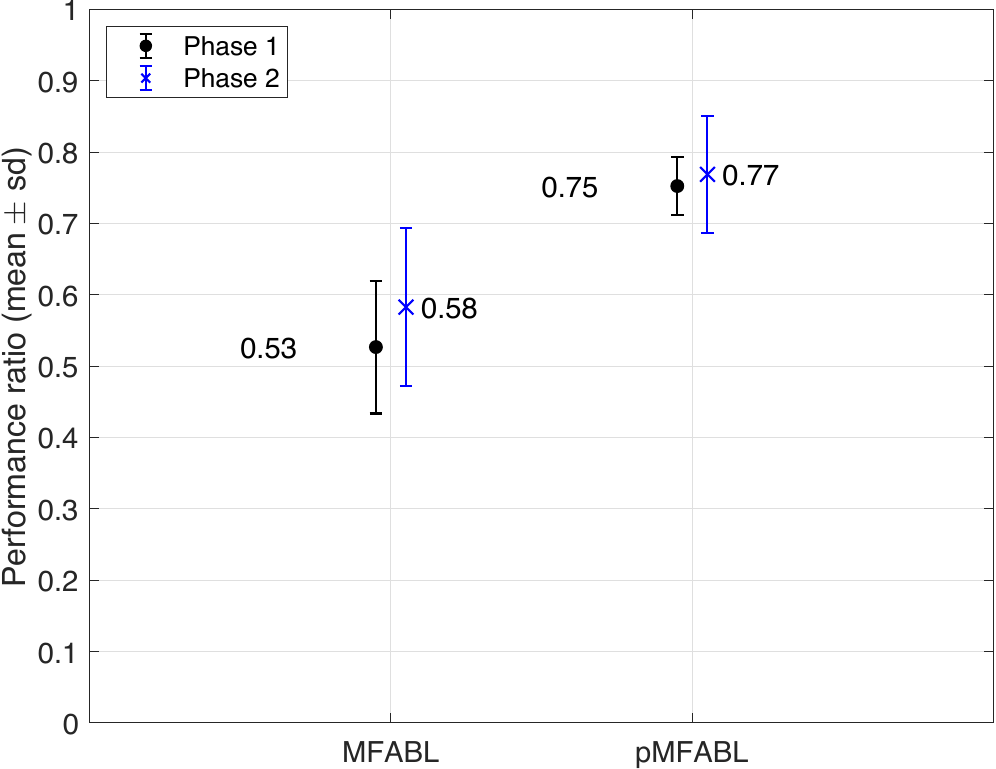}
\caption{\small $(N_1,N_2)=(1, 1000000)$}
\label{fig:ConceptGradual999999}
\end{subfigure}
\caption{\normalfont Performance of MFABL and pMFABL under gradual concept shift for three different settings of $(N_1, N_2)$.  
Though the shift here is gradual, in the legend of the plots, ``Phase 1'' refers to the first 500000 consumers and ``Phase 2'' to the last 500000 so that the numbers here are comparable to those in Figure \ref{fig:PRconcept}.}
\label{fig:GradualConcept}
\end{figure}

Given the three benchmarks (TS, PSRL, and QL-UCB) performed rather poorly under the two-phase setting of \S\ref{sec:conceptshift}, we do not expect them to perform any better under such a gradual shift. Hence, to efficiently manage our computational resources, we only simulate MFABL and pMFABL to test their robustness.  We do so via $R=100$ seeds (similar to before) and experiment with $(N_1, N_2) \in \{(400000, 600000), (250000, 750000), (1, 1000000)\}$ in order to capture various parameterizations of concept shift. The corresponding results are shown in Figure \ref{fig:GradualConcept}. Clearly, both MFABL and pMFABL are robust to such a modification in concept shift. In particular, the PR of both decreases slightly (compared to Figure \ref{fig:PRconcept}) during phase 1 (i.e., first 500000 consumers) but this decrease is offset by the increase during phase 2 (i.e., the last 500000 consumers). There is monotonicity in the sense that their phase 1 performance decreases as we move further away from the two-phase setting (i.e., from Figure \ref{fig:PRconcept} to \ref{fig:ConceptGradual200000} to \ref{fig:ConceptGradual500000} to \ref{fig:ConceptGradual999999}), but their phase 2 performance increases.  In particular, MFABL's phase 1 PR decreases from 0.64 to 0.61 to 0.58 to 0.53 whereas its phase 2 PR increases from 0.52 to 0.56 to 0.57 to 0.58. Similarly, pMFABL's phase 1 PR decreases from 0.81 to 0.79 to 0.78 to 0.75 whereas its phase 2 PR increases from 0.67 to 0.71 to 0.73 to 0.77.

\end{document}